\documentclass[acmtog,authorversion,nonacm]{acmart}

\usepackage{booktabs} 

\setcitestyle{square}

\usepackage[ruled]{algorithm2e} 

\SetAlFnt{\small}
\SetAlCapFnt{\small}
\SetAlCapNameFnt{\small}
\SetAlCapHSkip{0pt}
\IncMargin{-\parindent}

\acmJournal{TOG}

\acmSubmissionID{218}

\setcopyright{authorversion}

\gdef\cropInsets{0}

\usepackage{animate}

\usepackage{booktabs} 
\usepackage{tabularx}
\usepackage{xargs}
\usepackage{mathtools}
\usepackage{amsmath}
\usepackage{microtype}
\usepackage{units}
\usepackage[binary-units]{siunitx}
\usepackage{subcaption}
\usepackage{overpic}
\usepackage{cancel}
\usepackage{wrapfig}
\usepackage{placeins}
\usepackage{empheq}

\usepackage{multirow}

\usepackage{url} 
\usepackage[utf8]{inputenc}

\def\commentType{1}

\ifnum\commentType=0
    \newcommandx{\customComment}[3]{}
    \newcommandx{\customTODO}[3]{}
\fi
\ifnum\commentType=1
    \newcommandx{\customComment}[3]{\textcolor{#2}{\textsl{#1: #3}}}
    \newcommandx{\customTODO}[3]{\textcolor{#2}{\textsl{#1: #3}}}
\fi
\ifnum\commentType=2
    \usepackage{pdfcomment}
    \newcommandx{\customComment}[3]{\pdfcomment[icon=Comment,opacity=0.5,color=#2,author=#1]{#3}}
    \newcommandx{\customTODO}[3]{\pdfcomment[icon=Note,opacity=0.5,color=#2,author=#1]{#3}}
\fi
\ifnum\commentType=3
    \usepackage{pdfcomment} 
    \usepackage{todonotes} 
    \usepackage{silence}
    \newcommandx{\customComment}[3]{\todo[color=#2!40,size=\small]{\textbf{#1:} #3}}
    \newcommandx{\customTODO}[3]{\todo[color=#2!40,size=\small]{\textbf{#1:} #3}}
\fi

\let\originalleft\left 
\let\originalright\right 
\renewcommand{\left}{\mathopen{}\mathclose\bgroup\originalleft} 
\renewcommand{\right}{\aftergroup\egroup\originalright} 

\definecolor{amber}{rgb}{1.0, 0.49, 0.0}
\definecolor{darkgreen}{rgb}{0.0, 0.5, 0.0}
\newcommandx{\All}[1]{\customComment{All}{red}{#1}}
\newcommandx{\Jan}[1]{\customComment{Jan}{magenta}{#1}}
\newcommandx{\Alex}[1]{\customComment{Alex}{blue}{#1}}
\newcommandx{\Fabrice}[1]{\customComment{Fabrice}{amber}{#1}}
\newcommandx{\Thomas}[1]{\customComment{Thomas}{darkgreen}{#1}}

\newcommandx{\TODO}[1]{\customTODO{TODO}{red}{#1}}
\newcommandx{\JanTODO}[1]{\customTODO{Jan}{magenta}{#1}}
\newcommandx{\FabriceTODO}[1]{\customTODO{Fabrice}{amber}{#1}}
\newcommandx{\ThomasTODO}[1]{\customTODO{Thomas}{darkgreen}{#1}}

\newcommand{\REMOVE}[1]{} 
\newcommand{\ADD}[1]{#1} 

\newcommand{\useAnimations}{}

\def\equationautorefname~#1\null{%
  Equation~(#1)\null
}

\def\approxprop{%
  \def\p{%
    \setbox0=\vbox{\hbox{$\propto$}}%
    \ht0=0.6ex \box0 }%
  \def\s{%
    \vbox{\hbox{$\sim$}}%
  }%
  \mathrel{\raisebox{0.7ex}{%
      \mbox{$\underset{\s}{\p}$}%
    }}%
}

\newcommand{\Domain}{\mathcal{D}}
\newcommand{\DomainZ}{{\mathcal{P}}}

\newcommand{\Sphere}{{S^2}}

\newcommand{\Diff}[1]{\,\mathrm{d}#1}

\newcommand{\pdf}{p}

\newcommand{\PdfMC}{p}

\newcommand{\PdfNIS}{p_{\NIS}}

\newcommand{\PdfBSDF}{p_{\bsdf}}
\newcommand{\PdfUniform}{p_{\mathcal{U}}}

\newcommand{\PdfVarEst}{q}

\newcommand{\OneBlob}{\text{ob}}
\newcommand{\StopGradient}{\text{sg}}

\newcommand{\cv}{g}
\newcommand{\CV}{G}
\newcommand{\selectionProb}{c}
\newcommand{\acv}{\hat{\cv}}
\newcommand{\aCV}{\hat{\CV}}
\newcommand{\cvCoef}{\alpha}
\newcommand{\cvShape}{\bar{\cv}}
\newcommand{\Params}{\theta}
\newcommand{\cvParams}{\Params_{\cv}}
\newcommand{\CVParams}{\Params_{\CV}}
\newcommand{\acvParams}{\Params_{\acv}}
\newcommand{\aCVParams}{\Params_{\aCV}}
\newcommand{\pdfParams}{\Params_{\PdfMC}}
\newcommand{\selectionProbParams}{\Params_{\selectionProb}}
\newcommand{\nisParams}{\Params_{\NIS}}
\newcommand{\cvShapeParams}{\Params_{\cvShape}}
\newcommand{\cvCoefParams}{\Params_{\cvCoef}}
\newcommand{\NIS}{\mathrm{NIS}}
\newcommand{\conditionals}{y}

\newcommand{\JacobianProduct}{J}

\newcommand{\GenericParams}{\phi}

\newcommand{\R}{\mathbb{R}}

\newcommand{\dataSpace}{\mathcal{X}}

\newcommand{\latentSpace}{\mathcal{L}}

\newcommand{\map}{h}

\newcommand{\compoundMap}{\widehat{\map}}

\newcommand{\nnet}{m}

\newcommand{\warp}{h}
\newcommand{\warpIn}{x}
\newcommand{\warpInSpace}{\mathcal{X}}
\newcommand{\warpOut}{x^\prime}
\newcommand{\warpOutSpace}{\mathcal{X^\prime}}

\newcommand{\nlayers}{L}

\newcommand{\pos}{\mathbf{x}}
\newcommand{\diro}{\omega}
\newcommand{\diri}{\omega_\mathrm{i}}
\newcommand{\diriRV}{\Omega}
\newcommand{\normal}{{\vec{n}}}

\newcommand{\radiance}{L}
\newcommand{\inRadiance}{\radiance_{\mathrm{i}}}

\newcommand{\reflectedRadiance}{\radiance_{\mathrm{s}}}

\newcommand{\bsdf}{f_{\mathrm{s}}}

\newcommand{\fsangle}{\gamma}

\newcommand{\AreaHeuristic}{a}

\newcommand{\Loss}{\mathcal{L}}

\newcommand{\MC}[1]{{\langle {#1} \rangle}}

\newcommand{\CrossEntropy}{H}
\newcommand{\Expectation}{\mathbb{E}}
\newcommand{\Variance}{\mathbb{V}}

\newcommand{\Var}{\mathrm{Var}}
\newcommand{\Cov}{\mathrm{Cov}}

\newcommand{\N}{\mathbb{N}}

\newcommand{\Artroom}{\textsc{Artroom}}
\newcommand{\Bathroom}{\textsc{Bathroom}}
\newcommand{\Bedroom}{\textsc{Bedroom}}
\newcommand{\Bookshelf}{\textsc{Bookshelf}}
\newcommand{\Bottle}{\textsc{Bottle}}
\newcommand{\CornellBox}{\textsc{Cornell Box}}
\newcommand{\GlossyKitchen}{\textsc{Glossy Kitchen}}

\newcommand{\CountryKitchen}{\textsc{Country Kitchen}}
\newcommand{\SwimmingPool}{\textsc{Swimming Pool}}
\newcommand{\Spaceship}{\textsc{Spaceship}}
\newcommand{\SpectralBox}{\textsc{Spectral Box}}
\newcommand{\Sponza}{\textsc{Sponza Atrium}}
\newcommand{\CrytekSponza}{\textsc{Crytek Sponza}}

\newcommand{\Torus}{\textsc{Torus}}
\newcommand{\VeachDoor}{\textsc{Veach Door}}
\newcommand{\VeachLamp}{\textsc{Veach Lamp}}

\newcommand{\Staircase}{\textsc{Staircase}}

\newcommand{\Necklace}{\textsc{Necklace}}

\renewcommand{\fboxsep}{0pt}

\gdef\useCroppedImages{1}

\usepackage[export]{adjustbox}
\usepackage{calc}
\usepackage{tabularx}
\usepackage{tikz}
\usepackage{xstring}

\renewcommand{\fboxsep}{0pt}

\newlength{\beautyHeight}
\newlength{\beautyPixWidth}
\newlength{\beautyPixHeight}
\newlength{\insetvsep}
\gdef\useInsetA{0}
\gdef\useInsetB{0}
\gdef\useInsetC{0}

\newcommand{\setInset}[6]{%
    \expandafter\gdef\csname useInset#1\endcsname{1}%
    \expandafter\gdef\csname inset#1Color\endcsname{#2}%
    \expandafter\gdef\csname crop#1X\endcsname{#3}%
    \expandafter\gdef\csname crop#1Y\endcsname{#4}%
    \expandafter\gdef\csname crop#1W\endcsname{#5}%
    \expandafter\gdef\csname crop#1H\endcsname{#6}%
}

\newcommand{\unsetInset}[1]{%
    \expandafter\gdef\csname useInset#1\endcsname{0}%
}

\newcommand{\addBeautyCrop}[8]{%
    \pdfpxdimen=\dimexpr 1 in/72\relax
    \def\beauty{%
        \let\cropR\relax%
        \let\cropB\relax%
        \newlength\cropR%
        \newlength\cropB%
        \setlength\cropR{{#3 px}-{#5 px}-{#7 px}}%
        \setlength\cropB{{#4 px}-{#6 px}-{#8 px}}%
        \sbox0{\includegraphics[width=#2\textwidth,trim={#5px {\cropB} {\cropR} #6px},clip]{#1}}%
        \begin{tikzpicture}
            \node[anchor=north west,inner sep=0] at (0,0) {\usebox0};
            \begin{scope}[x=\wd0/#7, y=\ht0/#8]
            \if\useInsetA1{
                \draw[\insetAColor,thick] (\cropAX-#5,-\cropAY+#6) rectangle + (\cropAW,-\cropAH);
            }\fi
            \if\useInsetB1{
                \draw[\insetBColor,thick] (\cropBX-#5,-\cropBY+#6) rectangle + (\cropBW,-\cropBH);
            }\fi
            \if\useInsetC1{
                \draw[\insetCColor,thick] (\cropCX-#5,-\cropCY+#6) rectangle + (\cropCW,-\cropCH);
            }\fi
            \end{scope}
        \end{tikzpicture}
    }%
    \setlength\beautyHeight{\heightof{\beauty}}%
    \setlength\beautyPixWidth{#3px}%
    \setlength\beautyPixHeight{#4px}%
    \global\beautyHeight=\beautyHeight%
    \global\beautyPixWidth=\beautyPixWidth%
    \global\beautyPixHeight=\beautyPixHeight%
    \begin{adjustbox}{valign=t}
        \beauty{}
    \end{adjustbox}
}

\newcommand{\trimInset}[6]{%
    \let\cropR\relax%
    \let\cropB\relax%
    \newlength\cropR%
    \newlength\cropB%
    \setlength\cropR{{\beautyPixWidth}-{#3 px}-{#5 px}}%
    \setlength\cropB{{\beautyPixHeight}-{#4 px}-{#6 px}}%
    \color{#2}%
    \fbox{\includegraphics[width=1\linewidth,trim={{#3 px} {\cropB} {\cropR} {#4 px}},clip]{#1}}%
}

\newcommand{\addInset}[2]{%
    \color{#2}%
    \fbox{\includegraphics[width=1\linewidth]{#1}}%
}

\newcommand{\auxtimes}{x}
\newcommand{\auxplus}{+}
\newcommand{\auxspace}{ }

\newcommand{\addInsets}[1]{%
    \begin{adjustbox}{valign=t}
        \StrSubstitute{#1}{.}{-}[\baseFileName]
        \begin{adjustbox}{totalheight=1\beautyHeight,tabular={c}}
            \if\useInsetA1%
                \def\cropfile{\baseFileName-\cropAW\auxtimes\cropAH\auxplus\cropAX\auxplus\cropAY-crop}
                \if\cropInsets1
                    \immediate\write18{convert #1 -crop \cropAW\auxtimes\cropAH\auxplus\cropAX\auxplus\cropAY\auxspace -filter point -resize 800\% \cropfile.jpg}
                \fi
                \if\useCroppedImages1
                    \addInset{\cropfile.jpg}{\insetAColor}
                \else
                    \trimInset{#1}{\insetAColor}{\cropAX}{\cropAY}{\cropAW}{\cropAH}%
                \fi%
            \fi%
            \if\useInsetB1%
                \if\useInsetA1\\[\insetvsep]\fi%
                \def\cropfile{\baseFileName-\cropBW\auxtimes\cropBH\auxplus\cropBX\auxplus\cropBY-crop}
                \if\cropInsets1
                    \immediate\write18{convert #1 -crop \cropBW\auxtimes\cropBH\auxplus\cropBX\auxplus\cropBY\auxspace -filter point -resize 800\% \cropfile.jpg}
                \fi
                \if\useCroppedImages1
                    \addInset{\cropfile.jpg}{\insetBColor}
                \else
                    \trimInset{#1}{\insetBColor}{\cropBX}{\cropBY}{\cropBW}{\cropBH}%
                \fi%
            \fi%
            \if\useInsetC1%
                \if\useInsetB1\\[\insetvsep]\fi%
                \def\cropfile{\baseFileName-\cropCW\auxtimes\cropCH\auxplus\cropCX\auxplus\cropCY-crop}
                \if\cropInsets1
                    \immediate\write18{convert #1 -crop \cropCW\auxtimes\cropCH\auxplus\cropCX\auxplus\cropCY\auxspace -filter point -resize 800\% \cropfile.jpg}
                \fi
                \if\useCroppedImages1
                    \addInset{\cropfile.jpg}{\insetCColor}
                \else
                    \trimInset{#1}{\insetCColor}{\cropCX}{\cropCY}{\cropCW}{\cropCH}%
                \fi%
            \fi%
        \end{adjustbox}
    \end{adjustbox}
}

\definecolor{mathematicaBlue}{rgb}{0.38, 0.51, 0.71}
\definecolor{mathematicaOrange}{rgb}{0.88, 0.61, 0.14}
\definecolor{mathematicaGreen}{rgb}{0.56, 0.69, 0.19}
\definecolor{mathematicaRed}{rgb}{0.92,0.39, 0.21}
\definecolor{mathematicaPurple}{rgb}{0.53, 0.47, 0.7}

\definecolor{cvintegrand}{rgb}{1.0, 0.65, 0.0}
\definecolor{cvg}{rgb}{0.5, 0.0, 0.5}
\definecolor{cvG}{rgb}{0.67, 0.14, 0.19}

\definecolor{cvdifference}{rgb}{1.0, 0.65, 0.0}
\definecolor{cvpdf}{rgb}{0.5, 0.0, 0.5}

\begin{document}
\title{Neural Control Variates}

\author{Thomas M\"uller}
\affiliation{%
  \institution{NVIDIA}
  }
\email{tmueller@nvidia.com}

\author{Fabrice Rousselle}
\affiliation{%
  \institution{NVIDIA}
  }
\email{frousselle@nvidia.com}

\author{Jan Nov\'ak}
\affiliation{%
  \institution{NVIDIA}
  }
\email{jnovak@nvidia.com}

\author{Alexander Keller}
\affiliation{%
  \institution{NVIDIA}
  }
\email{akeller@nvidia.com}

\renewcommand\shortauthors{M\"uller et al.}

\begin{teaserfigure}
  \vspace{2mm}
  \begin{tabular}{cc}
    \begin{adjustbox}{valign=b}\begin{overpic}[height=4.8cm]{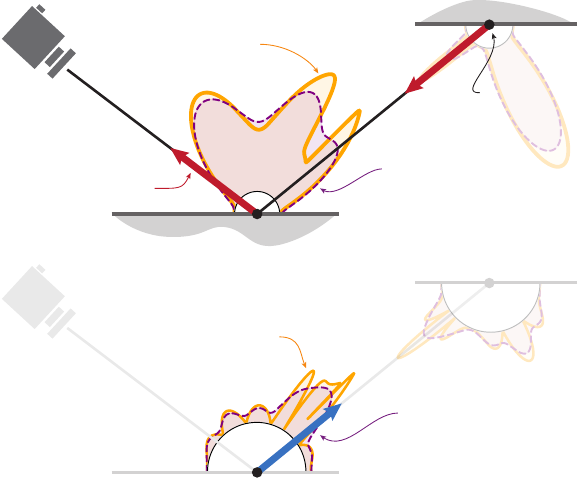}
      \put(13,80){ \small \textbf{(a)} Neural control variate }
      \put(26,74){ \footnotesize \color{cvintegrand}Integrand }
      \put(21,69){ \footnotesize \color{cvintegrand}$f = \bsdf \cdot \inRadiance \cdot \cos$ }
      \put(65,52){ \footnotesize \color{cvg}Learned control }
      \put(70,47){ \footnotesize \color{cvg}variate $\cv \approx f$ }
      \put(3,54){ \footnotesize \color{cvG}Learned }
      \put(-2,49){ \footnotesize \color{cvG}$\CV = \int \cv(\omega) \Diff{\omega}$ }
      \put(81.5,65){ \footnotesize \textbf{(c)} Heuristic }
      \put(79.5,60.5){ \footnotesize termination }
      \put(11,33){ \small \textbf{(b)} Residual neural sampling }
      \put(14,23.5){ \footnotesize \color{cvdifference}Absolute difference }
      \put(24.5,18.5){ \footnotesize \color{cvdifference}$|f - \cv|$ }
      \put(68,10){ \footnotesize \color{cvpdf}Learned PDF }
      \put(69,5){ \footnotesize \color{cvpdf}$\PdfMC \approxprop |f - \cv|$ }
    \end{overpic}\end{adjustbox} &
  \begin{adjustbox}{valign=b}\setlength{\fboxrule}{15pt}%
\setlength{\insetvsep}{30pt}%
\setlength{\tabcolsep}{-1.5pt}%
\renewcommand{\arraystretch}{1}%
\small%
\begin{tabular}{ccccc}
  & \multicolumn{2}{c}{Unbiased} & \multicolumn{1}{c}{Biased} & \\
  \cmidrule(lr){2-3}
  \cmidrule(lr){4-4}
  \textbf{(d)} Rendering results of the \VeachDoor{} scene & NIS++ & NCV & + heuristic & Reference \\
    \setInset{A}{red}{460}{650}{50}{80}%
    \setInset{B}{orange}{1135}{650}{50}{80}%
    \addBeautyCrop{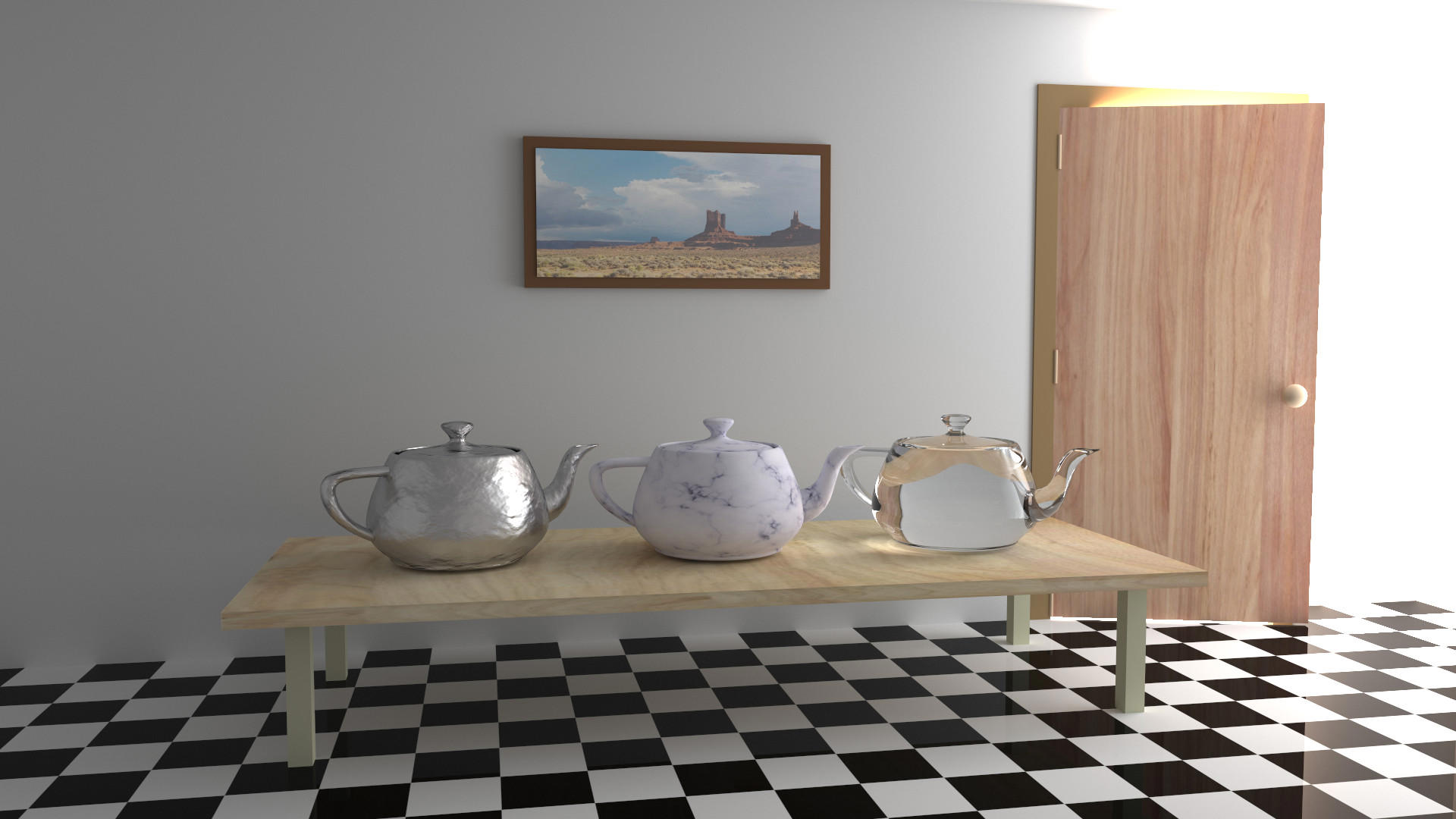}{0.333}{1920}{1080}{0}{0}{1520}{1080} &
    \addInsets{images/fig-teaser/veach-door-NIS++_u.jpg} &
    \addInsets{images/fig-teaser/veach-door-NCV_u.jpg} &
    \addInsets{images/fig-teaser/veach-door-NCV_b.jpg} &
    \addInsets{images/fig-references/veach-door-reference.jpg} \\
  \multicolumn{1}{r}{MAPE:} & 0.060 & 0.036 & 0.024 & \\
\end{tabular}
\end{adjustbox}
  \end{tabular}
  \caption{\label{fig:teaser}
    When applied to light-transport simulation, our neural control-variate algorithm \textbf{(a)} learns an approximation $\CV$ of the scattered light field and corrects the approximation error by estimating the difference between the original integrand $f$ and the corresponding learned control variate $\cv$. This is enabled by our construction that couples $\cv$ and $\CV$ such that $\cv$ always exactly integrates to $\CV$.
    To further reduce noise, we importance sample the absolute difference ${|f-\cv|}$ using a learned probability density function (PDF) $\PdfMC$ \textbf{(b)}. We also provide a heuristic \textbf{(c)} to terminate paths without estimating the difference. This reduces the mean path length and removes most of the remaining noise.
    On the right \textbf{(d)}, we compare the error of rendering the \VeachDoor{} scene using \ADD{an improved variant (NIS++) of neural importance sampling~\citep{mueller2019nis}} to our neural control variates (NCV) with and without our path termination heuristic.
  }
  \vspace{6mm}
\end{teaserfigure}
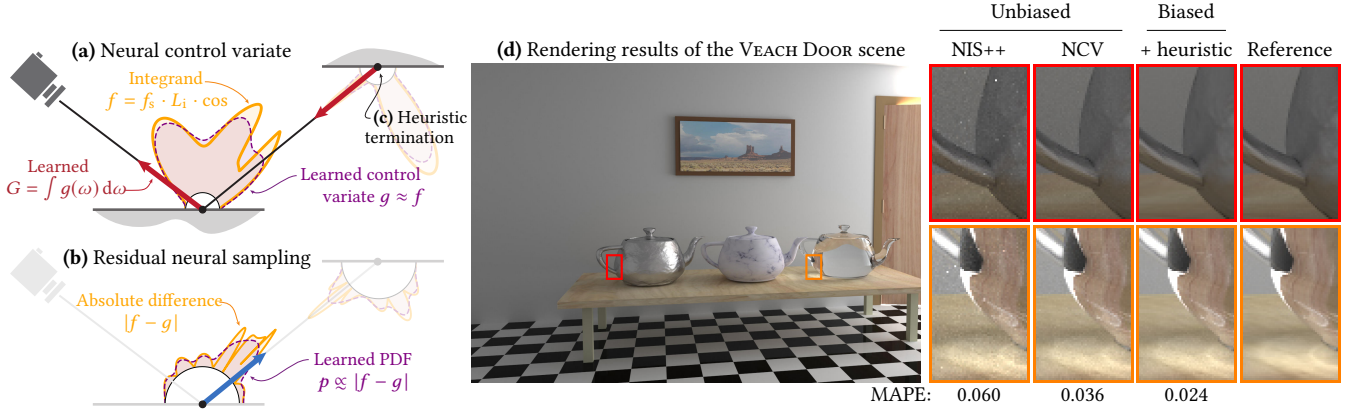

\begin{abstract}
We propose neural control variates (NCV) for unbiased variance reduction in parametric Monte Carlo integration.
So far, the core challenge of applying the method of control variates has been finding a good approximation of the integrand that is cheap to integrate.
We show that a set of neural networks can face that challenge:
a normalizing flow that approximates the shape of the integrand and another neural network that infers the solution of the integral equation.
We also propose to leverage a neural importance sampler to estimate the difference between the original integrand and the learned control variate.
To optimize the resulting parametric estimator, we derive a theoretically optimal, variance-minimizing loss function,
and propose an alternative, composite loss for stable online training in practice.
When applied to light transport simulation, neural control variates are capable of matching the state-of-the-art performance of other unbiased approaches, while providing means to develop more performant, practical solutions.
Specifically, we show that the learned light-field approximation is of sufficient quality
for high-order bounces, allowing us to omit the error correction and thereby dramatically reduce the noise at the cost of negligible visible bias.
\end{abstract}

\ccsdesc[500]{Computing methodologies~Neural networks}
\ccsdesc[500]{Computing methodologies~Ray tracing}
\ccsdesc[300]{Computing methodologies~Supervised learning by regression}
\ccsdesc[300]{Mathematics of computing~Sequential Monte Carlo methods}

\maketitle

\section{Introduction}%
\label{sec:introduction}

Monte Carlo (MC) integration is a simple numerical recipe for solving complicated integration problems.
The main drawback of the straightforward approach is the relatively slow convergence rate that manifests as high variance of MC estimators. Hence, many approaches have been developed to improve the efficiency.
Among the most frequently used ones are techniques focusing on carefully placing samples, e.g.\ antithetic sampling, stratification, quasi-random sampling, or importance sampling.
A complimentary way to further reduce variance is to leverage hierarchical integration or the concept of control variates.
In this article, we focus on the latter approach and present parametric control variates based on neural networks.

Reducing variance by control variates (CV) amounts to leveraging an approximate solution of the integral corrected by an estimate of the approximation error.
The principle is given by the following identity:
\begin{align}
    F = \int_\Domain f(x) \Diff{x} =  \cvCoef \cdot \CV + \int_\Domain f(x) -  \cvCoef \cdot \cv(x) \Diff{x}\,.
    \label{eq:basic-cv}
\end{align}
Instead of integrating the original function $f$ to obtain the solution $F$, we leverage an $\cvCoef$-scaled approximation $G$, that corresponds to integrating a (different) function $\cv$---the \emph{control variate}---over the same domain $\Domain$, i.e.\ $\CV = \int_\Domain \cv(x) \Diff{x}$ (we drop $\Domain$ in the rest of this section for brevity).
The approximation error is corrected by adding an integral of the difference $f(x) - \cvCoef\cdot\cv(x)$; this makes the right-hand side equal to the left-hand one.

The numerical efficiency of estimating the right-hand side, relative to estimating the original integral,
depends on the scaled control variate making the integration easier, for example by making the integrand smoother as illustrated in \autoref{fig:cv-illustration}.
This will typically be the case as long as $f$ and $\cv$ are (anti-)correlated.
In fact, the scaling coefficient $\cvCoef$, which controls the strength of applying the CV, should be derived from the correlation of the two functions.
In a nutshell, a successful application of control variates necessitates a $\cv$ that approximates the integrand $f$ sufficiently well, and permits an efficient evaluation and computation of $\CV$ and $\cvCoef$.

In this work, we propose to infer the control variate $\cv$ from observations of~$f$ using machine learning.
Since the control variate is learned, the main challenge becomes representing it in a form that permits (efficiently) computing its integral, $\CV = \int \cv(x) \Diff{x}$.
We propose to sidestep this integration problem by introducing a CV model that satisfies $\CV = \int \cv(x) \Diff{x}$ \emph{by construction}: we decompose the control variate $\cv(x)$ into its normalized form---the shape $\cvShape(x)$---and its integral $\CV$, such that $\cv(x) = \cvShape(x) \cdot \CV$.
The shape and the integral can be modeled independently.
We infer the integral $\CV$ and the coefficient $\cvCoef$ using one neural network for each.
For the shape $\cvShape$, we leverage a tailored variant of normalizing flows, which are capable of representing normalized functions.
The parameters of the flow are inferred using a set of neural networks.

When the control variate is designed well, the residual integral $\int f(x) - \cvCoef\cdot\cv(x)\Diff{x}$ carries less energy than the original integral $\int f(x) \Diff{x}$. However, the residual \emph{integrand} may feature shapes that are hard to sample with hand-crafted distributions; this is why many prior works in graphics did not demonstrate significant efficiency gains when combining control variates with importance sampling.

We address this by employing neural importance sampling (NIS) as proposed by \citet{mueller2019nis} that is capable of importance sampling arbitrary integrands, including the residual ones in our case.
We show that an estimator that utilizes both techniques, NCV and NIS, features the strengths of each approach as long as all trainable parameters are optimized jointly; to that end we derive two loss functions, one theoretically optimal and one that yields robust optimization in practice.

We demonstrate the benefits of neural control variates on light-transport simulations governed by Fredholm integral equations of the second kind.
These are notoriously difficult to solve efficiently due to their recursive nature, often necessitating high-dimensional samples in the form of multi-vertex transport paths (obtained using e.g.\ path tracing).
In this context, control variates offer two compelling advantages over prior works that only focus on placing the samples.
First, control variates reduce the number of constructed path vertices as the difference integral typically carries less energy than the original integral. Paths can thus be terminated earlier using the learned scattered radiance as an approximation of the true scattered radiance; we propose a heuristic that minimizes the resulting bias.
Second, control variates trivially support spectrally resolved path tracing by using a different $\cv$ for each spectral band.
To avoid computational overhead of using potentially \emph{many} control variates, we develop a novel normalizing flow that can represent multiple (ideally correlated) control variates at once.
Spectral noise, which is typical for importance sampling that only targets scalar distributions, is thus largely suppressed.
The benefits are clearly notable in several of our test scenes.

In summary, we present the following contributions:
\begin{itemize}
    \item a tractable neural control variate modeled as the product of normalized shape $\cvShape$ and the integral $\CV$,
    \item a multi-channel normalizing flow for efficient handling of spectral integrands,
    \item an estimator combining neural control variates with neural importance sampling, for which we present
    \item a variance-optimal loss derived from first principles, and an empirical composite loss that yields stable online optimization on noisy estimates of $f(x)$, and finally,
    \item a practical light-transport simulator that heuristically omits estimating the residual integral when visible bias is negligible.
\end{itemize}

\section{Related Work}%
\label{sec:related-work}

\begin{figure}[t]
    \vspace{4mm}
    \begin{overpic}[width=\columnwidth]{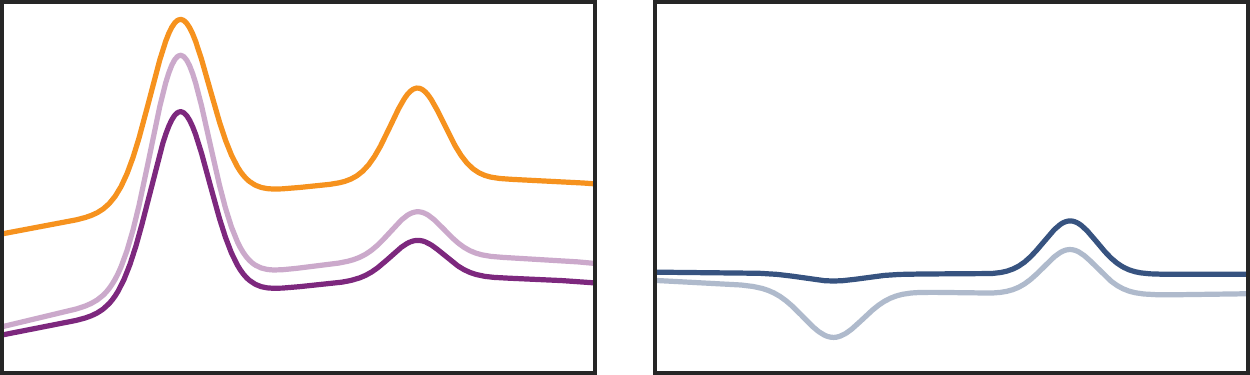}
        \put(40.5, 17){\small $f(x)$}
        \put(40.5, 10.5){\small $\cv(x)$}
        \put(38, 4.2){\small $\cvCoef\cv(x)$}
        \put(80.5, 14){\small $f(x) - \cvCoef\cv(x)$}
        \put(83, 3.2){\small $f(x) - \cv(x)$}
    \end{overpic}
    \vspace{-2mm}
    \caption{\label{fig:cv-illustration}
        Computing an integral $F$ of a function $f(x)$ with the help of a control variate $g(x)$ (left) amounts to using a known (or efficient-to-compute) integral $\CV=\int g(x)\Diff{x}$ and adding the integrated difference $f(x)-\cv(x)$ (right).
        The overall variance may be reduced by minimizing the variance of the difference using an $\cvCoef$-scaled control variate, where $\cvCoef = {\Cov(f, \cv)}/{\Var(\cv)}$.
    }
\end{figure}
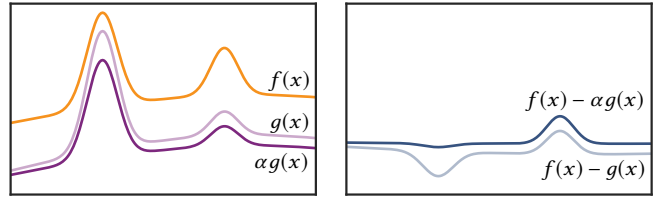

The application of control variates has been explored in many fields, predominantly in the field of financial mathematics and operations research, see~\citep{Kemna:1990,Broadie:1998,Hesterberg:1998} for examples.
Later on, \citet{Glynn:2002} focused on relating the concept to antithetic sampling, rotation sampling, and stratification, among other techniques. In computer graphics,
\citet{Rousselle:2016:ICV} link control variates to solving the Poisson equation in screen space and
\citet{Kondapaneni:2019} use the concept to interpret their optimally weighted multiple-importance sampler.

Since a poorly chosen control variate may even decrease efficiency, early research focused on an efficient and accurate estimation of the scaling coefficient $\cvCoef$.
While the optimal, variance-minimizing value of $\cvCoef$ is known to be ${\Cov(f, \cv)}/{\Var(\cv)}$ (see \autoref{fig:cv-illustration} for an illustration), estimating it numerically may introduce bias if done using samples correlated to the samples used for the actual estimate~\citep{Lavenberg:1982,Nelson:1990}.
We resolve this issue by providing recipes for obtaining $\cvCoef$ that do not bias the estimator.

We are not the first to apply control variates to light transport simulation.
Lafortune and Willems successfully leveraged CVs based on ambient illumination~\shortcite{Lafortune:1994:TAT} and hierarchically stored radiance values~\shortcite{Lafortune:1995:A5T} to accelerate the convergence of path tracing.
\citet{PegoraroEGSR08SMCALAPM,PegoraroIRT08TIGIESMCA} applied a similar idea to volumetric path tracing, but were restricted to near-isotropic volumes.
\citet{Fan:2006:OCV} and \citet{Kondapaneni:2019} use a linear combination of multiple importance-sampling densities as a control variate, which is an adaptation of a technique by \citet{owen00safe}.
Others proposed to apply CVs to carefully chosen subproblems, such as estimating direct illumination~\citep{Szecsi:2004:CCI,Clarberg:2008}, sampling free-flight distances in participating media~\citep{SzirmayKalos11,Novak:14:RRT,georgiev19integral}, or unbiased denoising and re-rendering~\citep{Rousselle:2016:ICV,yamaguchi18efficient}.

One of the challenges of successfully applying control variates is an efficient estimation of the residual integral $\int f(x) - \cv(x) \Diff{x}$; this is typically harder than (importance) sampling $f$ alone. We demonstrate that parametric trainable control variates can be well complemented by trainable importance samplers (such as neural importance sampling~\citep{mueller2019nis}) yielding better results than each technique in isolation.

\paragraph{Multi-level Monte Carlo integration}
Heinrich~\shortcite{Heinrich:1998,Heinrich:2000} proposed to apply the CV concept in a hierarchical fashion: Each successive estimator of a difference improves the estimate of its predecessor. This technique is known as multi-level Monte Carlo integration and it has been applied in stochastic modeling~\citep{Giles:2008}, solving partial differential equations~\citep{Barth:2011}, or image synthesis~\citep{Keller:2001}; see the review by \citet{giles:2013} for other applications.
While the allocation of samples across the estimators is key to efficiency, classic representations of functions quickly render the approach intractable in higher dimensions.

\paragraph{Realistic image synthesis with neural networks}
Similar to Monte Carlo methods for high-dimensional integration, neural networks are especially helpful in high-dimensional approximation.
In computer graphics, they have been shown very suitable for compressing and inferring fields of radiative quantities (or their approximations) in screen space~\citep{Nalbach2017b}, on surfaces~\citep{Ren:2013,Vicini:2019,Thies:2019,Maximov_2019_ICCV}, on point clouds~\citep{hermosilla2019}, or in free space~\citep{Kallweit2017DeepScattering,Lombardi:2019,Meka2019,Sitzmann2018DeepVoxelsLP}; see the survey by \cite{tewari2020state} for additional examples.
These approaches are largely orthogonal to our technique. In fact, many of these ideas may improve the learning and representation of the approximate solution $\CV$ in specific situations. For instance, one could employ voxel grids with warping fields instead of multi-layer perceptrons~\citep{Lombardi:2019}, combat overfitting using mip-level hierarchies~\citep{Thies:2019}, or handle scene partitions using dedicated networks~\citep{Ren:2013}; shall the application need it.
Leaving these as possible future extensions, we instead focus on a shortcoming that is common to all the aforementioned approaches: occasional deviations from the ground-truth solution observable as e.g.\ patchiness, loss of contrast, or dull highlights.
We propose to correct the errors using the mechanism of control variates, i.e.\ we add an estimate of the difference between the correct solution and the approximation to recover unbiased results with error manifesting merely as noise.
We view our neural control variates as a step towards bringing data-driven and physically-based rendering~closer.

\paragraph{Normalizing flows}
Normalizing flows~\citep{Tabak:2010,Tabak:2013} are a technique for mapping arbitrary distributions to a base distribution; e.g.\ the normal distribution.
The mappings are formally obtained by chaining an infinite series of infinitesimal transformations, hence the name flow.
The technique has been successfully leveraged for variational inference, either in the continuous form~\citep{chen2018neural} or as a finite sequence of warps~\citep{dinh2014nice,rezende2015variational}.
Numerous improvements followed soon after:\
the modeling power of individual transforms has been enhanced using non-volume preserving warps~\citep{dinh2016density}, piecewise-polynomial warps~\citep{mueller2019nis}, or by injecting learnable $1\times1$ convolutions between the warps~\citep{kingma18glow}.
Others have demonstrated benefits by formulating the estimation autoregressively~\citep{Huang2018NeuralAF,kingma2016improved,papamakarios2017masked}; we refer the reader to the surveys by \citet{papamakarios2019normalizing,kobyzev2019normalizing} for an introduction and comparisons of different approaches.

In light transport simulation, \citet{Zheng:2019} and \citet{mueller2019nis} leverage modified normalizing flows to learn and sample from parametric distributions.
In analogy, we use our multi-channel flow to represent the spectrally resolved per-channel normalized form $\cvShape$ of the control variate $\cv$.

\paragraph{(Neural) Control Variates based on Stein's identity and beyond}
\citet{PhysRevLett.83.4682} suggest representing the control variate in terms of the score function $s(x) = \nabla \log{\PdfMC(x)}$, where $\PdfMC(x)$ is the importance-sampling density.
The score function has zero expectation, i.e.\ ${\Expectation_{\PdfMC}[s(x)] = 0}$, trivially allowing its use as a control variate of a stochastic estimator.
Through Stein's~\shortcite{stein1972} identity, the score function can be reparameterized to act as an effective control variate.
Many such reparameterizations were proposed, be they parametric polynomials~\citep{PhysRevLett.83.4682,Mira2013ZeroVM}, non-parametric~\citep{Oates2014ControlFF}, or parameterized by neural networks~\citep{Grathwohletal2018relax,wan2019neural}.

In contrast to our use of normalizing flows, using the score function and Stein's identity as a control variate has one major limitation:\ the integral of the control variate $\CV$ is unknown---one only knows that the \emph{expectation} of the control variate under samples from $\PdfMC$ is zero.
This limitation results in the following practical shortcomings: (i) it is not possible to use the CV integral $\CV$ as a light-field approximation in the way we propose, and (ii) it is difficult to adapt the sampling density $\PdfMC$ to the control variate; optimizing the sampling density to importance sample the residual difference $|f - \cv|$ would alter the score function and thereby the control variate, creating a circular dependency.
In future work, it may be possible to derive a joint optimization between score-function-based control variates and importance sampling similar to our unbiased variance loss.
Beyond Stein's identity, neural networks were also used control variates based on the Martingale representation theorem for solving partial differential equations~\citep{vidales2018unbiased}.

\section{Parametric Trainable Control Variates}%
\label{sec:model}

In this section, we propose a novel model for trainable control variates in the context of integro-approximation: our goal is to reduce the variance of estimating the parametric integral
\begin{align}
    F(\conditionals) &= \int _\Domain f(x,\conditionals) \Diff{x} \nonumber \\
    &= \cvCoef(\conditionals) \cdot \CV(\conditionals) + \int_\Domain f(x,\conditionals) -  \cvCoef(\conditionals) \cdot \cv(x,\conditionals) \Diff{x}\,
    \label{eq:parametric-cv-equation-with-conditionals}
\end{align}
parameterized by $\conditionals$ using the control variate $\cv$.
This means that we need to represent and approximate functions besides computing integrals.
For instance, in the light transport application of~\autoref{sec:light-transport}, $F(\conditionals)$ is the reflected radiance and the parameter $\conditionals$ represents the reflection location and direction.

In many applications and especially in computer graphics, the functions $F(y)$ and $f(x,y)$ may have infinite variation and lack smoothness.
Their models thus need to be sufficiently flexible and highly expressive.
Therefore, we make the design decision to model the CV using neural networks driven by an optimizable set of parameters $\cvParams$;
a discussion of alternatives is deferred to \autoref{sec:discussion-and-future-work}.

\paragraph{Tractable neural control variates}
In order to use \autoref{eq:parametric-cv-equation-with-conditionals}, the neural model must permit an efficient evaluation of the control variate $\cv(x,y;\cvParams)$ and its integral $\CV(y)=\int \cv(x,y;\cvParams)\Diff{x}$.
This turns out to be the key challenge.
Modeling $\cv$ using a neural network may be sufficiently expressive, but computing the integral $\CV$ would require some form of numerical integration necessitating multiple forward passes to evaluate $\cv(x,y;\cvParams)$; a cost that is too high.

We avoid this issue by restricting ourselves to functions where the integral is known.
Specifically, we consider \emph{normalized} functions that integrate to $1$.
Arbitrary integrands can still be matched by scaling the normalized function by a (learned) factor.
Hence, our parametric control variate
\begin{align} \label{eq:parametric-cv-definition}
    \cv(x,y; \cvParams) := \cvShape(x,y;\cvShapeParams) \cdot \CV(y;\CVParams)
\end{align}
is defined as the product of two components: a parametric normalized function $\cvShape(x, y;\cvShapeParams)$ and a parametric scalar value $\CV(y;\CVParams)$.
From now on, we refer to $\cvShape$ and $\CV$ as the \emph{shape} and the \emph{integral} of the CV, each of which is parameterized by its own set of parameters and $\cvParams := \cvShapeParams \cup \CVParams$.
This decomposition has the advantage that computing the integral $\CV$ amounts to evaluating a neural network once, rather than performing a costly numerical integration of $ \cv(x,y; \cvParams)$ that requires a large number of network evaluations.

The rest of this section proposes parametric models for the shape (\autoref{sec:modeling-shape}), the integral (\autoref{sec:modeling-integral}), and the coefficient (\autoref{sec:modeling-coefficient}) of the control variate.
Sections \ref{sec:mc-integration} and \ref{sec:optimization} then describe an efficient combination with a parametric importance sampler and the optimization of all trainable parameters.

\subsection{Modeling the Shape of the Control Variate}
\label{sec:modeling-shape}

We now address the main challenge of modeling CVs using neural networks: learning normalized functions that we use to represent the shape $\cv(x,y;\cvParams)$ of the CV.
Normalizing the output of a neural network is generally difficult.
We thus resort to a class of models where the network output is used to merely parameterize a transformation, which can be used to warp a function without changing its integral.
This allows for learning functions that are normalized \emph{by construction}.
Such models are referred to as normalizing flows (see e.g.~\cite{kobyzev2019normalizing,papamakarios2019normalizing}).
In what follows, we briefly review the concept of normalizing flows and discuss the details of using them to learn the shape of the CV.\@

\begin{figure*}[t]
    \vspace{0mm}
    \footnotesize
    \begin{overpic}[width=\linewidth]{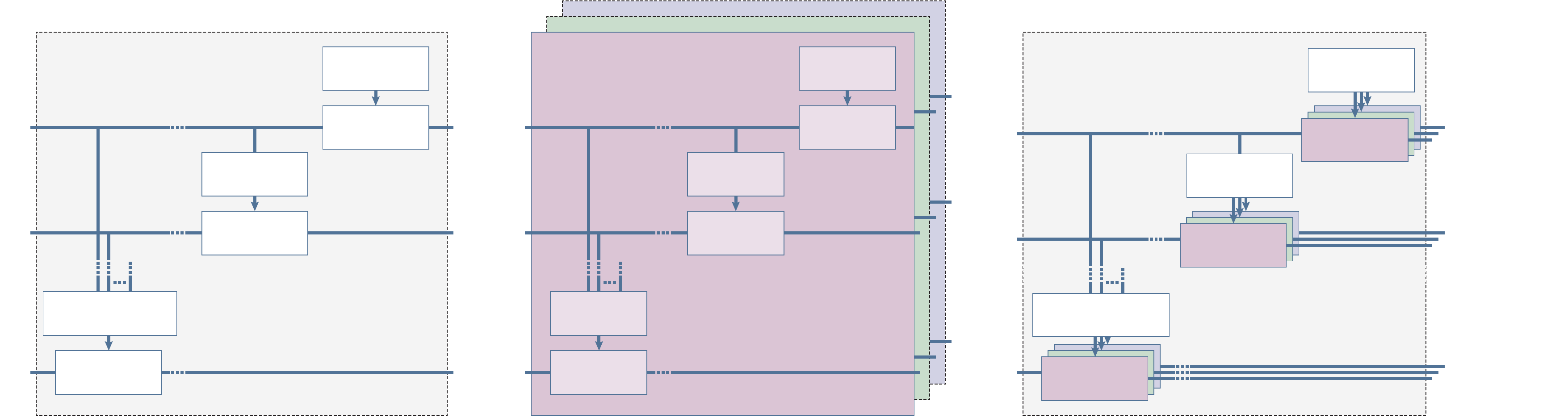}
        \put(8,-3){\textbf{(a)} Autoregressive sub-flow}
        \put(39,-3){\textbf{(b)} Per-channel sub-flows}
        \put(72,-3){\textbf{(c)} Multi-channel flow}
        \put(0.2, 18.3){$x_i^0$}
        \put(0.2, 11.5){$x_i^1$}
        \put(0.2, 2.7){$x_i^D$}
        \put(29.2, 18.3){$x_{i+1}^0$}
        \put(29.2, 11.5){$x_{i+1}^1$}
        \put(29.2, 2.7){$x_{i+1}^D$}
        \put(22, 21.9){$\nnet(\GenericParams_i^0)$}
        \put(21.8, 18.1){$\map(x_i^0;\cdot)$}
        \put(13.4, 15.2){$\nnet(x_i^0;\GenericParams_i^1)$}
        \put(14, 11.3){$\map(x_i^1;\cdot)$}
        \put(3.2, 6.3){$\nnet(x_i^{<\!D};\GenericParams_i^D)$}
        \put(4.5, 2.6){$\map(x_i^D;\cdot)$}
        \put(63, 17.9){$x^0$}
        \put(63, 11.1){$x^1$}
        \put(63, 2.7){$x^D$}
        \put(92.2, 17.9){$(\widehat{x}^0_\mathrm{r}, \widehat{x}^0_\mathrm{g}, \widehat{x}^0_\mathrm{b})$}
        \put(92.2, 11.1){$(\widehat{x}^1_\mathrm{r}, \widehat{x}^1_\mathrm{g}, \widehat{x}^1_\mathrm{b})$}
        \put(92.2, 2.7){$(\widehat{x}^D_\mathrm{r}, \widehat{x}^D_\mathrm{g}, \widehat{x}^D_\mathrm{b})$}
        \put(85.2, 21.6){$\nnet(\GenericParams^0)$}
        \put(84.2, 17.3){$\map(x_\mathrm{r}^0;\cdot)$}
        \put(76.1, 15.0){$\nnet(x^0;\GenericParams^1)$}
        \put(76.8, 10.7){$\map(x_\mathrm{r}^1;\cdot)$}
        \put(66.5, 6.0){$\nnet(x^{<\!D};\GenericParams^D)$}
        \put(67, 2.0){$\map(x_\mathrm{r}^D;\cdot)$}
    \end{overpic}
    \vspace{2mm}%
    \caption{
        \label{fig:autoregressive}
        We model the shape of the control variate using bijective transformations assuming autoregressive structure \textbf{(a)} as proposed by \citet{kingma2016improved} in the context of probabilistic generative models. Concatenating multiple autoregressive blocks (sub-flows) increases the expressivity of the model.
        To handle multiple control variates (e.g.\ one for each color channel), one can instantiate a flow for each ``channel'' \textbf{(b)};  we avoid repeating the expressions in (b) for brevity; the only difference to the left illustration is that all $x$ and $\GenericParams$ would have a \emph{channel} subscript.
        For applications where a single sub-flow is sufficient, such as the one discussed in \autoref{sec:light-transport}, we propose to use a single network across all channels \textbf{(c)} to keep the evaluation cost largely agnostic to the channel count.
    }
    \vspace{-1mm}
\end{figure*}
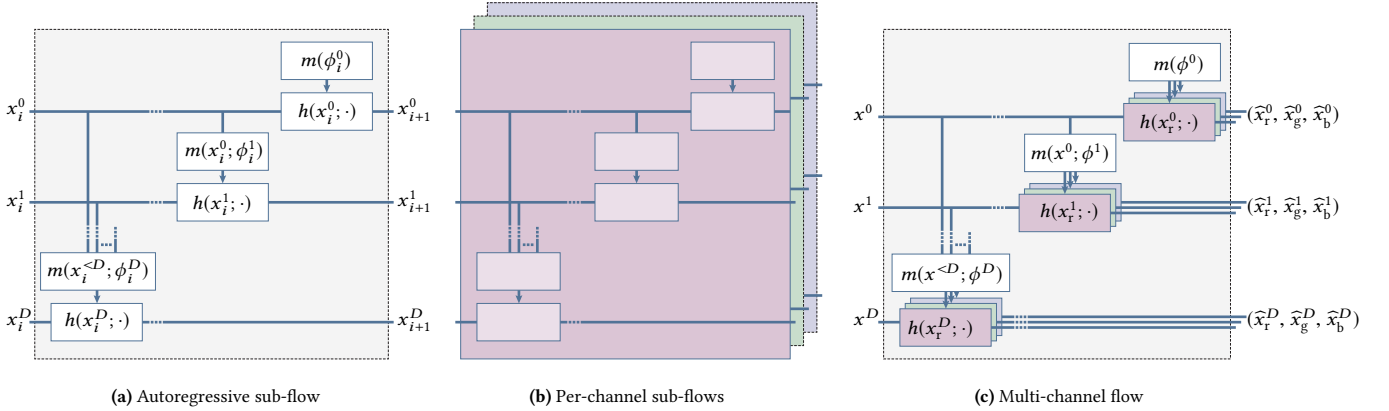

\paragraph{Normalizing flow preliminaries}
A normalizing flow is a differentiable, multi-dimensional, compound mapping for transforming probability densities.
The mapping $\compoundMap = \map_\nlayers \circ \cdots \circ \map_2 \circ \map_1$ comprises $\nlayers$ bijective warping functions; it is therefore also bijective as a whole.
The warping functions $\warp$: ${\warpInSpace \rightarrow \warpOutSpace}$ induce a density change according to the change-of-variables formula
\begin{align}
    \pdf_{\warpOutSpace}(\warpOut) = \pdf_\warpInSpace(\warpIn) \cdot \left| \det \left( \frac{\partial \warp(\warpIn)}{\partial \warpIn^T}\right) \right|^{-1}\,,
    \label{eq:change-of-variables}
\end{align}
where $\pdf$ is a probability density, ${\warpIn\in\warpInSpace}$ is the argument of the warp, ${\warpOut=\warp(\warpIn)\in\warpOutSpace}$ is the output of the warp, and $\left(\frac{\partial \warp(\warpIn)}{\partial \warpIn^T}\right)$ is the Jacobian matrix of $\warp$ at~$\warpIn$.

The density change induced by a chain of $\nlayers$ such warps can be obtained by invoking the chain rule. This yields the following product of absolute values of Jacobian determinants:
\begin{align}
    \JacobianProduct(x
    ) = \prod_{i=1}^{\nlayers} \, \left| \det \left(\frac{\partial \map_i(x_i
    )}{\partial x_i^T} \right)\right|\,,
    \label{eq:chain-jacobian}
\end{align}
where now ${x_1 \equiv x}$.
The $i$-th term in the product represents the absolute value of the Jacobian determinant of the $i$-th warp with respect to the output of warp ${i-1}$.

The transformed variable ${\widehat{x} = \compoundMap(x)}$ is often referred to as the latent variable in latent space $\latentSpace$.
Its distribution is related to the distribution of the input variable by combining Equations~(\ref{eq:change-of-variables},\ref{eq:chain-jacobian}):
\begin{align}
    \pdf_\latentSpace(\widehat{x}) = \frac{\pdf_\dataSpace(x)}{\JacobianProduct(x)}.
\end{align}
The distribution of latent variables $\pdf_\latentSpace(\widehat{x})$ is typically chosen to be simple and easy to sample; we use the uniform distribution $\pdf_\latentSpace(\widehat{x}) \equiv \PdfUniform(\widehat{x})$ over the unit hypercube.

In order to achieve high modeling power, neural normalizing flows utilize \emph{parametric} warps that are driven by the output of neural networks.
To allow for modeling correlations across dimensions, the outputs of individual warps need to be fed into neural networks conditioning the subsequent warps in the flow.
In the context of probabilistic modeling, two main approaches have been proposed to that end: autoregressive flows \cite{rezende2015variational,kingma2016improved,papamakarios2017masked,Huang2018NeuralAF} and coupling flows \cite{dinh2014nice,dinh2016density,mueller2019nis}.
Both of these approaches yield flows that are (i) invertible, (ii) avoid the cubic cost of computing determinants of dense Jacobian matrices, and (iii) avoid the need to differentiate through the neural network to compute relevant entries in the Jacobian.

In this work, efficient invertibility of the flow is not needed as modeling the CV shape requires evaluating the flow in only one direction.
However, we still take advantage of the previously proposed autoregressive formulation to ensure tractable Jacobian determinants. Furthermore, we show that the model can be further accelerated in cases when multiple densities---specifically, multiple channels of the control variate---are being learned.

\paragraph{Modeling the CV shape with normalizing flows}
Leveraging a normalizing flow to represent the shape $\cvShape$ of the control variate is straightforward.
We use the unit hypercube with the same number of dimensions $D$ as $\cv$ to be the latent space $\latentSpace$.
The normalized CV is then modeled as
\begin{align}
    \cvShape(x) := \pdf_\dataSpace(x;\cvShapeParams) = \pdf_\latentSpace(\widehat{x}) \cdot \JacobianProduct(x;\cvShapeParams).
    \label{eq:flow-cv-shape}
\end{align}
It is worth noting that the product on the right-hand side is normalized by construction:
the probability density $\pdf_\latentSpace$ is normalized by definition and each warp in the flow merely redistributes the density without altering the total mass.
This is key for ensuring that $\cvShape$ is and remains normalized during training.

In our implementation, the warps in the normalizing flow assume an autoregressive structure: dimension $d$ in the output $x_{i+1}$ of the $i$-th warp is conditioned on only the preceding dimensions in the input $x_i$:
\begin{align}
x_{i+1}^d =  \map\big(x_{i}^d; \nnet(x_{i}^{<d};\GenericParams_{i}^{d})\big)\,,
\end{align}
where the superscript $<\!d$ denotes the preceding dimensions, $\nnet$ is a neural network, and $\GenericParams_{i}^{d}$ are its parameters, with $\cvShapeParams = \bigcup_i \bigcup_d \GenericParams_{i}^{d}$.
This ensures tractable Jacobian determinants that are computed as the product of diagonal terms in the Jacobian matrix of $\map$.
The diagonal terms are specific to the transform $\map$ being used---we use piecewise-quadratic warping functions proposed by \citet{mueller2019nis} in our implementation.

\autoref{fig:autoregressive}(a) illustrates the autoregressive structure of the $i$-th warp in the normalizing flow. We adopt the terminology of \citet{papamakarios2019normalizing} and refer to one autoregressive block as the ``sub-flow''.
We utilize an independent network for inferring the warp of each dimension. The alternative of using a single network for all dimensions requires elaborate masking~\citep{germain2015made,papamakarios2017masked} to enforce the autoregressive structure.
Having an independent network per dimension simplifies the implementation and, more importantly, facilitates network sharing when dealing with multi-channel control variates.

\paragraph{Multi-channel CV}
Many integration problems simultaneously operate on multiple, potentially correlated channels.
In this article, for instance, we estimate spectrally resolved integrals; one for each RGB channel.
In order to minimize the variance per channel, it is advantageous to use a separate control variate for each channel rather than sharing one CV across all channels.

The most straightforward solution is to instantiate a distinct normalizing flow for each channel; the per-channel sub-flows are illustrated in \autoref{fig:autoregressive}(b); symbols were dropped for brevity.
Unfortunately, this makes the computation cost linear in the number of channels---a penalty that we strive to avoid.

We propose to keep the cost largely constant by sharing corresponding neural networks across the channels.
However, since network sharing introduces correlations across channels, e.g.\ red dimensions can influence green dimensions, special care must be taken to constrain the model correctly.

Merely concatenating the inputs to the $k$-th network across the per-channel flows, and instrumenting the network to produce parameters for warping dimensions in \emph{all} $n$ channels, is problematic as it corresponds to predicting a \emph{single} normalized ${(n\times D)}$-dimensional function. Instead, we need $n$ \emph{individually} normalized, $D$-dimensional functions, like in the case of instantiating a distinct flow for each channel.
We must ensure that each channel of the CV is normalized \emph{individually}.

Note that since channels can influence each other only after the first sub-flow, the first sub-flow produces individually normalized functions, even if the networks are shared across the channels.
This is easy to verify by inspecting the $nD \times nD$ Jacobian matrix constructed for all dimensions in all channels.
The matrix will have a block-diagonal structure, where each $d\times d$ block corresponds to the Jacobian matrix of one of the channels. All entries outside of the blocks on the diagonal will be zero.
This observation allows us to share the networks as long as we use only \emph{one} sub-flow to model each channel of the CV shape; as illustrated in \autoref{fig:autoregressive}(c).
The benefits of sharing the networks are studied in \autoref{fig:toy}.

\subsection{Modeling the Integral of the Control Variate}
\label{sec:modeling-integral}
Representing the integral value by a neural network $\CV(y; \CVParams)$ is fairly straightforward as we can use any architecture.
We exponentiate the network output to ensure that the CV integral is always positive.
The combination of the exponentiated network output and the normalizing flow for the CV shape constrains the CV to be a non-negative function; negative values are excluded by design. This is desired for the light-transport application in \autoref{sec:light-transport} that deals with non-negative integrands only.

Note that even without the exponentiation, the neural CV may only be non-negative or non-positive.
Then, signed integrands may be handled using the extension described in \autoref{sec:handling-of-signed-integrands}.

\subsection{Modeling the CV Coefficient}
\label{sec:modeling-coefficient}
Since the control variate may not match $f$ perfectly---our neural CV is no exception---the variate is weighted by the CV coefficient $\cvCoef$ that controls its contribution.
The optimal, variance-minimizing value of $\cvCoef(\conditionals)$ is known to be ${\Cov(f(x,\conditionals), \cv(x,\conditionals))}/{\Var(\cv(x,\conditionals))}$~\citep{Lavenberg:1982,
Nelson:1990}. However, computing the optimal value, which generally varies with $\conditionals$, can be prohibitively expensive in practice.
We thus model the coefficient using a neural network $\cvCoef(y;\cvCoefParams)$ \ADD{with a sigmoid output activation that constrains its value to the interval $(0, 1)$ for numerical robustness.}
The network is trained to output the appropriate contribution of the CV in dependence on the parameter $\conditionals$.
In \autoref{sec:optimization}, we contribute a loss function for optimizing the neural network $\cvCoef(y;\cvCoefParams)$ from Monte Carlo estimates such that it minimizes variance.

Since both $\cvCoef$ and $\CV$ are mere scaling factors of $\cv$, one could model the product $\cvCoef \cdot \CV$ directly.
We choose to keep them separate as this enables approximating $F$ by (unweighted) $\CV$ without evaluating the residual integral. We exercise this option in a biased version of our light-transport estimator whenever the approximation error is heuristically determined to be low; details follow in in \autoref{sec:path-truncation}.

\section{Monte Carlo Integration with NCV}%
\label{sec:mc-integration}

As an evolution of the parametric integral in \autoref{eq:parametric-cv-equation-with-conditionals}, our parametric trainable control variate yields
\begin{align}
    F(y) &= \cvCoef(y; \cvCoefParams) \cdot \CV(y; \CVParams) \nonumber \\
    &\qquad+ \int_\Domain f(x, y) -  \cvCoef(y; \cvCoefParams) \cdot \cv(x, y;\cvParams) \Diff{x}\,,
    \label{eq:parametric-cv-equation}
\end{align}
where the various $\theta$ denote the corresponding model parameter sets.
For the sake of readability, we omit the dependency on $y$ in the following derivations and
define the shorthands $\acv$ and $\aCV$ that represent the $\cvCoef$-weighted CV and its corresponding integral:\@
\begin{align}
    \aCV(\aCVParams) &:=\cvCoef(\cvCoefParams) \cdot \CV(\CVParams) \,;& \aCVParams &:=\cvCoefParams \cup \CVParams \,,
    \\
    \acv(x;\acvParams) &:=\cvCoef(\cvCoefParams) \cdot \cv(x;\cvParams) \,;& \acvParams &:=\cvCoefParams \cup \CVParams \cup \cvShapeParams \,.
\end{align}
Applying these notational simplifications, a one-sample Monte Carlo estimator of \autoref{eq:parametric-cv-equation} amounts to
\begin{align}
    \MC{F} = \aCV(\aCVParams) + \frac{f(X) - \acv(X;\acvParams)}{\PdfMC(X;\pdfParams)} \,,
    \label{eq:mc-estimator}
\end{align}
where $\PdfMC(X;\pdfParams) \equiv \PdfMC(X, y;\pdfParams)$ is the parametric probability density of drawing sample $X$.

\begin{figure*}
    \setlength{\tabcolsep}{0.6pt}%
\renewcommand{\arraystretch}{1}%
\scriptsize%
\hspace*{-3mm}%
\begin{tabular}{rc@{\hskip 7pt}ccc@{\hskip 7pt}ccc@{\hskip 7pt}ccc@{\hskip 7pt}c}
{} & {\textbf{(a)} NIS} & \multicolumn{3}{c}{\textbf{(b)} NCV---monochromatic (1 flow)} & \multicolumn{3}{c}{\textbf{(c)} NCV---spectral (3 flows)} & \multicolumn{3}{c}{\textbf{(d)} NCV---spectral (1 multi-channel flow)} & \textbf{(e)} Reference \\
\cmidrule(lr{12pt}){2-2} \cmidrule(lr{12pt}){3-5} \cmidrule(lr{12pt}){6-8} \cmidrule(lr{12pt}){9-11} \cmidrule(lr){12-12}{} & $\PdfMC \approxprop f$ & $\cv \approx f$ & $|f - \cv|$ & $\PdfMC \approxprop |f - \cv|$ & $\cv \approx f$ & $|f - \cv|$ & $\PdfMC \approxprop |f - \cv|$ & $\cv \approx f$ & $|f - \cv|$ & $\PdfMC \approxprop |f - \cv|$ & Integrand $f$ \\
\rotatebox{90}{\hspace{0.62cm}Kandinsky}
& \includegraphics[width=0.084\textwidth,trim={50 0 50 0},clip]{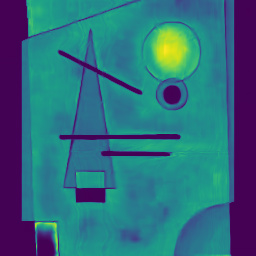}
& \includegraphics[width=0.084\textwidth,trim={50 0 50 0},clip]{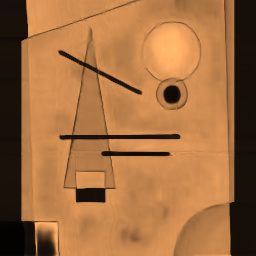}
& \includegraphics[width=0.084\textwidth,trim={50 0 50 0},clip]{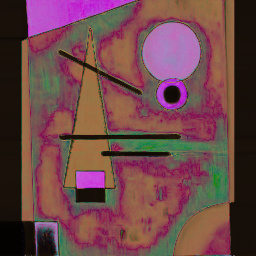}
& \includegraphics[width=0.084\textwidth,trim={50 0 50 0},clip]{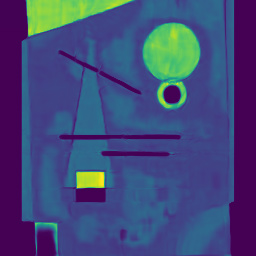}
& \includegraphics[width=0.084\textwidth,trim={50 0 50 0},clip]{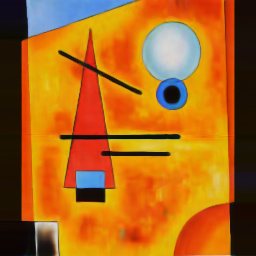}
& \includegraphics[width=0.084\textwidth,trim={50 0 50 0},clip]{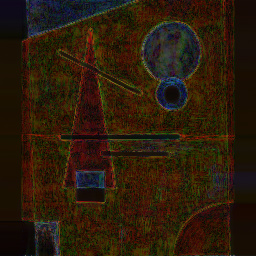}
& \includegraphics[width=0.084\textwidth,trim={50 0 50 0},clip]{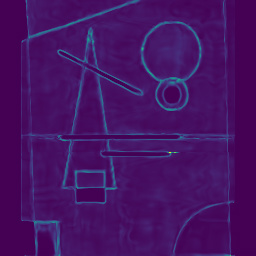}
& \includegraphics[width=0.084\textwidth,trim={50 0 50 0},clip]{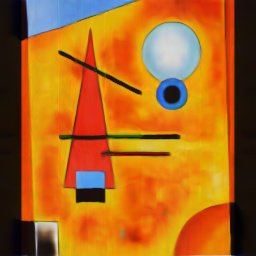}
& \includegraphics[width=0.084\textwidth,trim={50 0 50 0},clip]{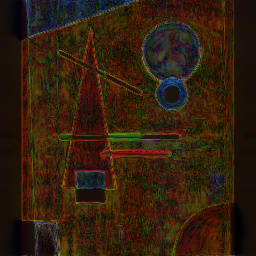}
& \includegraphics[width=0.084\textwidth,trim={50 0 50 0},clip]{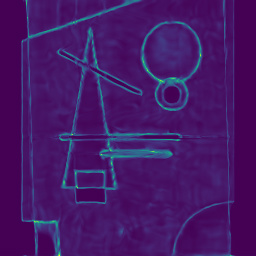}
& \includegraphics[width=0.084\textwidth,trim={50 0 50 0},clip]{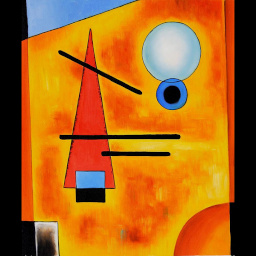}
\\[-0.5mm]
{} & Efficiency: 0.013 & {} & Efficiency: 0.010 & & {} & Efficiency: 0.216 & & {} & Efficiency: 0.228 & & {} \\[0.5mm]
\rotatebox{90}{\hspace{0.46cm}Feline Predator}
& \includegraphics[width=0.084\textwidth,trim={50 0 50 0},clip]{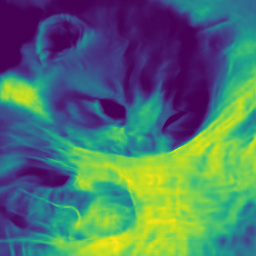}
& \includegraphics[width=0.084\textwidth,trim={50 0 50 0},clip]{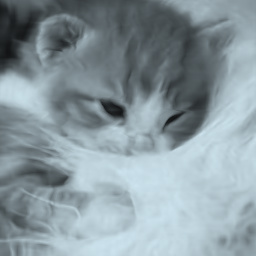}
& \includegraphics[width=0.084\textwidth,trim={50 0 50 0},clip]{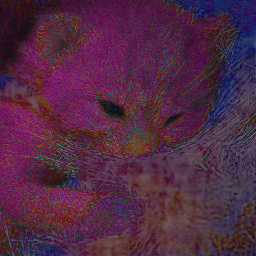}
& \includegraphics[width=0.084\textwidth,trim={50 0 50 0},clip]{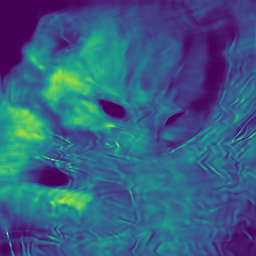}
& \includegraphics[width=0.084\textwidth,trim={50 0 50 0},clip]{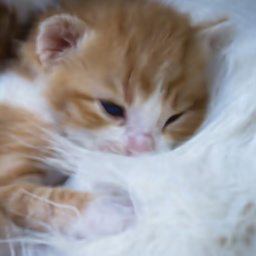}
& \includegraphics[width=0.084\textwidth,trim={50 0 50 0},clip]{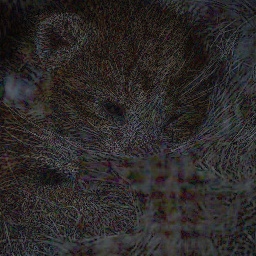}
& \includegraphics[width=0.084\textwidth,trim={50 0 50 0},clip]{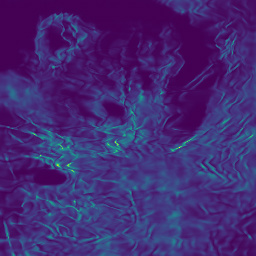}
& \includegraphics[width=0.084\textwidth,trim={50 0 50 0},clip]{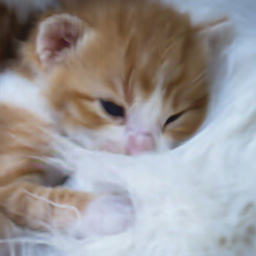}
& \includegraphics[width=0.084\textwidth,trim={50 0 50 0},clip]{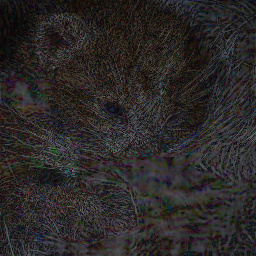}
& \includegraphics[width=0.084\textwidth,trim={50 0 50 0},clip]{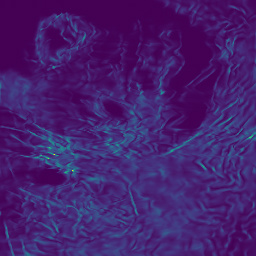}
& \includegraphics[width=0.084\textwidth,trim={50 0 50 0},clip]{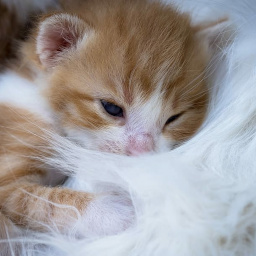}
\\[-0.5mm]
{} & Efficiency: 0.059 & & Efficiency: 0.084 & & & Efficiency: 0.242 & & & Efficiency: 0.312 & & {} \\[0.5mm]
\end{tabular}

    \vspace{-2mm}
    \caption{\label{fig:toy}
        Comparison of neural importance sampling \textbf{(a)}~\citep{mueller2019nis} and various flavors of our neural control variates \textbf{(b, c, d)} on two toy integration problems (rows). The integrands are 2D images with 3 color channels \textbf{(e)}; the goal is to estimate the average color of the images.
        We report the Monte Carlo efficiency, defined as ${\big(\Variance[\MC{F}] \cdot \textbf{runtime}\big)}^{-1}$, of using the different techniques after each technique's training converged. We also visualize the functions learned during the MC estimation, i.e.\ the sampling PDF $\PdfMC$ and the control variate $\cv$.
        NIS \textbf{(a)} is the least efficient method, because importance sampling can only target a scalar quantity---in this case the average of the 3 channels.
        Applying our NCVs, even with a single monochromatic flow \textbf{(b)}, improves efficiency, because the learned CV $\cv$ is able to match the average color of the integrand. The learned sampling PDF $\PdfMC$ therefore only needs to focus on the remaining color variation in the residual difference $|f - \cv|$.
        Using three independent flows \textbf{(c)} and using one of our multi-channel flows \textbf{(d)} for the CV $\cv$ both achieve great additional efficiency gains, because they can model color variation.
        The residual difference $|f - \cv|$ is thus much smaller and the sampling PDF $\PdfMC$ focuses on the remaining approximation error, which consists of sharp edges in the integrand.
        Our multi-channel flow yields the best MC efficiency: the fit is only slightly worse than with three flows but the evaluation is much faster.
    }
\end{figure*}

\paragraph{Importance sampling of the residual integral}%

To motivate the need for a parametric PDF model, we note that the variance of \autoref{eq:mc-estimator} is minimized when the PDF is proportional to the absolute correction term ${|f(X) - \acv(X;\acvParams)|}$.
Since the CV will be optimized progressively, the correction term will evolve over time.
In the ideal case, the absolute difference ${|f(X) - \acv(X;\acvParams)|}$ would get uniformly smaller and the optimal sampling distribution would be uniform, i.e.\ constant.
However, our experiments showed that despite the approximation power of neural networks, the numerator is never sufficiently uniformly bounded to permit a uniform PDF $\PdfUniform(x)$ to perform well in practice.

Accounting for the progressive optimization and the limited expressivity of the CV, we propose a sampling PDF that combines two samplers: a defensive sampler (in the following: uniform) that bootstraps the initial Monte Carlo estimates, and a learned parametric sampler that can capture the shape of the numerator once the CV has converged.
We combine these two sampling distributions using multiple importance sampling (MIS)~\citep{Veach:1995:MIS}
with learned probabilities for selecting the PDFs~\citep{mueller2019nis}.

We target the neural importance sampling probability density $\PdfNIS$ \citep{mueller2019nis} at the difference in the numerator of \autoref{eq:mc-estimator}.
To probabilistically select between uniform sampling $\PdfUniform$ and $\PdfNIS$
we train a parametric neural network $\selectionProb(x;\selectionProbParams)$ that approximates the variance-optimal selection probabilities of $\PdfNIS$.
We closely follow the approach by \citet{mueller2019nis} (including the prevention of degenerate training by the $\beta$ parameter)
optimizing $\selectionProb(x;\selectionProbParams)$ concurrently with the CV and PDF models to strike a good balance between uniform and neural importance sampling at any time during the training process.
The final PDF reads:
\begin{align}
    \PdfMC(x;\pdfParams) &=
    \big(1 - \selectionProb(x;\selectionProbParams)\big) \, \PdfUniform(x)
    + \selectionProb(x;\selectionProbParams) \, \PdfNIS(x;\nisParams) \,,
    \label{eq:sampling-pdf-blend-uniform}
\end{align}
where ${\pdfParams := \selectionProbParams \cup \nisParams}$.

\paragraph{Spectral 2D example}
We demonstrate the efficiency benefits of using our neural control variates for variance reduction in \autoref{fig:toy}.
We compare neural importance sampling~\citep{mueller2019nis} alone to three flavors of our full estimator from \autoref{eq:mc-estimator}: (i) a monochromatic, single-channel flow, (ii) multiple independent flows (one per channel), and (iii) the proposed multi-channel flow.
The multi-channel flow consistently achieves the highest efficiency, while learning only slightly worse control variates than multiple independent flows.
Note how the sampling PDF focuses on the high-frequency detail that our control variates do not perfectly capture.

\section{Optimization}
\label{sec:optimization}
In this section, we derive a theoretically optimal, variance-minimizing loss for optimizing the parameters of the CV and the sampling distribution. We then propose an empirical, composite loss that provided better performance and stable optimization in our experiments.

\subsection{Minimizing Variance by Optimization}
\label{sec:joint_loss}
Our goal is to minimize the variance of the CV estimator by training the neural networks using a convergent gradient-based optimizer.
Stochastic gradient descent provably converges to local optima when driven by \emph{unbiased estimates} of the loss gradient.\footnote{For a formal proof of convergence, the learning rate must approach zero at a carefully chosen rate, leading to an impractically slow optimization. Leaving the learning rate high, the optimization fluctuates around local minima, which is a widely accepted limitation in machine learning literature.}
In this section, we first derive the variance formula and then show that unbiased gradient estimates thereof can be computed using auto-differentiation.

We use the variance
\begin{align}
    \Variance\left[\MC{F}\right] &= \Expectation\left[ {\MC{F}}^2 \right] - {\Expectation\left[ \MC{F} \right]}^2 \nonumber \\
    &= \int_\Domain \frac{{\left(f(x) - \acv(x;\acvParams)\right)}^2}{\PdfMC(x;\pdfParams)} \Diff{x} - {\left(F - \aCV(\aCVParams)\right)}^2
    \label{eq:mc-variance}
\end{align}
of the estimator in \autoref{eq:mc-estimator} as the loss function.

\paragraph{Interpretation}

Minimizing the first term of \autoref{eq:mc-variance} corresponds to fitting $\acv$ to $f$ in terms of weighted least squares, where the weights are the inverse sampling density.
The weighted-least-squares distance is minimized when $\acv(x) = f(x)$, leading to zero variance.
Interestingly, the variance is also zero when the non-zero first term equals to the second term.
Due to this additional degree of freedom, there exists an entire family of CVs that yield zero variance.
A classical example of such a configuration is a control variate that matches $f$ up to an additive constant, $\acv(x) = f(x) + c$ for $c\in\R$, and a uniform $\PdfMC(x)$.

\paragraph{Variance with noisy estimates of \texorpdfstring{$f(x)$}{the integrand}}

In many applications, the original integrand $f(x)$ cannot be evaluated \emph{analytically}.
One such application is investigated in \autoref{sec:light-transport}, where we apply control variates to light transport simulation governed by a Fredholm integral equation.

Generalizing \autoref{eq:mc-variance}, we now demonstrate that noisy estimates of $f(x)$ pose no problem for the convergence of the optimizer.
Using the generic notation $f(x) := \int_\DomainZ f(x,z) \Diff{z}$ and
inserting it into the integral in \autoref{eq:parametric-cv-equation} (with $\conditionals$ being omitted for brevity as mentioned before), we obtain
\begin{align}
    F =\aCV(\aCVParams) + \int_\Domain \int_\DomainZ f(x,z) \Diff{z} - \acv(x;\acvParams) \Diff{x}\,.
    \label{eq:parametric-cv-equation-with-z}
\end{align}
A one-sample Monte Carlo estimator that leverages a single $(X,Z)$ sample to approximate $F$ reads
\begin{align}
    \MC{F} = \aCV(\aCVParams) + \frac{f(X,Z)}{\PdfMC(X,Z;\pdfParams)}  - \frac{\acv(X;\acvParams)}{\PdfMC(X;\pdfParams)} \,,
    \label{eq:mc-estimator-with-z}
\end{align}
where $\PdfMC(X,Z;\pdfParams) = \PdfMC(X;\pdfParams) \cdot \PdfMC(Z|X)$ is the joint probability density of sampling $X$ and $Z$, and $\PdfMC(X;\pdfParams)$ and $\PdfMC(Z|X)$ are the marginal and conditional densities, respectively.

The variance of the estimator in \autoref{eq:mc-estimator-with-z} can be derived in analogy to the variance of the estimator in \autoref{eq:mc-estimator}:
\begin{align}
\Variance\left[\MC{F}\right]
&= \!\int_\Domain \!\int_\DomainZ {\left( \frac{f(x,z)}{\PdfMC(z|x)} - \acv(x;\acvParams)\right)}^2\frac{\PdfMC(z|x)}{\PdfMC(x;\pdfParams)} \Diff{z}\!\Diff{x} \nonumber\\
& \qquad - {\big(F - \aCV(\aCVParams)\big)}^2;
\label{eq:mc-variance-with-tail}
\end{align}
see \autoref{sec:variance-derivation} for a complete derivation.

Finding optimal $\acvParams$, $\aCVParams$, $\pdfParams$ that minimize \autoref{eq:mc-variance-with-tail} in closed form is not practical as the equation contains the unknown integral $F$, which we are trying to compute in the first place, and a double integral, which for meaningful settings in computer graphics is infeasible to solve analytically.
Therefore, we resort to stochastic gradient-based optimizers that converge to the correct solution even if the loss is only approximated; provided that its approximation is unbiased.

\paragraph{Taking advantage of autograd functionality}

Using Leibniz's integral rule, we can swap the order of differentiation and MC estimation of variance: first estimate variance and then rely on auto-differentiation in modern optimization tools to compute the gradients.
Using Monte Carlo, the variance in \autoref{eq:mc-variance-with-tail} can be estimated using the following unbiased one-sample estimator:
\begin{align}
    \MC{\Variance\left[\MC{F}\right]}
    &=
    \frac{
        {\left(\frac{f(X,Z)}{\PdfMC(Z|X)} - \acv(X;\acvParams)\right)}^2 \PdfMC(Z|X)
    }{
        \PdfMC(X;\pdfParams) \, \PdfVarEst(X) \, \PdfVarEst(Z|X)
    } \nonumber \\
    & \qquad - {\left(\frac{  \MC{f(X)}  }{\PdfVarEst(X)} - \aCV(\aCVParams)\right)}^2 \,,
\end{align}
where $\PdfVarEst$ is the density of samples used for estimating the variance.
The estimator can be further simplified assuming that we use the same conditional densities in $\MC{F}$ and $\MC{\Variance}$, i.e.\ $\PdfMC(z|x) = \PdfVarEst(z|x)$,
and interpreting the fraction $\frac{f(X,Z)}{\PdfMC(Z|X)}$ as a one-sample estimator of $f(X)$:
\renewcommand{\fboxsep}{4pt}
\begin{empheq}[box=\fbox]{equation}
\begin{split}
    \MC{\Variance\left[\MC{F}\right]}
    &=
    \frac{
        {\big(\MC{f(X)} - \cvCoef(\cvCoefParams) \cv(X;\cvParams)\big)}^2
    }{
        \PdfMC(X;\pdfParams) \, \PdfVarEst(X)
    } \\
    & \qquad 
    - {\left( \frac{\MC{f(X)}  }{\PdfVarEst(X)} - \cvCoef(\cvCoefParams) \CV(\CVParams) \right)}^2,
    \label{eq:variance-estimator}
\end{split}
\end{empheq}
\renewcommand{\fboxsep}{0pt}%
where the symbols with hats were replaced by their definitions.

The variance estimate in \autoref{eq:variance-estimator} can be used as the loss function in modern optimization tools based on autograd.
Unfortunately, despite being theoretically optimal, our empirical analysis revealed poor overall performance and unstable optimization when using this loss.

\subsection{Composite Loss for Stable Optimization}%
\label{sec:composite-loss}

The variance of the parametric estimator, \autoref{eq:mc-variance-with-tail}, can be zero for an entire family of configurations of $\cvCoefParams, \cvParams, \CVParams$, and $\pdfParams$.
However, taking into account the entire \autoref{eq:mc-variance-with-tail} for each of the trainable components led to erratic optimization and often failed to approach one of the zero-variance configurations in our experiments.

We thus propose a composite loss that is more robust in the presence of noisy loss estimates.
Our composite loss imposes restrictions as it is zero \emph{only} for the following zero-variance configuration:
\begin{align}
    \CV(\CVParams) &= F\,, \label{eq:cvIntegral-constraint} \\
    \cvShape(x;\cvShapeParams) &= \frac{f(x)}{F}\,, \label{eq:cvShape-constraint} \\
    \PdfMC(x;\pdfParams) &= \frac{|f(x) - \cv(x; \cvParams)|}{\int_\Domain |f(x) - \cv(x; \cvParams)| \Diff{x}}\,, \text{ and}
    \label{eq:composite-optimum-pdf}
    \\
    \cvCoef(\cvCoefParams) &= 1\,. \label{eq:composite-optimum-cvCoef}
\end{align}
Despite being more restrictive, decomposing the optimization into smaller, better-understood optimization tasks leads to better results in practice than blindly relying on \autoref{eq:mc-variance-with-tail}.
Our composite loss is the sum of the individual terms:
\begin{align}
    \Loss =
    \underbrace{\Loss^2 \left(F, \CV;\CVParams\right)}_{\text{CV integral}} +
    \underbrace{\Loss_\CrossEntropy \left(\bar{f}, \cvShape;\cvShapeParams\right)}_{\text{CV shape}} +
    \underbrace{\Loss_\CrossEntropy \left(|f-g|, \PdfMC; \pdfParams\right)}_{\text{Sampling PDF}} +
    \underbrace{\Loss_{\Variance}(\cvCoefParams)}_{\text{$\cvCoef$-coefficient}},
    \label{eq:composite-loss}
\end{align}
which we detail in the following paragraphs.

\begin{figure*}
    \setlength{\fboxrule}{10pt}%
\setlength{\insetvsep}{20pt}%
\setlength{\tabcolsep}{-1pt}%
\renewcommand{\arraystretch}{1}%
\small%
\hspace*{-2mm}%
\begin{tabular}{rccccc}
  {} & {} & Variance loss & $\Loss^2$ loss & Relative $\Loss^2$ loss & Reference \\
    \setInset{A}{red}{670}{50}{285}{114}%
    \setInset{B}{orange}{650}{875}{285}{114}%
    \rotatebox{90}{\hspace{-1.70cm}\Bedroom{}}\hspace{0.14cm} & 
    \addBeautyCrop{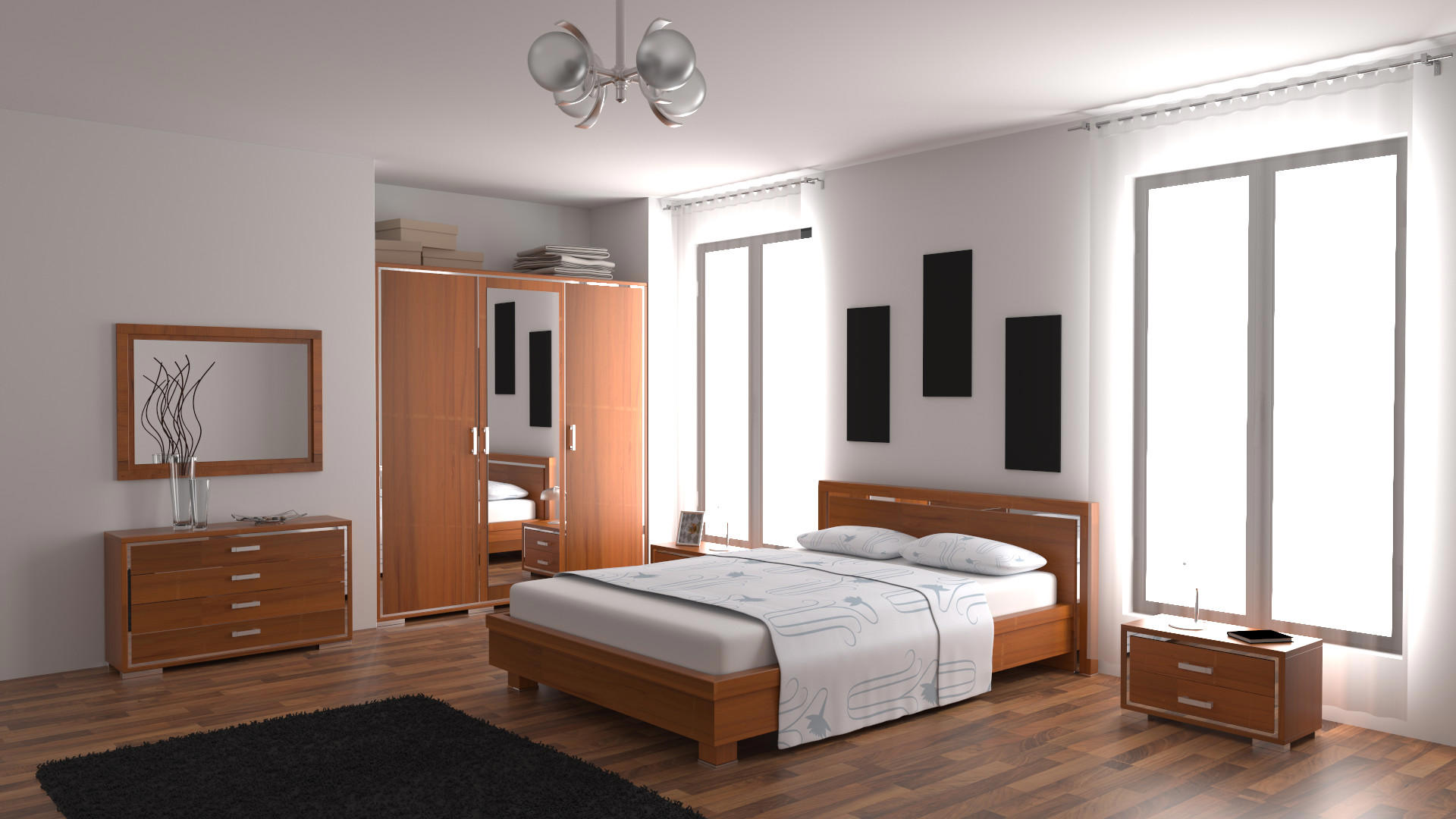}{0.27}{1920}{1080}{0}{0}{1920}{1080} &
    \addInsets{images/fig-scale-loss/bedroom-Variance-loss.jpg} &
    \addInsets{images/fig-scale-loss/bedroom-L2-loss.jpg} &
    \addInsets{images/fig-scale-loss/bedroom-Relative-L2-loss.jpg} &
    \addInsets{images/fig-references/bedroom-reference.jpg} \\
  & \multicolumn{1}{r}{MAPE:} & 0.425 & 0.162 & \textbf{0.042} \\
\end{tabular}

    \vspace{-1mm}
    \caption{\label{fig:scale}
        Control Variate integral optimization. Using the estimator variance loss (left) to optimize the CV integral $\CV$ effectively merges it with the coefficient $\alpha$, resulting in a darker output where the CV shape is a poor match for the target.
        Using the $\Loss^2$ loss (middle) decouples the CV integral and the coefficient $\alpha$.
        The relative $\Loss^2$ loss (right) further improves the model prediction in dark regions, such as the floor under the bed.
    }%
    \label{fig:scale-loss}
\end{figure*}

\begin{figure*}
    \setlength{\fboxrule}{10pt}%
\setlength{\insetvsep}{20pt}%
\setlength{\tabcolsep}{-1pt}%
\renewcommand{\arraystretch}{1}%
\small%
\hspace*{-2mm}%
\begin{tabular}{rccccccc}
  {} & {} &      Variance      & + rel. $\Loss^2$ & + cross entropy & + rel. variance & + cross entropy \\
  {} & {} & for all components &  for CV integral  &  for CV shape   &  for CV alpha   &     for IS      & Reference \\
    \setInset{A}{red}{170}{500}{60}{38}%
    \setInset{B}{orange}{965}{975}{60}{38}%
    \rotatebox{90}{\hspace{-1.80cm}\Bathroom{}}\hspace{0.14cm} & 
    \addBeautyCrop{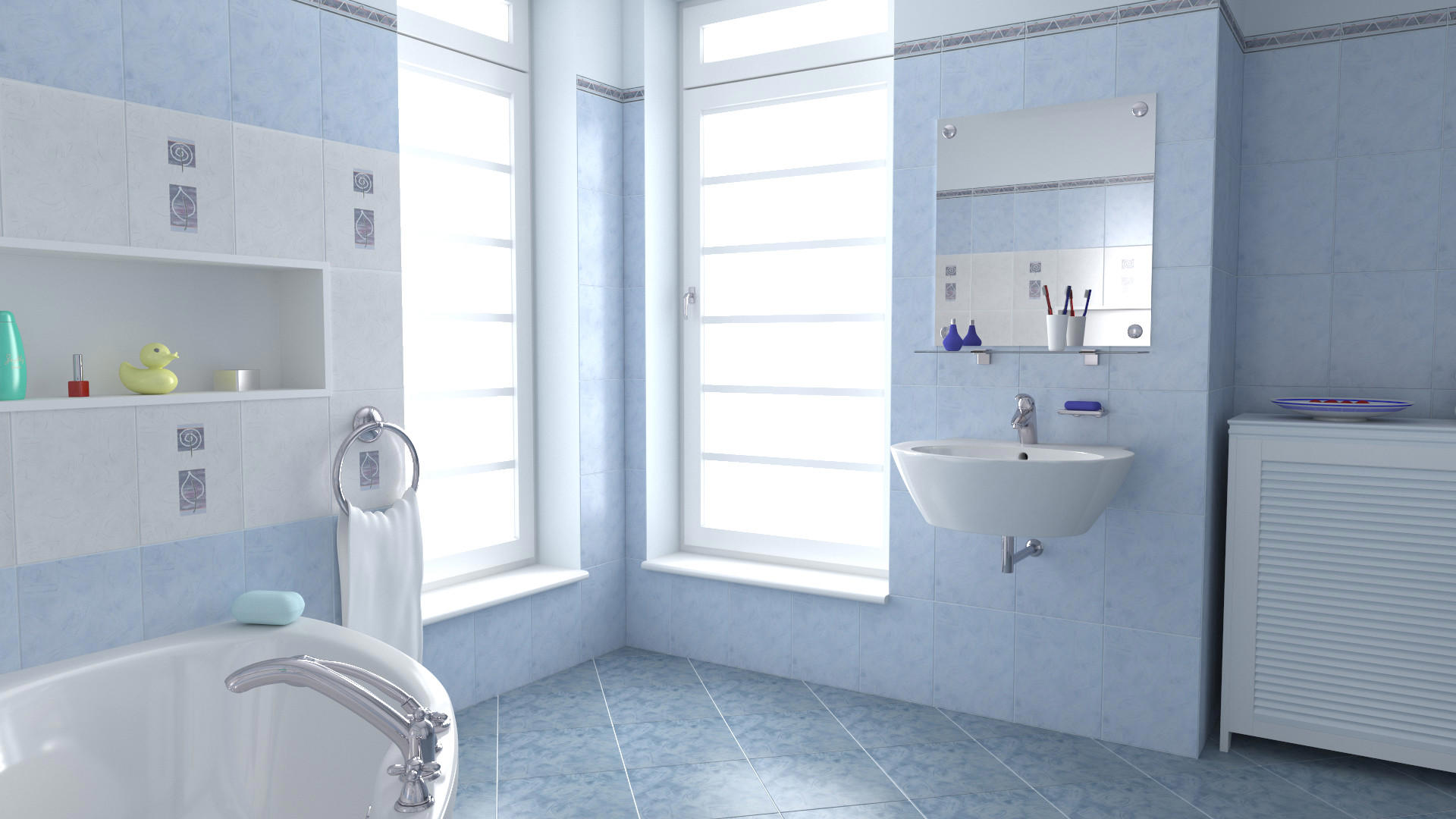}{0.27}{1920}{1080}{0}{0}{1920}{1080} &
    \addInsets{images/fig-ablation/bathroom-Variance-loss.jpg} &
    \addInsets{images/fig-ablation/bathroom-CV-integral-loss.jpg} &
    \addInsets{images/fig-ablation/bathroom-CV-shape-loss.jpg} &
    \addInsets{images/fig-ablation/bathroom-CV-alpha-loss.jpg} &
    \addInsets{images/fig-ablation/bathroom-IS-loss.jpg} &
    \addInsets{images/fig-references/bathroom-reference.jpg} \\
  & \multicolumn{1}{r}{MAPE:} & 0.058 & 0.056 & 0.053 & 0.053 & \textbf{0.046} \\
\end{tabular}

    \vspace{-1mm}
    \caption{\label{fig:ablation}
        Impact of the individual composite-loss terms at equal sample count (512 spp).
        First, we jointly optimize all components using the variance loss described in~\autoref{sec:joint_loss}.
        Then, we progressively replace each variance term by the corresponding term from the composite loss described in~\autoref{sec:composite-loss}.
        Optimizing the CV integral using the relative $\Loss^2$ loss has a minor impact on the final reconstruction,
        \ADD{
        but it greatly improves the visual accuracy of the CV integral (as shown in \autoref{fig:scale-loss}), which in turn enables its use in our biased reconstruction.}
        \ADD{The accurate CV integral is also a key enabler of the cross entropy loss that we use to optimize the CV shape and IS component, which proved crucial for the overall robustness and efficiency of our algorithm.}
        Aside of the numerical and visual improvements, the main benefit of our composite loss lies in its robustness: rendering the \Bathroom{} scene using our proposed composite loss consistently converged to a satisfying result, whereas using the variance loss frequently resulted in diverging optimizations.
    }%
\end{figure*}

\paragraph{CV integral optimization}

To satisfy the constraint in \autoref{eq:cvIntegral-constraint}, we minimize a relative $\Loss^2$ metric
\begin{align}
    \Loss^2 \left(F, \CV;\CVParams\right) = \frac{{\big(F - \CV(\CVParams)\big)}^2}{{\StopGradient(\CV(\CVParams))}^2+\epsilon} \,,
    \label{eq:relative-l2}
\end{align}
where $\StopGradient(x)$ indicates that $x$ is treated as a constant, i.e.\ no gradients w.r.t.\ it are computed, and $\epsilon=\num{1e-2}$.
Our choice of a relative $\Loss^2$ metric has two reasons: first, the $\Loss^2$ metric admits unbiased gradient estimates when $F$ is noisy, and second, relative losses are robust with respect to a high dynamic range of values.
We use ${\CV(\CVParams)}^2$ as the normalization constant, as proposed by \citet{Lehtinen:2018}, because normalizing by $F^2$~\citep{Rousselle:2011} is infeasible---our goal is to estimate $F$ in the first place.
${\CV(\CVParams)}^2$ merely serving as an approximation of $F^2$ in the denominator is the reason why it must be treated as a constant for the optimization to be correct---hence the $\StopGradient(\,\cdot\,)$ around it.
It follows that our Monte Carlo estimator of $\Loss^2 \left(F, \CV;\CVParams\right)$, which we feed to automatic differentiation, reads
\begin{align}
    \MC{\Loss^2 \left(F, \CV;\CVParams\right)} = \frac{{\big(\MC{F} - \CV(\CVParams)\big)}^2}{{\StopGradient(\CV(\CVParams))}^2+\epsilon} \,.
\end{align}
In \autoref{fig:scale}, we illustrate the learned integral when optimizing either the variance, $\Loss^2$, or relative $\Loss^2$ in the setting of light-transport simulation as explored in \autoref{sec:light-transport}.
The relative $\Loss^2$ loss achieves the most accurate fit.

\paragraph{CV shape optimization}

The CV shape is modeled using a normalizing flow, the parameters of which are optimized using the cross entropy.
The cross entropy measures the similarity between two normalized functions and yields more robust convergence than minimizing variance directly~\cite{mueller2019nis}.
Since we aim to satisfy the constraint in \autoref{eq:cvShape-constraint}, we minimize the cross entropy of the normalized integrand, $\bar{f}(x) = f(x)/F$, to the shape of the CV, $\cvShape$:
\begin{align}
    \Loss_\CrossEntropy \left(\bar{f}, \cvShape;\cvShapeParams\right) = -\int_\Domain \bar{f}(x) \log \left(\cvShape(x;\cvShapeParams)\right) \Diff{x}\,.
\end{align}
The main caveat of the cross entropy is that it requires normalizing the integrand.
\ADD{Despite $F$ not being known exactly, approximate normalization \ $\bar{f}(x) \approx f(x)/\CV(\CVParams)$ is feasible by using the learned CV integral $\CV(\CVParams)$ instead of the exact normalization constant.}
With this observation, \ADD{an approximately normalized} MC estimator of the cross entropy that can be fed to automatic differentiation reads
\begin{align}
    \MC{\Loss_\CrossEntropy \left(\bar{f}, \cvShape;\cvShapeParams\right)} = -\frac{\MC{f(X)}}{\StopGradient(\CV(\CVParams))+\epsilon} \frac{\log \left(\cvShape(X;\cvShapeParams)\right)}{\PdfVarEst(X)} \,.
\end{align}

\paragraph{Sampling distribution optimization}
Our parametric sampling distribution $\PdfMC(x;\pdfParams)$ consists of a normalizing flow $\PdfNIS(x;\nisParams)$ as well as the selection probability $\selectionProb(x;\selectionProbParams)$, which are both optimized using the cross entropy---the same as in neural importance sampling (NIS)~\citep{mueller2019nis}.
However, in contrast to NIS, which optimizes the flow to match the normalized integrand $\bar{f}(x)$,
we optimize the flow to approximate the normalized absolute difference in \autoref{eq:composite-optimum-pdf}.
\ADD{Once again, we approximate the normalization constant by $\CV(\CVParams)$.}
In addition, we approximate the difference ${|f(x) - \cv(x; \cvParams)|}$ using the biased estimator \[{\Delta_{f,\cv}(X) = |\MC{f(X)}-\StopGradient(\cv(X;\cvParams))|} \, ,\] resulting in the following cross-entropy estimator for automatic differentiation:
\begin{align}
    \MC{\Loss_\CrossEntropy \left(|f-g|, \PdfMC; \pdfParams\right)} = -\frac{\Delta_{f,\cv}(X)}{\StopGradient(\CV(\CVParams))+\epsilon}
    \frac{\log \left(\PdfMC(X;\pdfParams)\right)}{\PdfVarEst(X)} \,.
    \label{eq:pdf-cross-entropy}
\end{align}
Note that $\Delta_{f,\cv}(X)$ is biased due to Jensen's inequality: taking the absolute value of an estimator \emph{overestimates} the absolute value of the estimator's expectation.
As a result, the above cross-entropy estimator is an \emph{upper bound} to the true cross entropy between $|f-\cv|$ and $\PdfMC$.
Crucially, since the upper bound has the same minimum as the cross entropy (when the flow matches the normalized absolute difference) minimizing the upper bound does not prevent convergence and worked sufficiently well in our experiments.

\paragraph{$\cvCoef$-coefficient optimization}

As given by the constraint in \autoref{eq:composite-optimum-cvCoef}, we only achieve zero variance using ${\cvCoef = 1}$.
However, this identity assumes that our parametric control variate and sampling distribution \emph{exactly} match their targets, which is unlikely in practice.
In such cases, the $\cvCoef$-coefficient allows for downweighting the control variate to avoid increased variance due to a poor fit.
We therefore employ a parametric model for $\cvCoef$, too, and optimize it to minimize the \emph{relative} variance of the complete CV estimator:
\begin{align}
    \Loss_{\Variance}(\cvCoefParams) = \frac{\Variance\left[\MC{F}\right]}{{\StopGradient(\CV(\CVParams))}^2+\epsilon} \,,
\end{align}
where we use a relative loss for the same reason as in \autoref{eq:relative-l2}: to be robust with respect to a high dynamic range of values.
The $\cvCoef$ coefficient is thus the only component of our model that is optimized with respect to the variance loss in \autoref{eq:mc-variance-with-tail}; we use the estimator in \autoref{eq:variance-estimator} to estimate the numerator of $\Loss_{\Variance}(\cvCoefParams)$ for optimizing~$\cvCoefParams$:
\begin{align}
    \MC{\Loss_{\Variance}(\cvCoefParams)} = \frac{\MC{\Variance\left[\MC{F}\right]}}{{\StopGradient(\CV(\CVParams))}^2+\epsilon} \,,
\end{align}
In \autoref{fig:ablation}, we demonstrate the additional robustness of using our composite loss instead of the theoretically optimal variance loss.
We note that the variance loss result, as well as the intermediate ablation results, are generally very unstable; we had to run many optimization runs to produce these results.
In contrast, the final composite loss consistently produces useful results.

\newlength{\oldparskip}
\setlength{\oldparskip}{\parskip}
\setlength{\parskip}{\baselineskip}

\setlength{\parskip}{\oldparskip}

\begin{figure*}
    \input{all-components-figure\useAnimations}
    \vspace{-2mm}
    \caption{\label{fig:all-components}
        All learned components of our method when applied to light transport simulation:
        we visualize the learned CV integral $\CV(\pos, \diro; \CVParams)$, the coefficient $\cvCoef(\pos, \diro; \cvCoefParams)$, and selection probability $\selectionProb(\pos, \diro; \selectionProbParams)$ at the primary vertices (first non-delta interaction) of each pixel.
        Furthermore, we show the directionally resolved learned CV $\cv(\pos, \diro, \diri; \cvShapeParams)$ and PDF $\PdfMC(\pos, \diro, \diri; \pdfParams)$ at the spatial location marked in red.
        The CV integral approximates the scattered light field $\reflectedRadiance(\pos, \diro)$ remarkably well.
        In places where either the CV integral or the shape is inaccurate, the learned alpha-coefficient weighs down the contribution of the CV to our unbiased estimator.
        Lastly, the learned selection probability blends between BSDF sampling (red) and residual neural importance sampling (green) such that variance is minimized.
        Note how glossy surfaces tend to favor BSDF sampling, whereas rougher surfaces often favor residual NIS.\@
    }
\end{figure*}

\section{Application to Light Transport}%
\label{sec:light-transport}

With trainable control variates at hand, we are ready to demonstrate their benefits in light-transport simulation.
Physically based image synthesis is concerned with estimating the scattered radiance
\begin{align}
    \reflectedRadiance(\pos,\diro) = \int_{\Sphere} \bsdf(\pos,\diro,\diri) \inRadiance(\pos,\diri) \left| \cos \fsangle \right| \Diff{\diri}
    \label{eq:reflected-radiance}
\end{align}
that leaves surface point $\pos$ in direction $\diro$~\citep{PBRT},
where $\bsdf$ is the bidirectional scattering distribution function, $\inRadiance$ is radiance arriving at $\pos$ from direction $\diri$, and $\fsangle$ is the foreshortening angle.

The correspondence to \autoref{eq:parametric-cv-equation} is established as follows: the scattered radiance $\reflectedRadiance(\pos,\diro)$ corresponds to the parametric integral $F(y)$, where\ $y \equiv (\pos,\diro)$, which we refer to as the \emph{query location}.
The integration domain and the integration variable are the unit sphere and the direction of incidence, i.e.\ ${\Domain \equiv \Sphere}$ and ${x \equiv \diri}$, respectively.

Our goal is to reduce estimation variance by leveraging the parametric CV from~\autoref{sec:model}.
Its integral component serves as an approximation of the scattered radiance, i.e.\ $\CV(\pos,\diro; \CVParams) \approx \reflectedRadiance(\pos,\diro)$, while its shape component $\cvShape(\pos,\diro,\diri;\cvShapeParams)$ approximates the normalized integrand.
In analogy to \autoref{eq:mc-estimator}, a one-sample MC estimator of \autoref{eq:reflected-radiance} with the trainable CV from \autoref{sec:model} reads:
\begin{align}
    \MC{\reflectedRadiance(\pos,\diro)} &= \aCV(\pos,\diro;\CVParams) \nonumber \\
    &\qquad + \frac{\bsdf(\pos,\diro,\diriRV) \inRadiance(\pos,\diriRV) \left| \cos \fsangle \right| - \acv(\pos,\diro,\diriRV;\cvShapeParams) }
    {\PdfMC(\diriRV| \pos,\diro;\pdfParams)} \,.
    \label{eq:reflected-radiance-cv-estimator}
\end{align}
We made one small modification to $\PdfMC(\diriRV| \pos,\diro;\pdfParams)$: instead of mixing NIS with uniform sampling as proposed
in \autoref{sec:mc-integration},
we mix NIS with BSDF sampling $\PdfBSDF$, which in rendering in many cases is a better baseline than uniform sampling.
This results in the following PDF:\@
\begin{align}
    \PdfMC(\diriRV|\pos,\diro;\pdfParams) &=
    \big(1 - \selectionProb(\pos,\diro;\selectionProbParams)\big) \, \PdfBSDF(\diriRV|\pos,\diro) \nonumber \\
    &\qquad+ \selectionProb(\pos,\diro;\selectionProbParams) \, \PdfNIS(\diriRV|\pos,\diro;\nisParams)\,.
    \label{eq:sampling-pdf-blend-bsdf}
\end{align}
\autoref{fig:all-components} visualizes how each component of our trainable CVs fits into the light-transport integral equation.

\subsection{Path Termination}%
\label{sec:path-truncation}

\begin{figure*}
    \setlength{\fboxrule}{10pt}%
\setlength{\insetvsep}{20pt}%
\setlength{\tabcolsep}{-1pt}%
\renewcommand{\arraystretch}{1}%
\small%
\hspace*{-2mm}%
\begin{tabular}{rcccccccc}
  {} & {} & \multicolumn{4}{c}{Unbiased} & \multicolumn{2}{c}{Biased} \\
  \cmidrule(lr){3-6}
  \cmidrule(lr){7-8}
  {} & {} & PT & PPG & NIS++ & NCV & NCV + heuristic & CV Integral & Reference \\
    \setInset{A}{red}{180}{500}{51}{38}%
    \setInset{B}{orange}{975}{975}{51}{38}%
    \rotatebox{90}{\hspace{-1.80cm}\Bathroom{}}\hspace{0.14cm} &
    \addBeautyCrop{images/fig-references/bathroom-reference-lq.jpg}{0.27}{1920}{1080}{0}{0}{1920}{1080} &
    \addInsets{images/fig-results/bathroom-PT_u.jpg} &
    \addInsets{images/fig-results/bathroom-PPG_u.jpg} &
    \addInsets{images/fig-results/bathroom-NIS++_u.jpg} &
    \addInsets{images/fig-results/bathroom-NCV_u.jpg} &
    \addInsets{images/fig-results/bathroom-NCV_b.jpg} &
    \addInsets{images/fig-results/bathroom-CV-Integral_b.jpg} &
    \addInsets{images/fig-references/bathroom-reference.jpg} \\
  & \multicolumn{1}{r}{MAPE:} & 0.112 & 0.073 & 0.037 & \textbf{0.030} & \textbf{0.017} & 0.023 \\
    \setInset{A}{red}{720}{90}{51}{38}%
    \setInset{B}{orange}{1150}{730}{51}{38}%
    \rotatebox{90}{\hspace{-1.70cm}\Bedroom{}}\hspace{0.14cm} &
    \addBeautyCrop{images/fig-references/bedroom-reference-lq.jpg}{0.27}{1920}{1080}{0}{0}{1920}{1080} &
    \addInsets{images/fig-results/bedroom-PT_u.jpg} &
    \addInsets{images/fig-results/bedroom-PPG_u.jpg} &
    \addInsets{images/fig-results/bedroom-NIS++_u.jpg} &
    \addInsets{images/fig-results/bedroom-NCV_u.jpg} &
    \addInsets{images/fig-results/bedroom-NCV_b.jpg} &
    \addInsets{images/fig-results/bedroom-CV-Integral_b.jpg} &
    \addInsets{images/fig-references/bedroom-reference.jpg} \\
  & \multicolumn{1}{r}{MAPE:} & 0.064 & 0.035 & 0.031 & \textbf{0.026} & \textbf{0.020} & 0.035 \\
    \setInset{A}{red}{400}{415}{51}{38}%
    \setInset{B}{orange}{450}{750}{51}{38}%
    \rotatebox{90}{\hspace{-1.75cm}\Bookshelf{}}\hspace{0.14cm} &
    \addBeautyCrop{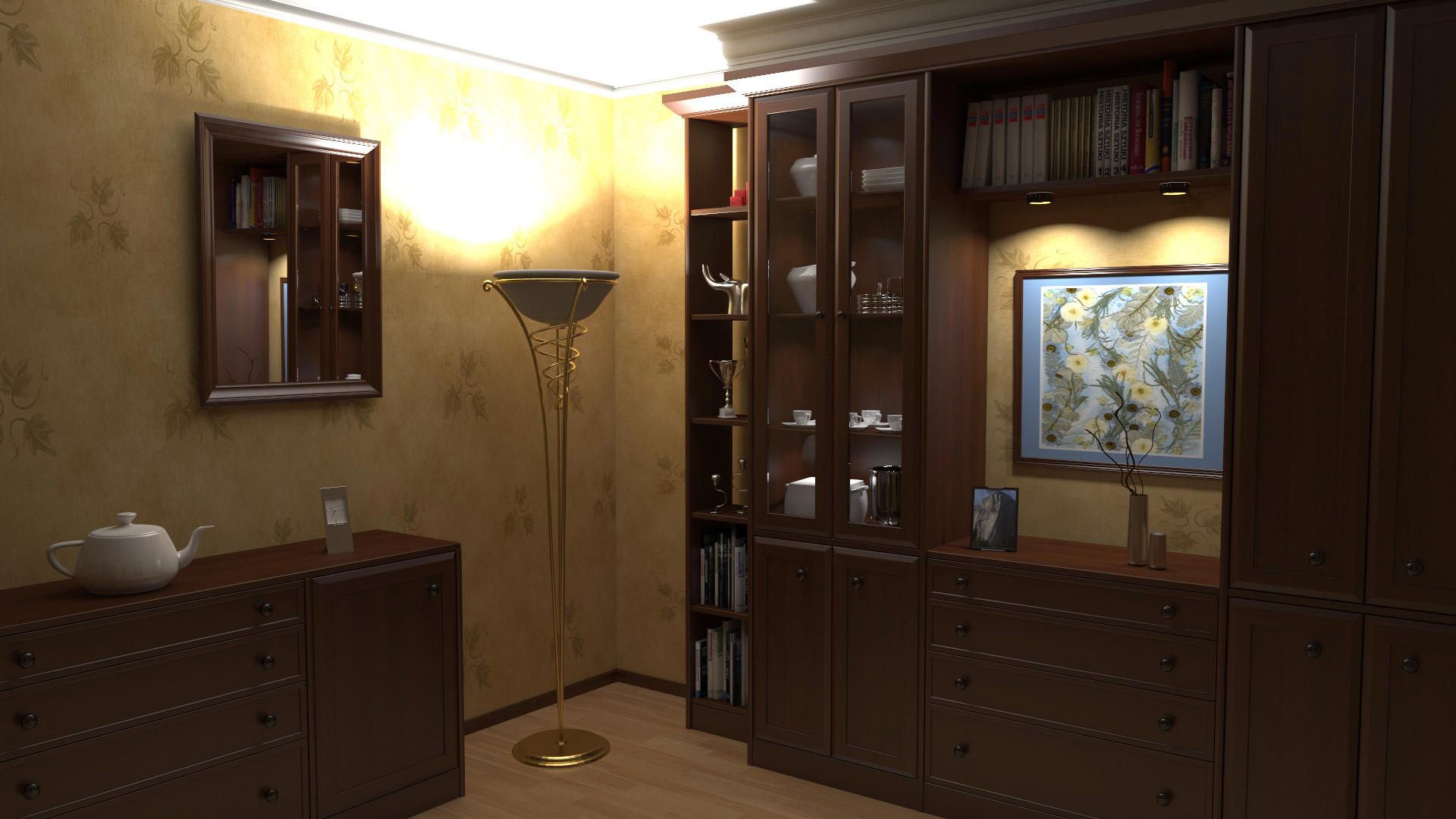}{0.27}{1920}{1080}{0}{0}{1920}{1080} &
    \addInsets{images/fig-results/bookshelf-PT_u.jpg} &
    \addInsets{images/fig-results/bookshelf-PPG_u.jpg} &
    \addInsets{images/fig-results/bookshelf-NIS++_u.jpg} &
    \addInsets{images/fig-results/bookshelf-NCV_u.jpg} &
    \addInsets{images/fig-results/bookshelf-NCV_b.jpg} &
    \addInsets{images/fig-results/bookshelf-CV-Integral_b.jpg} &
    \addInsets{images/fig-references/bookshelf-reference.jpg} \\
  & \multicolumn{1}{r}{MAPE:} & 0.658 & 0.045 & 0.047 & \textbf{0.030} & \textbf{0.025} & 0.080 \\
    \setInset{A}{red}{700}{400}{51}{38}%
    \setInset{B}{orange}{1095}{575}{51}{38}%
    \rotatebox{90}{\hspace{-1.75cm}\Bottle{}}\hspace{0.14cm} &
    \addBeautyCrop{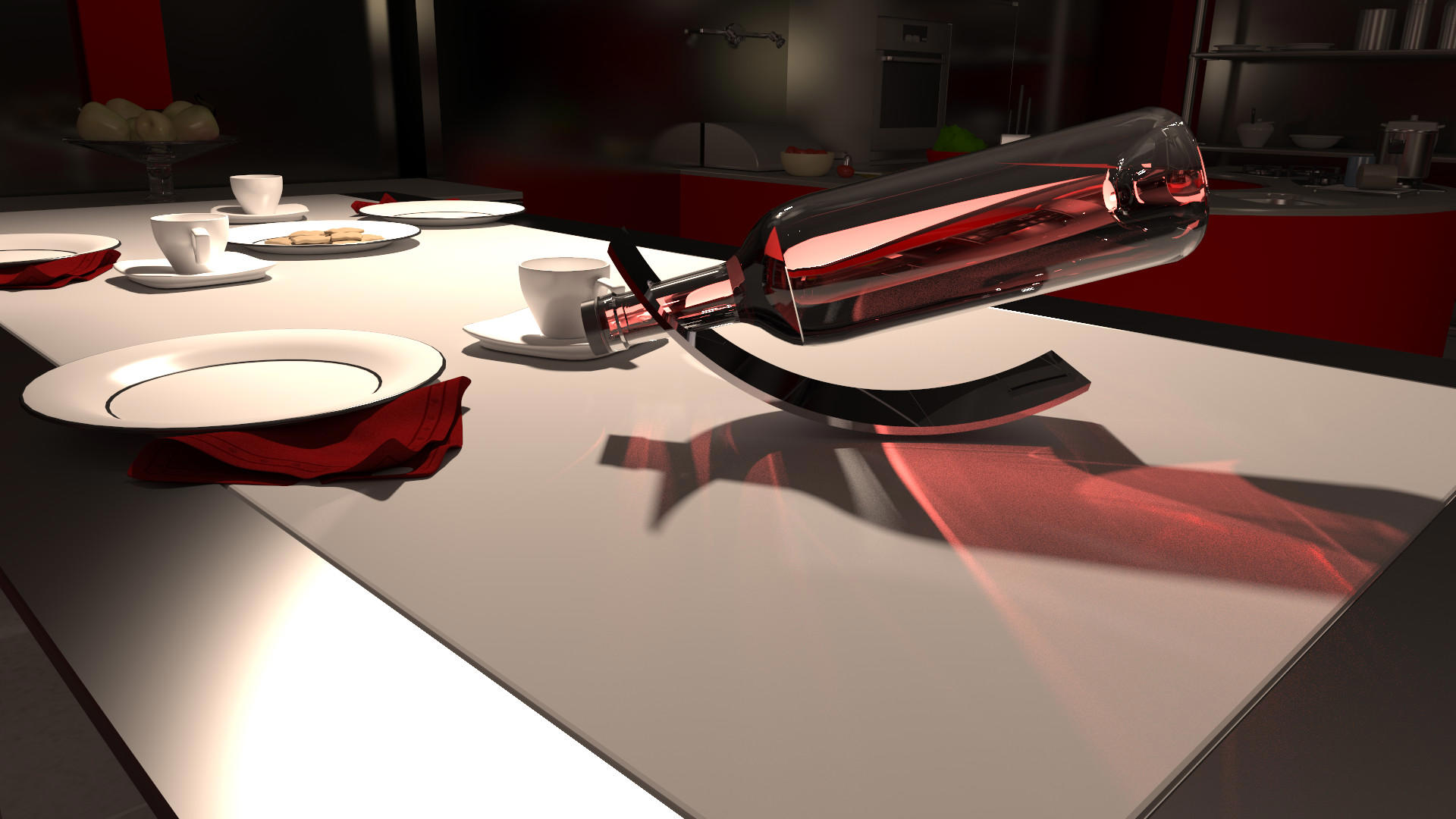}{0.27}{1920}{1080}{0}{0}{1920}{1080} &
    \addInsets{images/fig-results/bottle-PT_u.jpg} &
    \addInsets{images/fig-results/bottle-PPG_u.jpg} &
    \addInsets{images/fig-results/bottle-NIS++_u.jpg} &
    \addInsets{images/fig-results/bottle-NCV_u.jpg} &
    \addInsets{images/fig-results/bottle-NCV_b.jpg} &
    \addInsets{images/fig-results/bottle-CV-Integral_b.jpg} &
    \addInsets{images/fig-references/bottle-reference.jpg} \\
  & \multicolumn{1}{r}{MAPE:} & 0.848 & 0.088 & 0.062 & \textbf{0.056} & \textbf{0.046} & 0.163 \\
    \setInset{A}{red}{675}{135}{51}{38}%
    \setInset{B}{orange}{1020}{290}{51}{38}%
    \rotatebox{90}{\hspace{-1.95cm}\SpectralBox{}}\hspace{0.14cm} &
    \addBeautyCrop{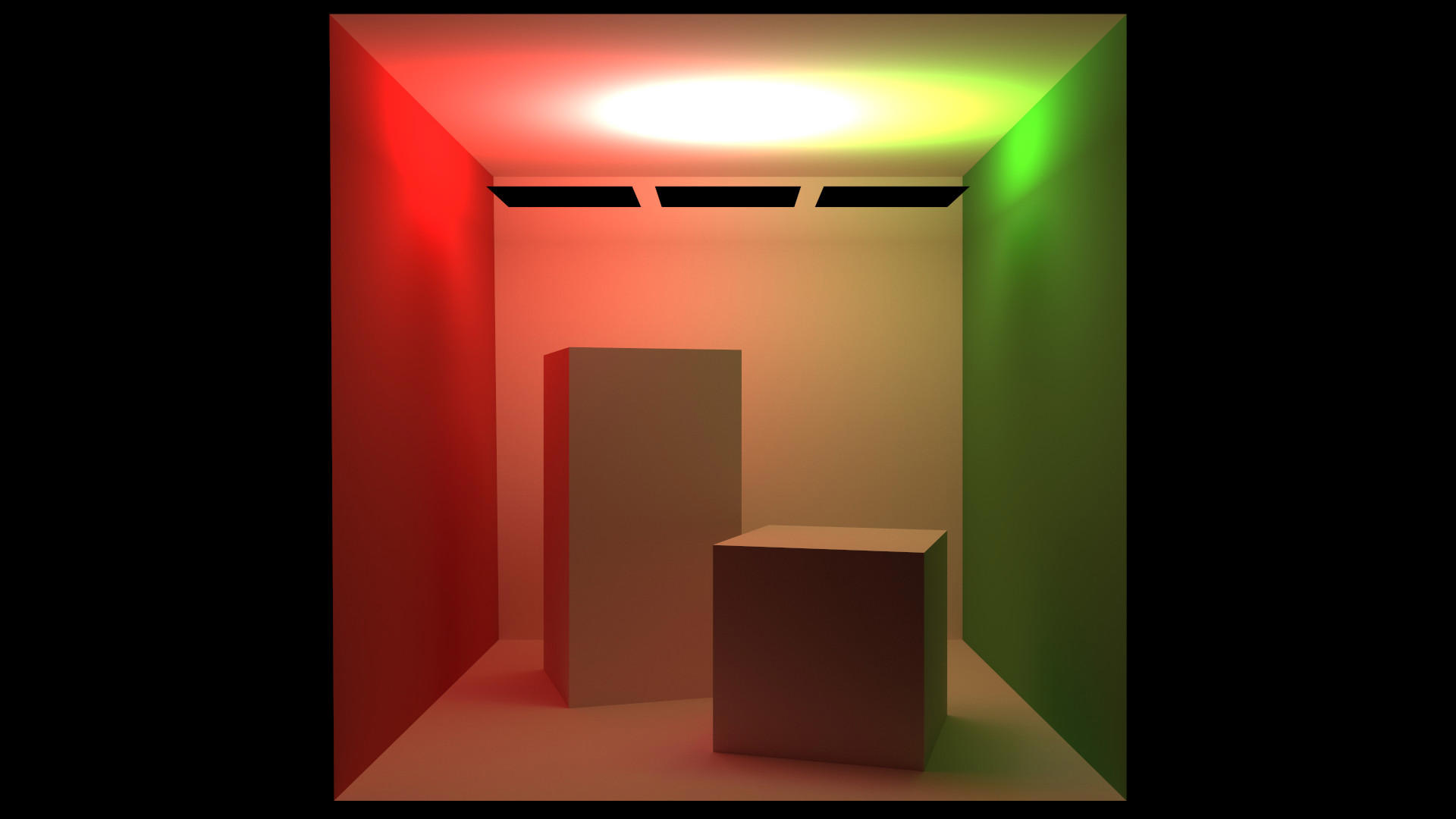}{0.27}{1920}{1080}{0}{0}{1920}{1080} &
    \addInsets{images/fig-results/spectral-box-PT_u.jpg} &
    \addInsets{images/fig-results/spectral-box-PPG_u.jpg} &
    \addInsets{images/fig-results/spectral-box-NIS++_u.jpg} &
    \addInsets{images/fig-results/spectral-box-NCV_u.jpg} &
    \addInsets{images/fig-results/spectral-box-NCV_b.jpg} &
    \addInsets{images/fig-results/spectral-box-CV-Integral_b.jpg} &
    \addInsets{images/fig-references/spectral-box-reference.jpg} \\
  & \multicolumn{1}{r}{MAPE:} & 0.030 & 0.014 & 0.015 & \textbf{0.009} & \textbf{0.008} & 0.019 \\
    \setInset{A}{red}{475}{675}{51}{38}%
    \setInset{B}{orange}{1135}{700}{51}{38}%
    \rotatebox{90}{\hspace{-2.00cm}\VeachDoor{}}\hspace{0.14cm} &
    \addBeautyCrop{images/fig-references/veach-door-reference-lq.jpg}{0.27}{1920}{1080}{0}{0}{1920}{1080} &
    \addInsets{images/fig-results/veach-door-PT_u.jpg} &
    \addInsets{images/fig-results/veach-door-PPG_u.jpg} &
    \addInsets{images/fig-results/veach-door-NIS++_u.jpg} &
    \addInsets{images/fig-results/veach-door-NCV_u.jpg} &
    \addInsets{images/fig-results/veach-door-NCV_b.jpg} &
    \addInsets{images/fig-results/veach-door-CV-Integral_b.jpg} &
    \addInsets{images/fig-references/veach-door-reference.jpg} \\
  & \multicolumn{1}{r}{MAPE:} & 0.532 & 0.084 & 0.060 & \textbf{0.036} & \textbf{0.024} & 0.034 \\
\end{tabular}

    \vspace{-2mm}
    \caption{\label{fig:images}
        Neural control variates (NCV) compared to \ADD{our improved variant (NIS++) of neural importance sampling~\citep{mueller2019nis}}, practical path guiding with recent improvements (PPG)~\citep{mueller19guiding}, and uni-directional path tracing (PT).
        All images have a resolution of $1920 \times 1080$ pixels and were rendered in 2 hours.
        Unbiased NCV consistently achieves a moderate MAPE improvement.
        \ADD{The biggest benefit is seen in scenes with smooth, indirect illumination (\Bathroom{}, \SpectralBox{}, and \VeachDoor{}).
        Approximating the tail contribution of light-transport paths by the learned CV integral driven by our heuristic described in \autoref{sec:path-truncation} (``NCV + heuristic'' column) allows for additional improvements, most noticeable in the \Bathroom{} and \VeachDoor{} scenes.}
        This variant of NCV reduces noise significantly while introducing only little visible bias, unlike na\"{\i}vely evaluating the CV integral at the first non-specular path vertex (``CV Integral'' column).
    }
\end{figure*}

The recursive estimation of radiance terminates when the path escapes the scene or hits a black-body radiator that does not scatter light.
Since the integral component of the CV approximates the scattered light field well in many cases, we considered skipping the evaluation of the correction term, thereby truncating the path and producing a biased radiance estimate.
\autoref{fig:images} (column CV Integral) visualizes the neural scattered light field $\CV$ at non-specular surfaces that are directly visible from the camera or seen through specular interactions.
Compared to the reference (right-most column) the approximation error of the neural light field, which manifests as low-frequency variations and blurry appearance, is not suitable for direct visualization.
However, deferring the approximation error to higher-order bounces (such as in final gathering for photon mapping) may strike a good balance between visual quality and computation cost (``Biased NCV (Ours)'' column in \autoref{fig:images}).

We utilize a simple criterion for ignoring the correction term.
The criterion measures the stochastic area-spread of path vertices, which \citep{bekaert2003} proposed to use as the photon-mapping filter radius.
Once the area spread becomes sufficiently large, we terminate the path and approximate $\reflectedRadiance(\pos,\diro)$ by $\CV(\pos,\diro; \CVParams)$.

Sampling of direction $\diro$ using $\PdfMC(\diriRV | \pos, \diro)$ at path vertex $\pos$ induces the area spread of
\begin{align}
  \AreaHeuristic(\pos',\pos) = \frac{1}{\PdfMC(\pos' | \pos, \diro)} = \frac{\|\pos - \pos'\|^2}{\PdfMC(\diriRV | \pos, \diro) \, | \cos{\fsangle'} |}
\end{align}
around the next path vertex $\pos'$, where $\fsangle'$ is the incidence angle at~$\pos'$.
The cumulative area spread at the $n$-th path vertex is the convolution of the spreads induced at all previous vertices.
Assuming isotropic Gaussian spreads with variance $\sqrt{\AreaHeuristic(\pos',\pos)}$ and parallel surfaces, this convolution can be approximated as:
\begin{align}
    \AreaHeuristic(\pos_1, \ldots, \pos_n) = {\left( \sum_{i=2}^n \sqrt{\AreaHeuristic(\pos_i,\pos_{i-1})} \right)}^2 \,.
\end{align}
\begin{wrapfigure}{r}{0.40\columnwidth}
    \centering
    \begin{overpic}[width=0.40\columnwidth]{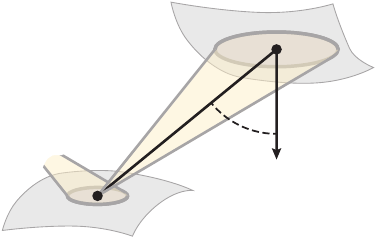}
        \put(24,4) {$\pos$}
        \put(77,48) {$\pos'$}
        \put(65,35) {$\fsangle'$}
        \put(28,50) {$\AreaHeuristic(\pos',\pos)$}
    \end{overpic}
\end{wrapfigure}
We compare this cumulative area spread to the pixel footprint projected onto the primary vertex $\pos_1$.
If the projected pixel footprint is more than \num{10000}$\times$ smaller than the path's cumulative area spread---loosely corresponding to a \num{100}-pixel-wide image-space filter---we terminate the path into $\CV(\pos,\diri)$.
Otherwise, we keep applying our unbiased control variates and recursively evaluate the heuristic at the next path vertex.

The heuristic path termination shortens the mean path length and removes a significant amount of noise at the cost of a small amount of visible bias; see \autoref{fig:teaser} and \autoref{sec:results}.

Our heuristic area spread is a simplified version of path differentials and
could be made more accurate by taking into account anisotropy and additional dimensions of variation, for instance via covariance tracing~\cite{Belcour:2013:COV}.
In our experiments, our heuristic worked sufficiently fine and hence we leave this extension to future work.
Stochastic path termination via Russian roulette is discussed in \autoref{sec:discussion-and-future-work}.

\subsection{Implementation}

We implemented our neural control variates as well as neural importance sampling within Tensorflow~\cite{tensorflow2015-whitepaper}.
The rendering algorithm is implemented in the Mitsuba renderer~\cite{Mitsuba}, interfacing with Tensorflow to invoke the neural networks.

Rendering and training happen \emph{simultaneously}, following the methodology of Neural Importance Sampling~\citep{mueller2019nis}:\ we begin by initializing the trainable parameters using Xavier initialization~\cite{Glorot:2010} and then optimize our composite loss~(\autoref{eq:composite-loss}) using Adam~\cite{KingmaB14}.
We use the CPU to perform light-transport computations and two GPUs to perform the neural-network-related computations.
One GPU is responsible solely for training whereas the other is responsible for utilizing the current trained model to reduce variance as per~\autoref{eq:reflected-radiance-cv-estimator}.
Training and variance reduction mutually benefit each other---the models are synced every second---making our algorithm a variant of reinforcement learning.

Mitsuba communicates with Tensorflow in batches of \num{65536} samples, where every \emph{path vertex} is a single sample.
At each path vertex $(\pos,\diro)$, we initially proceed identically to NIS~\citep{mueller2019nis}:\ the renderer first queries the MIS selection probabilities $\selectionProb(\pos,\diro;\Params_{{\bsdf}})$ and $\selectionProb(\pos,\diro;\Params_{{\NIS}})$.
Next, according to the selection probabilities, the renderer probabilistically selects either BSDF sampling or NIS.\@
If BSDF sampling is selected, the algorithm queries the NIS PDF for the sampled direction~$\diri$. If NIS is selected, the algorithm queries a sample of $\diri$ via NIS.\@
The renderer continues by querying the CV integral~$\CV(\pos,\diro)$, the CV shape~$\cv(\pos,\diro,\diri)$, as well as $\cvCoef(\pos,\diro)$, and applies them according to \autoref{eq:reflected-radiance-cv-estimator}.
After a light path has been completed, the reflected radiance at each vertex, along with the vertex's metadata, is put into a ring buffer that keeps track of the past \num{1048576} vertices.
The training GPU continuously assembles training batches by picking random samples from the ring buffer to minimize correlations within each batch.

\paragraph{Specular BSDFs}
BSDFs with Dirac-delta components (henceforth referred to as ``specular'') typically require special treatment because they are not square integrable. 
Inserting specular components into our equations results in the following behavior that needs to be explicitly implemented.
There are two cases: (i) the BSDF has specular \emph{and} non-specular components. In this case, the selection probability $\selectionProb$ is used in the standard way to select either BSDF sampling or NIS.\@ If BSDF sampling is selected and one of its specular components is sampled, then the NIS PDF and our parametric control variate are treated as zero. Otherwise (i.e.\ when either NIS or a smooth BSDF component is sampled), we apply the neural control variate, but with $\cv$ multiplied by the total probability of sampling NIS or a smooth BSDF component; $\CV$ should not be multiplied by this number.
(ii) the BSDF has only specular components. In this case, regular path tracing is used (without the influence of any of our parametric models).

\paragraph{Iterative rendering}
We apply the same iterative rendering scheme as \citet{mueller2019nis}:\ we render $M = \lfloor \log_2 (N+1) \rfloor$ images with power-of-two sample counts $2^i; i \in \{0, \ldots, M\}$, except for the last iteration which may have fewer samples due to running out of render time.
To obtain the final image, we average all images, weighted by the reciprocal of a robust numerical estimate of their mean pixel variance~\citep{mueller19guiding} in order to limit the impact of high-variance initial samples.

\begin{table}
    \caption{\label{tab:parameters}
        Parameters $y$ that are fed to our parametric models along with their encoding and dimensionality.
        We apply one-blob (ob) encoding~\citep{mueller2019nis} to all parameters except for the reflectances and the transmittance.
    }
    \begin{tabularx}{\columnwidth}{rcc}
        \toprule
        Parameter & Symbol & Encoding \\
        \midrule
        Scattered dir. & $\diro \in \Sphere$ & $\OneBlob(\diro / 2 + 0.5) \in \R^{3\times32}$ \\
        Position & $\pos \in \R^3$ & $\OneBlob(\pos) \in \R^{3\times32}$ \\
        Path length & $k \in \N$ & $\OneBlob(k / k_\text{max}) \in \R^{32}$ \\
        \midrule
        Surface normal & $\normal(\pos) \in \Sphere$ & $\OneBlob(\normal(\pos) / 2 + 0.5) \in \R^{3\times32}$ \\
        Surface roughness & $r(\pos, \diro) \in \R$ & $\OneBlob\left(1 - e^{-r(\pos, \diro)}\right) \in \R^{32}$ \\
        Diffuse reflectance & $f_\mathrm{dr}(\pos, \diro) \in \R^3$ & $f_\mathrm{dr}(\pos, \diro) \in \R^3$ \\
        Specular reflectance & $f_\mathrm{sr}(\pos, \diro) \in \R^3$ & $f_\mathrm{sr}(\pos, \diro) \in \R^3$ \\
        Transmittance & $f_\mathrm{t}(\pos, \diro) \in \R^3$ & $f_\mathrm{t}(\pos, \diro) \in \R^3$ \\
        \bottomrule
    \end{tabularx}
\end{table}

\paragraph{Parameter augmentation for neural networks}
As observed by \citet{Ren:2013}, the approximation power of a parametric model to learn the light field as a function of ${(\pos,\diro)}$ may be dramatically improved when additional quantities are provided as input.
\autoref{tab:parameters} lists all parameters that we feed to our parametric models in addition to
the query location and direction ${(\pos,\diro)}$: the surface normal, the surface roughness, the diffuse and specular reflectance, and the transmittance.
Directions are parameterized in a global coordinate frame as done by \citet{mueller2017practical}.

We also include the path length $k$ when the maximum path length is capped to some finite number $k_\text{max}$; in all our results we use $k_\text{max} = 10$.
In this case, the networks must learn progressively less indirect illumination as $k$ approaches $k_\text{max}$.

All quantities are normalized such that they fall within the unit hypercube of their respective dimensionality. 
Those quantities that have a highly non-linear relationship with the light field (all but the reflectances and the transmittance) are additionally one-blob encoded~\citep{mueller2019nis}, denoted by $\OneBlob(x)$.

\paragraph{Network and flow architecture}
For both normalizing flows, i.e.\ the multi-channel flow for the CV shape and the standard flow for NIS, we use the piecewise-quadratic warp proposed by~\citep{mueller2019nis} with \num{64} bins and a uniform latent distribution $\pdf_\latentSpace(x') \equiv \PdfUniform(x')$.
Both flows use ${L=2}$ warps to make the total number of warps, neural networks, and trainable parameters comparable to standalone NIS~\citep{mueller2019nis}, which uses a single flow with ${L=4}$ warps.

All neural networks---i.e.\ those that parameterize our warps as well as the one that predicts $\CV$, $\cvCoef$, and $\selectionProb$---use the same architecture: a fully connected residual network~\citep{He:2015} with \num{2} residual blocks that each have \num{2} layers with \num{256} neurons.
We also experimented with other architectures, such as multi-layer perceptrons (MLPs) and U-nets (as proposed for NIS), and remark that architectural differences had only small, almost immeasurable impact on the results in our tests.

\ADD{Lastly, matrix multiplications are computed at half precision in order to take advantage of dedicated hardware.}

\begin{table*}
    \caption{\label{tab:results}
        We report equal-time \emph{mean absolute percentage error}~(MAPE) of several machine-learning-based variance reduction techniques on 18 test scenes.
        Bold entries indicate lowest error among unbiased/biased techniques.
        All images have a resolution of $1920 \times 1080$ pixels and were rendered in 2 hours.
        The achieved samples per pixel are written next to the error numbers.
        Our unbiased neural control variates~(NCV) outperform neural importance sampling~(NIS and our improved NIS++) on almost all scenes, \ADD{except for the \Sponza{}.
        The performance advantage is larger when the illumination is mostly indirect, such as in interior scenes (e.g.\ \Bathroom{}, \SpectralBox{}, \Bookshelf{} and \VeachDoor{}).
        Using the integral $\CV$ of the neural control variate without correcting for its error after interactions with rough surfaces (NCV + heuristic) results in a biased image, but in most cases with a significant additional error reduction---this implies that the variance reduction outweighs the amount of introduced bias.
        In one scene (\VeachLamp{})}, lowest MAPE is achieved by using uncorrected $\CV$ already at the first camera vertex (CV Integral), but the produced image suffers from visually displeasing artifacts (see \autoref{fig:images} and the supplementary material).
    }
    \vspace{-3mm}
    \small
\rowcolors{2}{white}{gray!8}
\setlength{\tabcolsep}{6.33pt}
\begin{tabular*}{\textwidth}{rccccccc}
  \toprule
  {} & \multicolumn{5}{c}{Unbiased} & \multicolumn{2}{c}{Biased} \\
  \cmidrule(lr){2-6}
  \cmidrule(lr){7-8}
  \rowcolor{white}
  {} & {} & \citep{mueller19guiding} & \citep{mueller2019nis} & \multicolumn{4}{c}{Ours} \\
  \cmidrule(lr){3-3}
  \cmidrule(lr){4-4}
  \cmidrule(lr){5-8}
  \rowcolor{white}
  {} & PT & PPG & NIS & NIS++ & NCV & NCV + heuristic & CV Integral \\
  \midrule
  \small{\Artroom{}} & \small{\makebox[0.6cm]{1.393} \, \makebox[0.8cm]{\textcolor{gray}{1,756spp}}} & \small{\makebox[0.6cm]{0.108} \, \makebox[0.8cm]{\textcolor{gray}{2,998spp}}} & \small{\makebox[0.6cm]{0.079} \, \makebox[0.8cm]{\textcolor{gray}{1,231spp}}} & \small{\makebox[0.6cm]{0.062} \, \makebox[0.8cm]{\textcolor{gray}{1,271spp}}} & \small{\makebox[0.6cm]{\textbf{0.056}} \, \makebox[0.8cm]{\textcolor{gray}{1,188spp}}} & \small{\makebox[0.6cm]{\textbf{0.044}} \, \makebox[0.8cm]{\textcolor{gray}{1,188spp}}} & \small{\makebox[0.6cm]{0.091} \, \makebox[0.8cm]{\textcolor{gray}{1,188spp}}} \\
  \small{\Bathroom{}} & \small{\makebox[0.6cm]{0.112} \, \makebox[0.8cm]{\textcolor{gray}{1,785spp}}} & \small{\makebox[0.6cm]{0.073} \, \makebox[0.8cm]{\textcolor{gray}{2,155spp}}} & \small{\makebox[0.6cm]{0.041} \, \makebox[0.8cm]{\textcolor{gray}{736spp}}} & \small{\makebox[0.6cm]{0.037} \, \makebox[0.8cm]{\textcolor{gray}{717spp}}} & \small{\makebox[0.6cm]{\textbf{0.030}} \, \makebox[0.8cm]{\textcolor{gray}{756spp}}} & \small{\makebox[0.6cm]{\textbf{0.017}} \, \makebox[0.8cm]{\textcolor{gray}{756spp}}} & \small{\makebox[0.6cm]{0.023} \, \makebox[0.8cm]{\textcolor{gray}{756spp}}} \\
  \small{\Bedroom{}} & \small{\makebox[0.6cm]{0.064} \, \makebox[0.8cm]{\textcolor{gray}{1,727spp}}} & \small{\makebox[0.6cm]{0.035} \, \makebox[0.8cm]{\textcolor{gray}{2,187spp}}} & \small{\makebox[0.6cm]{0.031} \, \makebox[0.8cm]{\textcolor{gray}{719spp}}} & \small{\makebox[0.6cm]{0.031} \, \makebox[0.8cm]{\textcolor{gray}{636spp}}} & \small{\makebox[0.6cm]{\textbf{0.026}} \, \makebox[0.8cm]{\textcolor{gray}{705spp}}} & \small{\makebox[0.6cm]{\textbf{0.020}} \, \makebox[0.8cm]{\textcolor{gray}{705spp}}} & \small{\makebox[0.6cm]{0.035} \, \makebox[0.8cm]{\textcolor{gray}{705spp}}} \\
  \small{\Bookshelf{}} & \small{\makebox[0.6cm]{0.658} \, \makebox[0.8cm]{\textcolor{gray}{2,113spp}}} & \small{\makebox[0.6cm]{0.045} \, \makebox[0.8cm]{\textcolor{gray}{2,749spp}}} & \small{\makebox[0.6cm]{0.054} \, \makebox[0.8cm]{\textcolor{gray}{1,012spp}}} & \small{\makebox[0.6cm]{0.047} \, \makebox[0.8cm]{\textcolor{gray}{1,036spp}}} & \small{\makebox[0.6cm]{\textbf{0.030}} \, \makebox[0.8cm]{\textcolor{gray}{1,017spp}}} & \small{\makebox[0.6cm]{\textbf{0.025}} \, \makebox[0.8cm]{\textcolor{gray}{1,017spp}}} & \small{\makebox[0.6cm]{0.080} \, \makebox[0.8cm]{\textcolor{gray}{1,017spp}}} \\
  \small{\Bottle{}} & \small{\makebox[0.6cm]{0.848} \, \makebox[0.8cm]{\textcolor{gray}{2,110spp}}} & \small{\makebox[0.6cm]{0.088} \, \makebox[0.8cm]{\textcolor{gray}{3,612spp}}} & \small{\makebox[0.6cm]{0.084} \, \makebox[0.8cm]{\textcolor{gray}{1,322spp}}} & \small{\makebox[0.6cm]{0.062} \, \makebox[0.8cm]{\textcolor{gray}{1,451spp}}} & \small{\makebox[0.6cm]{\textbf{0.056}} \, \makebox[0.8cm]{\textcolor{gray}{1,280spp}}} & \small{\makebox[0.6cm]{\textbf{0.046}} \, \makebox[0.8cm]{\textcolor{gray}{1,280spp}}} & \small{\makebox[0.6cm]{0.163} \, \makebox[0.8cm]{\textcolor{gray}{1,280spp}}} \\
  \small{\CornellBox{}} & \small{\makebox[0.6cm]{0.035} \, \makebox[0.8cm]{\textcolor{gray}{8,614spp}}} & \small{\makebox[0.6cm]{0.009} \, \makebox[0.8cm]{\textcolor{gray}{3,873spp}}} & \small{\makebox[0.6cm]{0.009} \, \makebox[0.8cm]{\textcolor{gray}{906spp}}} & \small{\makebox[0.6cm]{0.009} \, \makebox[0.8cm]{\textcolor{gray}{908spp}}} & \small{\makebox[0.6cm]{\textbf{0.005}} \, \makebox[0.8cm]{\textcolor{gray}{1,290spp}}} & \small{\makebox[0.6cm]{\textbf{0.006}} \, \makebox[0.8cm]{\textcolor{gray}{1,290spp}}} & \small{\makebox[0.6cm]{0.020} \, \makebox[0.8cm]{\textcolor{gray}{1,290spp}}} \\
  \small{\CrytekSponza{}} & \small{\makebox[0.6cm]{1.340} \, \makebox[0.8cm]{\textcolor{gray}{1,518spp}}} & \small{\makebox[0.6cm]{0.056} \, \makebox[0.8cm]{\textcolor{gray}{2,417spp}}} & \small{\makebox[0.6cm]{0.065} \, \makebox[0.8cm]{\textcolor{gray}{605spp}}} & \small{\makebox[0.6cm]{0.060} \, \makebox[0.8cm]{\textcolor{gray}{533spp}}} & \small{\makebox[0.6cm]{\textbf{0.051}} \, \makebox[0.8cm]{\textcolor{gray}{711spp}}} & \small{\makebox[0.6cm]{\textbf{0.048}} \, \makebox[0.8cm]{\textcolor{gray}{711spp}}} & \small{\makebox[0.6cm]{0.234} \, \makebox[0.8cm]{\textcolor{gray}{711spp}}} \\
  \small{\GlossyKitchen{}} & \small{\makebox[0.6cm]{1.450} \, \makebox[0.8cm]{\textcolor{gray}{2,092spp}}} & \small{\makebox[0.6cm]{0.071} \, \makebox[0.8cm]{\textcolor{gray}{2,391spp}}} & \small{\makebox[0.6cm]{0.075} \, \makebox[0.8cm]{\textcolor{gray}{988spp}}} & \small{\makebox[0.6cm]{0.049} \, \makebox[0.8cm]{\textcolor{gray}{978spp}}} & \small{\makebox[0.6cm]{\textbf{0.045}} \, \makebox[0.8cm]{\textcolor{gray}{875spp}}} & \small{\makebox[0.6cm]{\textbf{0.031}} \, \makebox[0.8cm]{\textcolor{gray}{875spp}}} & \small{\makebox[0.6cm]{0.114} \, \makebox[0.8cm]{\textcolor{gray}{875spp}}} \\
  \small{\CountryKitchen{}} & \small{\makebox[0.6cm]{0.696} \, \makebox[0.8cm]{\textcolor{gray}{2,070spp}}} & \small{\makebox[0.6cm]{0.068} \, \makebox[0.8cm]{\textcolor{gray}{3,013spp}}} & \small{\makebox[0.6cm]{0.073} \, \makebox[0.8cm]{\textcolor{gray}{1,039spp}}} & \small{\makebox[0.6cm]{0.066} \, \makebox[0.8cm]{\textcolor{gray}{1,026spp}}} & \small{\makebox[0.6cm]{\textbf{0.059}} \, \makebox[0.8cm]{\textcolor{gray}{1,014spp}}} & \small{\makebox[0.6cm]{\textbf{0.046}} \, \makebox[0.8cm]{\textcolor{gray}{1,014spp}}} & \small{\makebox[0.6cm]{0.067} \, \makebox[0.8cm]{\textcolor{gray}{1,014spp}}} \\
  \small{\Necklace{}} & \small{\makebox[0.6cm]{0.289} \, \makebox[0.8cm]{\textcolor{gray}{10,280spp}}} & \small{\makebox[0.6cm]{0.057} \, \makebox[0.8cm]{\textcolor{gray}{9,449spp}}} & \small{\makebox[0.6cm]{0.038} \, \makebox[0.8cm]{\textcolor{gray}{2,958spp}}} & \small{\makebox[0.6cm]{0.033} \, \makebox[0.8cm]{\textcolor{gray}{2,541spp}}} & \small{\makebox[0.6cm]{\textbf{0.031}} \, \makebox[0.8cm]{\textcolor{gray}{2,726spp}}} & \small{\makebox[0.6cm]{\textbf{0.030}} \, \makebox[0.8cm]{\textcolor{gray}{2,726spp}}} & \small{\makebox[0.6cm]{0.159} \, \makebox[0.8cm]{\textcolor{gray}{2,726spp}}} \\
  \small{\SwimmingPool{}} & \small{\makebox[0.6cm]{0.451} \, \makebox[0.8cm]{\textcolor{gray}{4,271spp}}} & \small{\makebox[0.6cm]{0.035} \, \makebox[0.8cm]{\textcolor{gray}{5,771spp}}} & \small{\makebox[0.6cm]{0.040} \, \makebox[0.8cm]{\textcolor{gray}{1,973spp}}} & \small{\makebox[0.6cm]{0.034} \, \makebox[0.8cm]{\textcolor{gray}{1,956spp}}} & \small{\makebox[0.6cm]{\textbf{0.031}} \, \makebox[0.8cm]{\textcolor{gray}{1,971spp}}} & \small{\makebox[0.6cm]{\textbf{0.028}} \, \makebox[0.8cm]{\textcolor{gray}{1,971spp}}} & \small{\makebox[0.6cm]{0.076} \, \makebox[0.8cm]{\textcolor{gray}{1,971spp}}} \\
  \small{\Spaceship{}} & \small{\makebox[0.6cm]{0.017} \, \makebox[0.8cm]{\textcolor{gray}{6,489spp}}} & \small{\makebox[0.6cm]{0.009} \, \makebox[0.8cm]{\textcolor{gray}{7,529spp}}} & \small{\makebox[0.6cm]{0.009} \, \makebox[0.8cm]{\textcolor{gray}{2,775spp}}} & \small{\makebox[0.6cm]{0.008} \, \makebox[0.8cm]{\textcolor{gray}{2,707spp}}} & \small{\makebox[0.6cm]{\textbf{0.006}} \, \makebox[0.8cm]{\textcolor{gray}{2,585spp}}} & \small{\makebox[0.6cm]{\textbf{0.007}} \, \makebox[0.8cm]{\textcolor{gray}{2,585spp}}} & \small{\makebox[0.6cm]{0.027} \, \makebox[0.8cm]{\textcolor{gray}{2,585spp}}} \\
  \small{\SpectralBox{}} & \small{\makebox[0.6cm]{0.030} \, \makebox[0.8cm]{\textcolor{gray}{9,563spp}}} & \small{\makebox[0.6cm]{0.014} \, \makebox[0.8cm]{\textcolor{gray}{4,348spp}}} & \small{\makebox[0.6cm]{0.017} \, \makebox[0.8cm]{\textcolor{gray}{958spp}}} & \small{\makebox[0.6cm]{0.015} \, \makebox[0.8cm]{\textcolor{gray}{946spp}}} & \small{\makebox[0.6cm]{\textbf{0.009}} \, \makebox[0.8cm]{\textcolor{gray}{1,373spp}}} & \small{\makebox[0.6cm]{\textbf{0.008}} \, \makebox[0.8cm]{\textcolor{gray}{1,373spp}}} & \small{\makebox[0.6cm]{0.019} \, \makebox[0.8cm]{\textcolor{gray}{1,373spp}}} \\
  \small{\Sponza{}} & \small{\makebox[0.6cm]{1.614} \, \makebox[0.8cm]{\textcolor{gray}{1,904spp}}} & \small{\makebox[0.6cm]{0.060} \, \makebox[0.8cm]{\textcolor{gray}{3,007spp}}} & \small{\makebox[0.6cm]{0.049} \, \makebox[0.8cm]{\textcolor{gray}{599spp}}} & \small{\makebox[0.6cm]{\textbf{0.040}} \, \makebox[0.8cm]{\textcolor{gray}{550spp}}} & \small{\makebox[0.6cm]{0.042} \, \makebox[0.8cm]{\textcolor{gray}{738spp}}} & \small{\makebox[0.6cm]{\textbf{0.022}} \, \makebox[0.8cm]{\textcolor{gray}{738spp}}} & \small{\makebox[0.6cm]{0.066} \, \makebox[0.8cm]{\textcolor{gray}{738spp}}} \\
  \small{\Staircase{}} & \small{\makebox[0.6cm]{0.137} \, \makebox[0.8cm]{\textcolor{gray}{1,458spp}}} & \small{\makebox[0.6cm]{0.029} \, \makebox[0.8cm]{\textcolor{gray}{2,553spp}}} & \small{\makebox[0.6cm]{0.023} \, \makebox[0.8cm]{\textcolor{gray}{1,288spp}}} & \small{\makebox[0.6cm]{0.022} \, \makebox[0.8cm]{\textcolor{gray}{1,156spp}}} & \small{\makebox[0.6cm]{\textbf{0.019}} \, \makebox[0.8cm]{\textcolor{gray}{943spp}}} & \small{\makebox[0.6cm]{\textbf{0.015}} \, \makebox[0.8cm]{\textcolor{gray}{943spp}}} & \small{\makebox[0.6cm]{0.036} \, \makebox[0.8cm]{\textcolor{gray}{943spp}}} \\
  \small{\Torus{}} & \small{\makebox[0.6cm]{0.214} \, \makebox[0.8cm]{\textcolor{gray}{12,108spp}}} & \small{\makebox[0.6cm]{0.021} \, \makebox[0.8cm]{\textcolor{gray}{9,470spp}}} & \small{\makebox[0.6cm]{0.020} \, \makebox[0.8cm]{\textcolor{gray}{3,714spp}}} & \small{\makebox[0.6cm]{0.018} \, \makebox[0.8cm]{\textcolor{gray}{2,955spp}}} & \small{\makebox[0.6cm]{\textbf{0.015}} \, \makebox[0.8cm]{\textcolor{gray}{3,497spp}}} & \small{\makebox[0.6cm]{\textbf{0.014}} \, \makebox[0.8cm]{\textcolor{gray}{3,497spp}}} & \small{\makebox[0.6cm]{0.014} \, \makebox[0.8cm]{\textcolor{gray}{3,497spp}}} \\
  \small{\VeachDoor{}} & \small{\makebox[0.6cm]{0.532} \, \makebox[0.8cm]{\textcolor{gray}{3,749spp}}} & \small{\makebox[0.6cm]{0.084} \, \makebox[0.8cm]{\textcolor{gray}{2,773spp}}} & \small{\makebox[0.6cm]{0.065} \, \makebox[0.8cm]{\textcolor{gray}{648spp}}} & \small{\makebox[0.6cm]{0.060} \, \makebox[0.8cm]{\textcolor{gray}{603spp}}} & \small{\makebox[0.6cm]{\textbf{0.036}} \, \makebox[0.8cm]{\textcolor{gray}{830spp}}} & \small{\makebox[0.6cm]{\textbf{0.024}} \, \makebox[0.8cm]{\textcolor{gray}{830spp}}} & \small{\makebox[0.6cm]{0.034} \, \makebox[0.8cm]{\textcolor{gray}{830spp}}} \\
  \small{\VeachLamp{}} & \small{\makebox[0.6cm]{0.532} \, \makebox[0.8cm]{\textcolor{gray}{4,079spp}}} & \small{\makebox[0.6cm]{0.069} \, \makebox[0.8cm]{\textcolor{gray}{2,204spp}}} & \small{\makebox[0.6cm]{0.077} \, \makebox[0.8cm]{\textcolor{gray}{582spp}}} & \small{\makebox[0.6cm]{0.068} \, \makebox[0.8cm]{\textcolor{gray}{537spp}}} & \small{\makebox[0.6cm]{\textbf{0.039}} \, \makebox[0.8cm]{\textcolor{gray}{723spp}}} & \small{\makebox[0.6cm]{0.026} \, \makebox[0.8cm]{\textcolor{gray}{723spp}}} & \small{\makebox[0.6cm]{\textbf{0.022}} \, \makebox[0.8cm]{\textcolor{gray}{723spp}}} \\
  \bottomrule
\end{tabular*}

    \vspace{-4mm}
\end{table*}

\begin{figure*}
    \hspace*{-1mm}\includegraphics[width=1.01\textwidth]{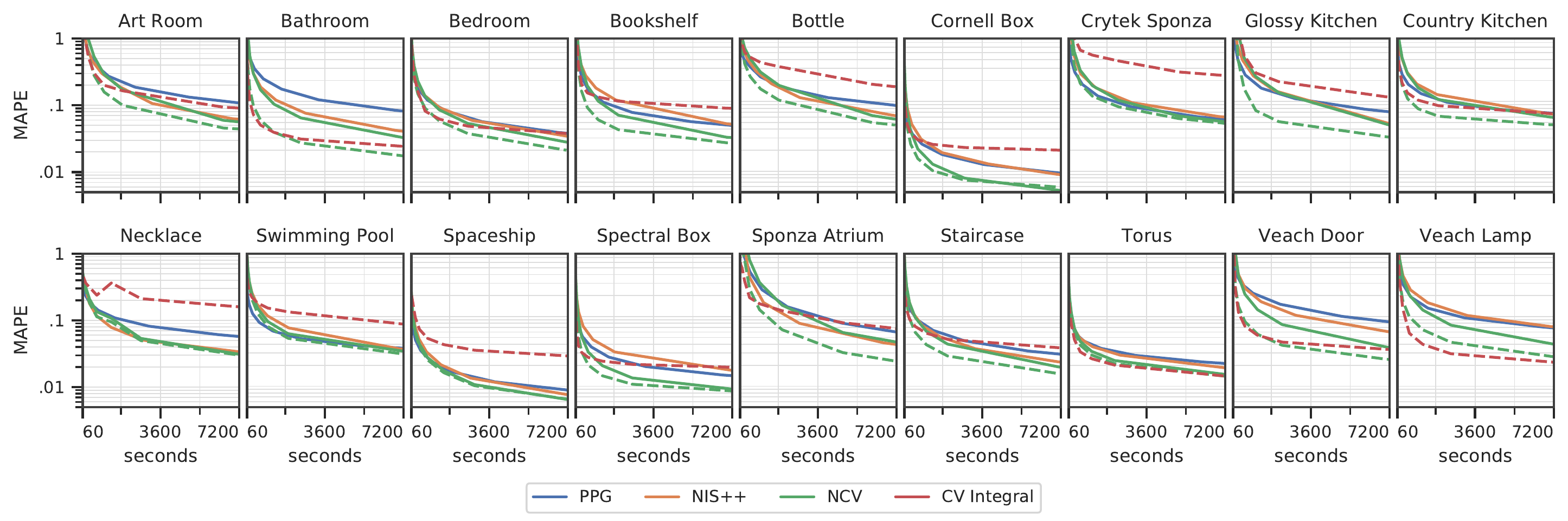}
    \vspace{-7mm}%
    \caption{\label{fig:convergence}
        MAPE convergence plots of practical path guiding (PPG)~\citep{mueller2017practical}, \ADD{our improved variant (NIS++) of neural importance sampling~\citep{mueller2019nis}}, and our neural control variates (NCV).
        The dashed red line corresponds to using the uncorrected CV integral at the first non-specular vertex.
        The dashed green line corresponds to the biased variant of our algorithm, where the uncorrected CV integral is used at a vertex determined by the path termination heuristic.
        It consistently outperforms all other techniques, except for the \Torus{} and the \VeachLamp{}, where using the CV integral at the first non-specular vertex performs best.
    }
    \vspace{-2mm}
\end{figure*}

\paragraph{Optimization}
We optimize the neural networks \emph{during} rendering in a reinforcement-learning fashion:\ the vertices of traced paths are used to optimize the networks by minimizing \autoref{eq:composite-loss}, while \emph{simultaneously} the current neural networks are used to drive variance reduction via \autoref{eq:reflected-radiance-cv-estimator}.
The networks reduce variance of their own training data \emph{and} that of the final image.

We minimize \autoref{eq:composite-loss} using Adam~\citep{KingmaB14}
and use a learning rate of \num{1e-3}, which decays in two steps: (i) $\sqrt{10} \times 10^{-4}$ after \num{25}\% of the rendering process and (ii) \num{1e-4} after \num{50}\% of the rendering process.
This learning-rate decay addresses a problem pointed out by \citet{mueller2019nis}, where learned distributions exhibited prolonged fluctuations in the later stages of training.

Lastly, we note that we do \emph{not} use batch normalization~\citep{ioffe15batchnorm}, because it detrimentally affected computational \emph{and} qualitative performance.

\section{Results and Analysis}%
\label{sec:results}

All results were produced on an NVIDIA DGX-1, using one Intel Xeon E5--2698 v4 CPU (20 cores; 40 threads) and two Tesla V100 GPUs (comparable to two RTX 2080Ti).
To gauge the practical usefulness of our technique, we compare render quality at \emph{equal time}, but we recognize that the performance depends strongly on the particular hardware setup.
Therefore, we also report \emph{samples per pixel} in \autoref{tab:results} for completeness.

We quantify rendering error using the ``mean absolute percentage error''~(MAPE), which strikes a good balance between being perceptually accurate and correlating with Monte Carlo standard deviation.
MAPE is defined as $\frac{1}{N} \sum_{i=1}^N |v_i - \hat{v}_i| / (\hat{v}_i + \epsilon)$, where $\hat{v}_i$ is the value of the $i$-th pixel in the reference image, $v_i$ is the value of the $i$-th rendered pixel, and $\epsilon = 0.01$ prevents near-black pixels from dominating the metric.
A rough estimate of Monte Carlo efficiency can be obtained by the \emph{reciprocal square root} of MAPE---i.e.\ a $2\times$ smaller MAPE loosely corresponds to $4\times$ faster rendering.\@

\autoref{tab:results} and \autoref{fig:images} summarize our main results.
We report MAPE and samples per pixel after 2 hours of rendering at a resolution of \num{1920}x\num{1080}.
We compare unidirectional path tracing (PT), practical path guiding with recent improvements (PPG)~\citep{mueller2017practical,mueller19guiding}, \ADD{neural importance sampling~\citep{mueller2019nis} without~(NIS) and with (NIS++) our additionally proposed learning-rate decay and approximately normalized cross-entropy loss (\autoref{sec:composite-loss}, \autoref{eq:pdf-cross-entropy})}, and our neural control variates (NCV).
Russian roulette is \emph{not} used, which we elaborate on in \autoref{sec:discussion-and-future-work}.
Among these unbiased techniques, our NCVs usually yield the lowest error.

\paragraph{Comparison to NIS and NIS++}
To rule out the possibility of NCVs outperforming NIS simply because it uses additional neural networks for the CV components, we use only $L=2$ warps for importance sampling the residual integral (NIS uses $L=4$ warps).

Furthermore, we use a \emph{single} neural network to simultaneously predict the coefficient $\cvCoef$, the CV integral $\CV$, and the selection probability $\selectionProb$.\@
Our NCVs therefore use the same total number of neural networks (five), all with the same architecture, and the same total number of piecewise-quadratic warps (four) as NIS;\@ therefore the number of trainable parameters is the same and the performance is comparable.
Differences in the number of per-pixel samples are largely due to differences in importance sampling and therefore different mean path lengths.

\paragraph{Path termination using NCV}
We also show the results of applying our path termination heuristic (\autoref{sec:path-truncation}) as a by-product of NCVs.
The technique dramatically outperforms the unbiased algorithm at the cost of minimally visible artifacts (cf.\ the ``NCV + heuristic'' column in \autoref{fig:images}). Please refer to the supplementary material with an interactive image viewer for full-resolution images.

\begin{figure}
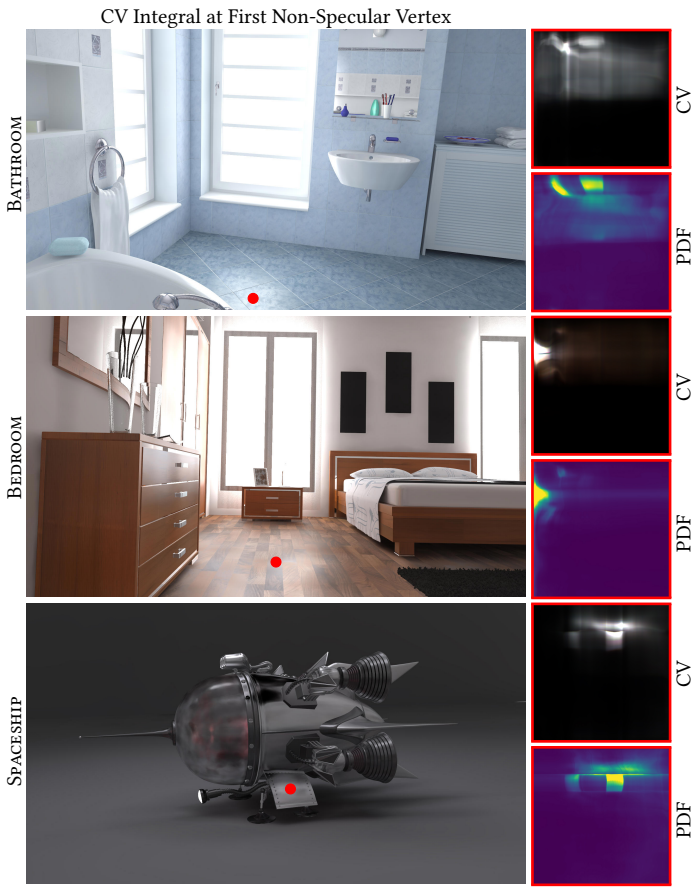

    \vspace{-1mm}
    \setlength{\fboxrule}{1pt}%
\setlength{\insetvsep}{20pt}%
\setlength{\tabcolsep}{1pt}%
\renewcommand{\arraystretch}{0.75}%
\small%
\hspace*{-3mm}%
\begin{tabular}{rccl}
& CV Integral at First Non-Specular Vertex & & \\
    & & \color{red}\fbox{\animategraphics[width=0.10006009615384616\textwidth,autoplay,loop,palindrome,every=3,alttext=none,poster=12]{20}{images/fig-light-field/cam-0-bathroom/downsampled-dist-0-cv-}{0}{149}} & \rotatebox{90}{\hspace{0.7cm}CV}\\
    \multirow[t]{2}{*}{\rotatebox{90}{\hspace{1.30cm}\Bathroom{}}} & 
    \multirow[t]{2}{*}{\begin{animateinline}[autoplay,loop,palindrome,alttext=none,poster=12]{20}\multiframe{50}{i=0+3}{\begin{tikzpicture}\useasboundingbox (0,0) rectangle (0.37\textwidth,0.208125\textwidth);
\node [inner sep=0pt] (pic) at (0.185\textwidth,0.1040625\textwidth) {\includegraphics[width=0.37\textwidth]{images/fig-light-field/cam-0-bathroom/downsampled-control-variate-\i.jpg}};
\draw[red,fill=red] (0.16784895833333335\textwidth,0.0083828125\textwidth) circle (0.0037\textwidth);
\end{tikzpicture}}\end{animateinline}} & 
    \color{red}\fbox{\animategraphics[width=0.10006009615384616\textwidth,autoplay,loop,palindrome,every=3,alttext=none,poster=12]{20}{images/fig-light-field/cam-0-bathroom/downsampled-dist-0-pdf-}{0}{149}} & \rotatebox{90}{\hspace{0.6cm}PDF}\\
    & & \color{red}\fbox{\animategraphics[width=0.10006009615384616\textwidth,autoplay,loop,palindrome,every=3,alttext=none,poster=12]{20}{images/fig-light-field/cam-0-bedroom/downsampled-dist-0-cv-}{0}{149}} & \rotatebox{90}{\hspace{0.7cm}CV}\\
    \multirow[t]{2}{*}{\rotatebox{90}{\hspace{1.30cm}\Bedroom{}}} & 
    \multirow[t]{2}{*}{\begin{animateinline}[autoplay,loop,palindrome,alttext=none,poster=12]{20}\multiframe{50}{i=0+3}{\begin{tikzpicture}\useasboundingbox (0,0) rectangle (0.37\textwidth,0.208125\textwidth);
\node [inner sep=0pt] (pic) at (0.185\textwidth,0.1040625\textwidth) {\includegraphics[width=0.37\textwidth]{images/fig-light-field/cam-0-bedroom/downsampled-control-variate-\i.jpg}};
\draw[red,fill=red] (0.18480729166666665\textwidth,0.025630208333333335\textwidth) circle (0.0037\textwidth);
\end{tikzpicture}}\end{animateinline}} & 
    \color{red}\fbox{\animategraphics[width=0.10006009615384616\textwidth,autoplay,loop,palindrome,every=3,alttext=none,poster=12]{20}{images/fig-light-field/cam-0-bedroom/downsampled-dist-0-pdf-}{0}{149}} & \rotatebox{90}{\hspace{0.6cm}PDF}\\
    & & \color{red}\fbox{\animategraphics[width=0.10006009615384616\textwidth,autoplay,loop,palindrome,every=3,alttext=none,poster=12]{20}{images/fig-light-field/cam-0-spaceship/downsampled-dist-0-cv-}{0}{149}} & \rotatebox{90}{\hspace{0.7cm}CV}\\
    \multirow[t]{2}{*}{\rotatebox{90}{\hspace{1.30cm}\Spaceship{}}} & 
    \multirow[t]{2}{*}{\begin{animateinline}[autoplay,loop,palindrome,alttext=none,poster=12]{20}\multiframe{50}{i=0+3}{\begin{tikzpicture}\useasboundingbox (0,0) rectangle (0.37\textwidth,0.208125\textwidth);
\node [inner sep=0pt] (pic) at (0.185\textwidth,0.1040625\textwidth) {\includegraphics[width=0.37\textwidth]{images/fig-light-field/cam-0-spaceship/downsampled-control-variate-\i.jpg}};
\draw[red,fill=red] (0.19553536458333332\textwidth,0.07014583333333332\textwidth) circle (0.0037\textwidth);
\end{tikzpicture}}\end{animateinline}} & 
    \color{red}\fbox{\animategraphics[width=0.10006009615384616\textwidth,autoplay,loop,palindrome,every=3,alttext=none,poster=12]{20}{images/fig-light-field/cam-0-spaceship/downsampled-dist-0-pdf-}{0}{149}} & \rotatebox{90}{\hspace{0.6cm}PDF}\\
\end{tabular}

    \vspace{-3mm}
    \caption{\label{fig:light-field}
        Visualization of the learned scattered light field (the CV integral $\CV$) from novel viewpoints in several scenes.
        The light field was trained while rendering the corresponding scenes in \autoref{fig:images}. As the shown camera views were not used for rendering, they were learned by our neural networks from secondary path vertices.
        We synthesize the visualizations by evaluating the learned light field for each pixel at the first non-specular path vertex (left).
        We also show the learned CV (top-right) and sampling PDF (bottom-right) for the given viewing direction at the marked locations (red).
    }
    \vspace{-2mm}
\end{figure}

\paragraph{Quality of approximations}
\autoref{tab:results} and \autoref{fig:images} include a column listing the MAPE obtained when evaluating the learned CV integral at the first non-specular path vertex.
Even though the CV integral exhibits visible bias, its relatively low MAPE is an indicator of the excellent approximation power of neural networks.

To further explore the limits of our parametric models, we visualize them \emph{from novel viewpoints} in \autoref{fig:light-field}.
For each scene in the figure, the neural networks were trained while rendering the corresponding entries in \autoref{tab:results}.
In the \Bedroom{} and \Bathroom{} scenes, the specular highlights on the floor and the furniture are at the correct positions. This observation supports the claim that the neural networks learn the actual 5D light field, as opposed to a mere screen-space approximation of it.
On the other hand, in the \Spaceship{} scene, the highly glossy transport is not accurately captured from the novel viewpoint, despite the good performance of our NCVs in terms of MAPE.\@
We show animated camera trajectories in the supplementary video, which also features the \Necklace{} scene as another failure case of incorrectly learned glossy light transport.

\begin{figure*}
    \input{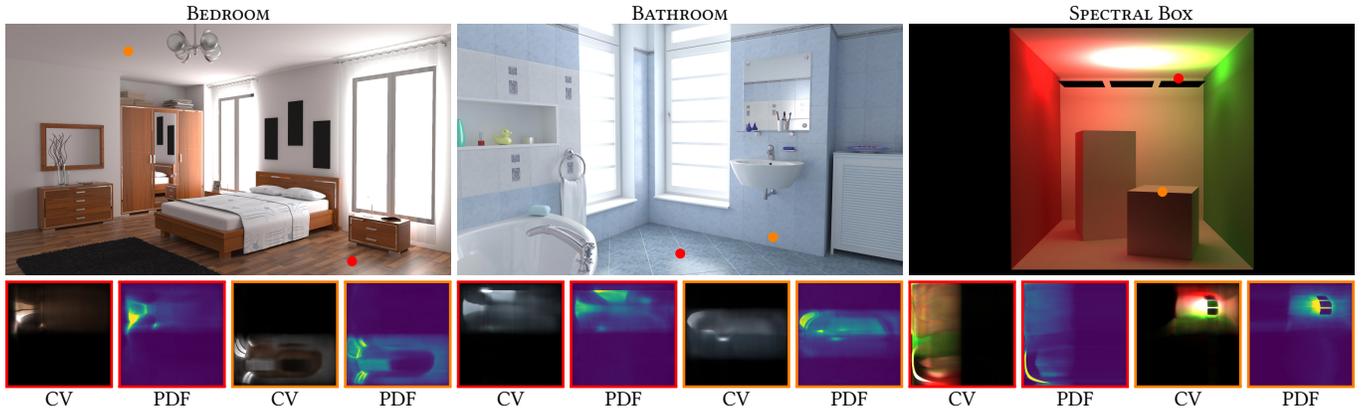}
    \vspace{-1mm}
    \caption{\label{fig:visualizer}
        Visualization of the learned control variates and importance sampling distributions in several scenes.
        We show CVs (left) and sampling PDFs (right) at two locations (red and orange) per scene.
        The CVs and PDFs are parametrized in world space via cylindrical coordinates as learned by our neural networks.
    }
\end{figure*}

Finally, we demonstrate the \emph{spatial} adaptivity of our model in \autoref{fig:visualizer}, where we show the learned CV and PDF at several locations in space.
Combined with the observations from \autoref{fig:light-field}, the spatial adaptivity confirms that the learned CV and PDF capture the full 7-dimensional integrand of the rendering equation with a reasonable accuracy.
We visualize the spatio-directional variations of the learned distributions in the supplementary video.

\begin{figure*}
    \setlength{\fboxrule}{1pt}%
\setlength{\insetvsep}{20pt}%
\setlength{\tabcolsep}{1pt}%
\renewcommand{\arraystretch}{0.75}%
\small%
\hspace*{-2.5mm}%
\begin{tabular}{rccccccl}
& & NCV & + heuristic & 1st vertex & 2nd vertex & 3rd vertex & \\
 & & \color{red}\fbox{\includegraphics[height=0.07964999999999998\textwidth]{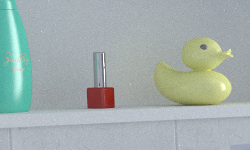}}
 & \color{red}\fbox{\includegraphics[height=0.07964999999999998\textwidth]{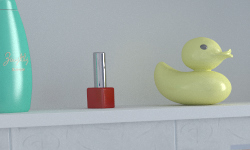}}
 & \color{red}\fbox{\includegraphics[height=0.07964999999999998\textwidth]{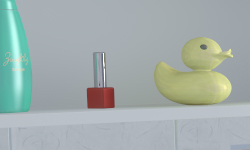}}
 & \color{red}\fbox{\includegraphics[height=0.07964999999999998\textwidth]{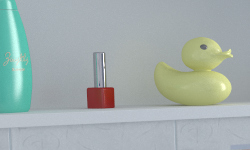}}
 & \color{red}\fbox{\includegraphics[height=0.07964999999999998\textwidth]{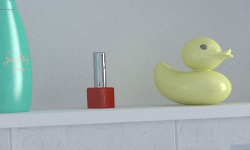}}
 & \rotatebox{90}{\hspace{0.3cm}Render} \\
   \multirow[t]{2}{*}{\rotatebox{90}{\hspace{0.79cm}\Bathroom{}}} 
 & \multirow[t]{2}{*}{\begin{tikzpicture}\useasboundingbox (0,0) rectangle (0.295\textwidth,0.1659375\textwidth);\node [inner sep=0pt] (pic) at (0.1475\textwidth,0.08296875\textwidth) {\includegraphics[width=0.295\textwidth]{images/fig-references/bathroom-reference-lq.jpg}};\draw[red,thick] (0.0004609375\textwidth,0.10217447916666665\textwidth) rectangle (0.03887239583333334\textwidth,0.07912760416666666\textwidth);\end{tikzpicture}}
 & \color{red}\fbox{\includegraphics[height=0.07964999999999998\textwidth]{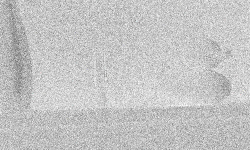}}
 & \color{red}\fbox{\includegraphics[height=0.07964999999999998\textwidth]{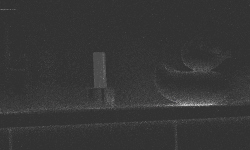}}
 & \color{red}\fbox{\includegraphics[height=0.07964999999999998\textwidth]{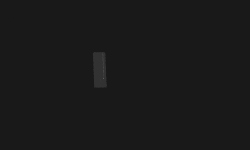}}
 & \color{red}\fbox{\includegraphics[height=0.07964999999999998\textwidth]{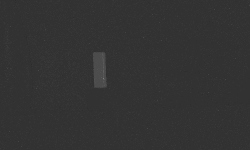}}
 & \color{red}\fbox{\includegraphics[height=0.07964999999999998\textwidth]{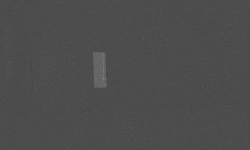}}
 &  \rotatebox{90}{\hspace{0.05cm}Path length} \\
 & \multicolumn{1}{r}{MAPE:} & 0.030 & \textbf{0.017} & 0.023 & 0.017 & 0.020 \\
 & & \color{red}\fbox{\includegraphics[height=0.07964999999999998\textwidth]{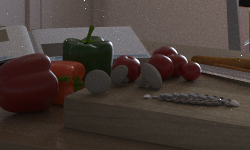}}
 & \color{red}\fbox{\includegraphics[height=0.07964999999999998\textwidth]{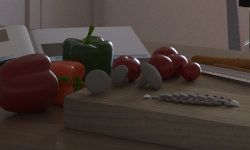}}
 & \color{red}\fbox{\includegraphics[height=0.07964999999999998\textwidth]{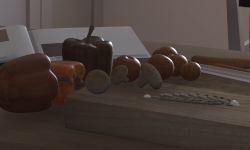}}
 & \color{red}\fbox{\includegraphics[height=0.07964999999999998\textwidth]{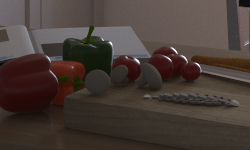}}
 & \color{red}\fbox{\includegraphics[height=0.07964999999999998\textwidth]{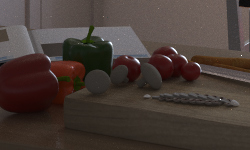}}
 & \rotatebox{90}{\hspace{0.3cm}Render} \\
   \multirow[t]{2}{*}{\rotatebox{90}{\hspace{0.39cm}\CountryKitchen{}}} 
 & \multirow[t]{2}{*}{\begin{tikzpicture}\useasboundingbox (0,0) rectangle (0.295\textwidth,0.1659375\textwidth);\node [inner sep=0pt] (pic) at (0.1475\textwidth,0.08296875\textwidth) {\includegraphics[width=0.295\textwidth]{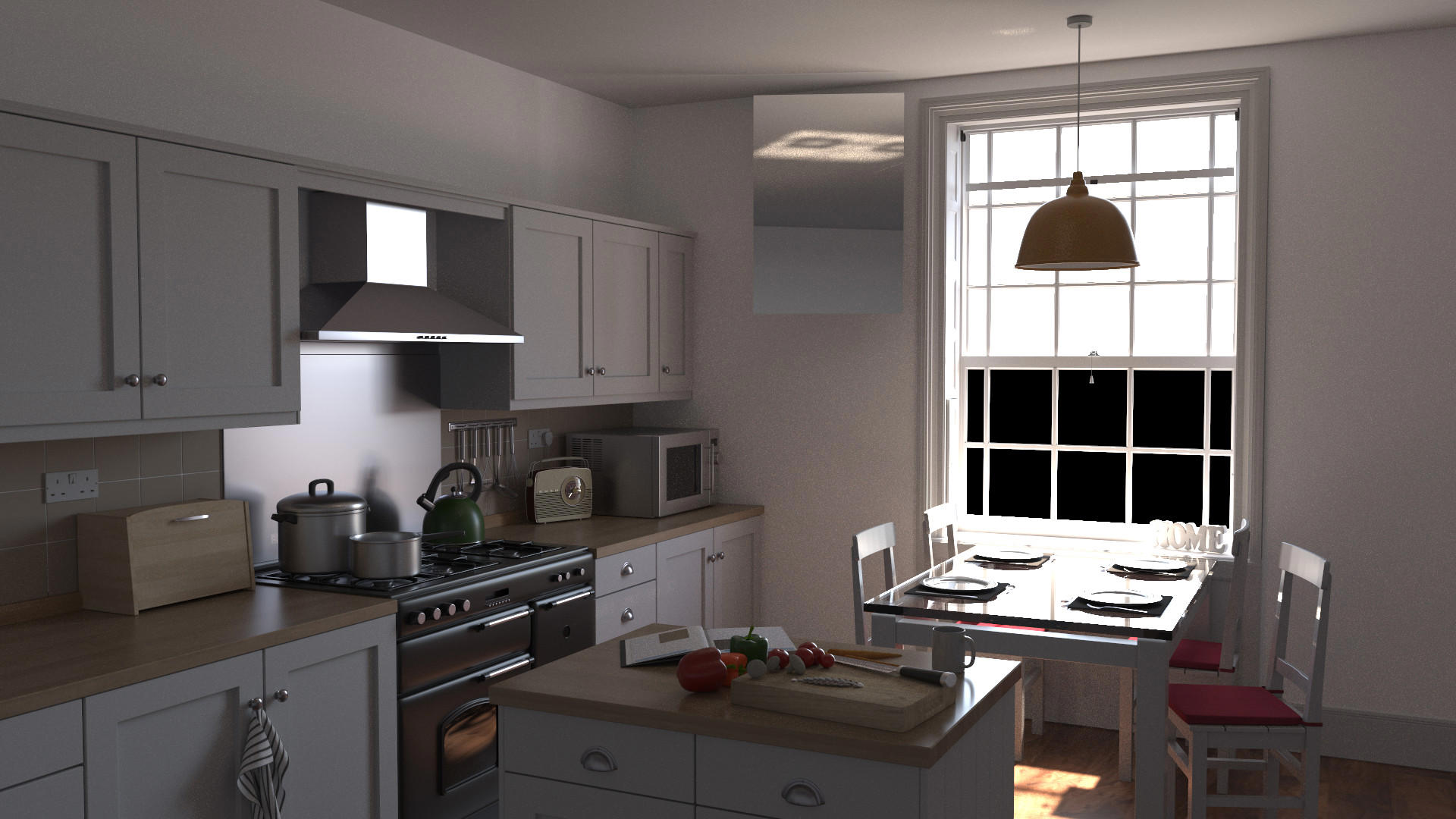}};\draw[red,thick] (0.13828125\textwidth,0.04302083333333333\textwidth) rectangle (0.17669270833333334\textwidth,0.01997395833333332\textwidth);\end{tikzpicture}}
 & \color{red}\fbox{\includegraphics[height=0.07964999999999998\textwidth]{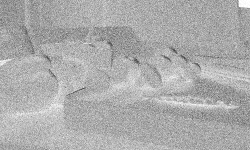}}
 & \color{red}\fbox{\includegraphics[height=0.07964999999999998\textwidth]{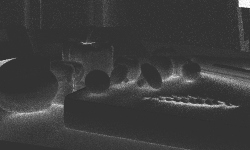}}
 & \color{red}\fbox{\includegraphics[height=0.07964999999999998\textwidth]{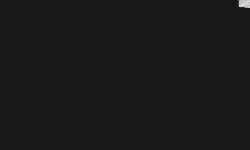}}
 & \color{red}\fbox{\includegraphics[height=0.07964999999999998\textwidth]{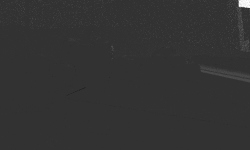}}
 & \color{red}\fbox{\includegraphics[height=0.07964999999999998\textwidth]{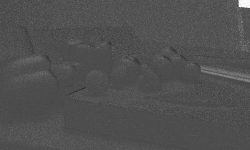}}
 &  \rotatebox{90}{\hspace{0.05cm}Path length} \\
 & \multicolumn{1}{r}{MAPE:} & 0.061 & \textbf{0.049} & 0.070 & 0.051 & 0.055 \\
\end{tabular}

    \vspace{-2mm}
    \caption{\label{fig:path-length}
        Different path termination strategies.
        We compare our unbiased NCVs (no early termination) to our heuristic (\autoref{sec:path-truncation}) and to terminating at the 1st, 2nd, or 3rd non-specular vertex.
        The top row shows the resulting images and the bottom row shows the average path length in each pixel (brighter means longer).
        The heuristic consistently produces less noise than always terminating at the 3rd vertex while simultaneously avoiding visible artifacts in creases and on rough surfaces (as seen when terminating always at the 1st or 2nd vertex).
    }
\end{figure*}

\begin{figure*}
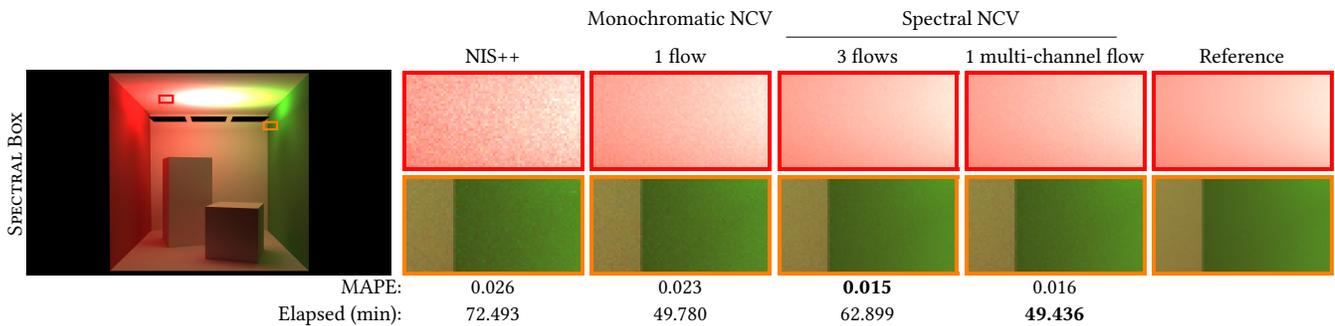

    \setlength{\fboxrule}{10pt}%
\setlength{\insetvsep}{20pt}%
\setlength{\tabcolsep}{-1pt}%
\renewcommand{\arraystretch}{1}%
\small%
\hspace*{-2mm}%
\begin{tabular}{rcccccc}
  {} & {} & {}  & Monochromatic NCV & \multicolumn{2}{c}{Spectral NCV} \\
  \cmidrule(lr{12pt}){5-6}
  {} & {} & NIS++ & 1 flow            & 3 flows & 1 multi-channel flow   & Reference \\
    \setInset{A}{red}{700}{135}{71}{38}%
    \setInset{B}{orange}{1250}{275}{71}{38}%
    \rotatebox{90}{\hspace{-1.95cm}\SpectralBox{}}\hspace{0.14cm} & 
    \addBeautyCrop{images/fig-references/spectral-box-reference-lq.jpg}{0.27}{1920}{1080}{0}{0}{1920}{1080} &
    \addInsets{images/fig-mono-spectral/spectral-box----.jpg} &
    \addInsets{images/fig-mono-spectral/spectral-box-1-flow.jpg} &
    \addInsets{images/fig-mono-spectral/spectral-box-3-flows.jpg} &
    \addInsets{images/fig-mono-spectral/spectral-box-1-multi-channel-flow.jpg} &
    \addInsets{images/fig-references/spectral-box-reference.jpg} \\
  & \multicolumn{1}{r}{MAPE:} & 0.026 & 0.023 & \textbf{0.015} & 0.016 \\
  & \multicolumn{1}{r}{Elapsed (min):} & 72.493 & 49.780 & 62.899 & \textbf{49.436} \\
\end{tabular}

    \vspace{-2mm}
    \caption{\label{fig:mono-vs-spectral}
        Comparison of monochromatic and spectral control variates at an equal sampling rate (512 spp).
        On scenes where the neural control variate is most useful, such as the \SpectralBox{} shown here, the 3-flows spectral approach can offer a significant improvement over the monochromatic one, but at a higher computational cost. Our multi-channel flow offers a good trade-off between the two approaches.
        The higher rendering time of NIS is due to the fact that it tends to yield longer paths on the average than the CV approaches in this scene. The impact could be smaller or even reversed in other scenes.
    }%
\end{figure*}

\begin{figure*}
    \vspace{-2mm}
    \setlength{\fboxrule}{10pt}%
\setlength{\insetvsep}{20pt}%
\setlength{\tabcolsep}{-1pt}%
\renewcommand{\arraystretch}{1}%
\small%
\hspace*{-2mm}%
\begin{tabular}{rcccccccc}
  {} & {} & 8 spp & 32 spp & 128 spp & 512 spp & 2048 spp & 8192 spp & Reference \\
    \setInset{A}{red}{750}{500}{153}{114}%
    \setInset{B}{orange}{440}{780}{153}{114}%
    \rotatebox{90}{\hspace{-2.05cm}\SwimmingPool{}}\hspace{0.14cm} &
    \addBeautyCrop{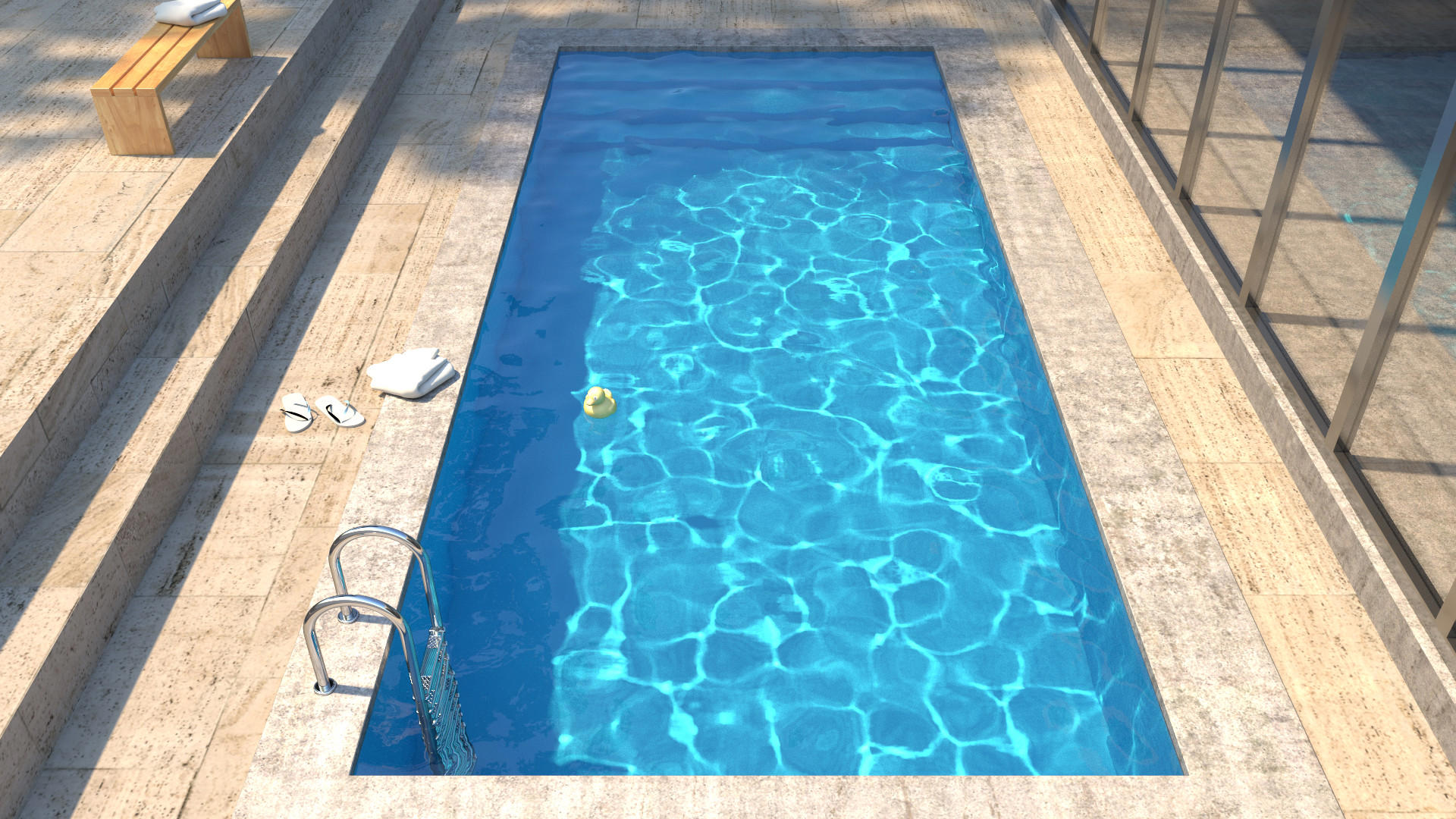}{0.27}{1920}{1080}{0}{0}{1920}{1080} &
    \addInsets{images/fig-cv-conv/pool-it2.jpg} &
    \addInsets{images/fig-cv-conv/pool-it4.jpg} &
    \addInsets{images/fig-cv-conv/pool-it6.jpg} &
    \addInsets{images/fig-cv-conv/pool-it8.jpg} &
    \addInsets{images/fig-cv-conv/pool-it10.jpg} &
    \addInsets{images/fig-cv-conv/pool-it12.jpg} &
    \addInsets{images/fig-references/pool-reference.jpg} \\
  & \multicolumn{1}{r}{MAPE:} & 0.375 & 0.192 & 0.139 & 0.108 & 0.068 & 0.042 \\
\end{tabular}

    \vspace{-2mm}
    \caption{\label{fig:cv-conv}
        CV integral training convergence.
        Training \ADD{a dedicated CV integral network} for longer than 2 hours enables more accurate approximations.
        This suggests that our method is not limited by the approximation power of the neural networks, but by the speed of their training.
    }
    \vspace{-2mm}
\end{figure*}

\paragraph{Convergence plots}
In \autoref{fig:convergence}, we plot MAPE vs.\ time (seconds) for PPG, NIS++, and our unbiased and biased NCV applications.
The unbiased NCVs (green line) are mostly on-par or slightly better than NIS++, except for the \Bookshelf{}, \CornellBox{}, \SpectralBox{}, \VeachDoor{}, and \VeachLamp{} scenes, where the difference is more pronounced.
Adding the heuristic path termination (dashed green line) significantly improves results in most scenes.
Interestingly, the almost noise-free CV integral, when applied at the first non-specular path vertex, initially performs much better than the other techniques in terms of MAPE.\@
However, as soon as a sufficient number of samples are drawn, the (nearly) unbiased techniques overtake the significantly biased learned CV integral in most scenes.

\section{Discussion and future work}%
\label{sec:discussion-and-future-work}

The use of neural networks makes our technique more computationally expensive than many non-neural approaches such as PPG (see \autoref{tab:results}) or Gaussian~\cite{Vorba:2014:OnlineLearningPMMinLTS} and von Mises-Fischer mixture models~\cite{Herholz:2019}.
It is worth emphasizing that the different methods in the equal-time comparisons do not all use the same hardware: PPG utilizes only the CPU while NIS and NCV additionally leverage two GPUs and require roughly $3\times$ more power.
Nonetheless, even in a light-weight path tracer such as Mitsuba, the per-sample variance reduction of NCV sometimes outweighs the added computational cost and higher power consumption.
In cases with high tracing and shading costs---such as in production rendering---the relative overhead of the networks will be smaller and a larger efficiency attainable.
The practicality of NCV thus strongly depends on the use case and availability of hardware acceleration.

When the small amount of bias from our path termination heuristic is acceptable, the average path length is reduced by roughly \num{3}$\times$ (see \autoref{fig:path-length}).
Efficiency is therefore improved twofold:\ lower sample variance \emph{and} much cheaper paths.
Developing neural estimators that simulate all light bounces with short paths only, e.g.\ via Q-learning~\cite{Dahm16}, may enable synthesizing high-quality images at interactive and real-time rates in the future.
Leveraging samples across time may also become an enabler.

\paragraph{Separate learning of $\cvCoef$ and $\CV$}
Recall that the control variate is defined as the product $\acv = \cvCoef \cdot \CV \cdot \cvShape$.
Since \emph{both} $\cvCoef$ and $\CV$ are scaling factors of the control variate, it is possible to combine and learn them as a single variance-minimizing factor $\aCV$.
The reason for keeping them separate, in our case, is that a variance-minimizing $\aCV$ may not necessarily approximate $F$ well. However, we need a good approximation $\CV \approx F$ for two reasons: (i) to normalize the relative loss terms by $\CV^2$ and (ii) to approximate the path-tail contribution by $\CV$ when heuristically terminating the paths.

\paragraph{Benefit of unbiased control variates}
Our unbiased control variates improve the efficiency over pure NIS by \num{6}--\num{17}$\times$ on a toy problem (\autoref{fig:toy}) whereas they yield a much smaller benefit in light-transport simulations (\autoref{tab:results} and \autoref{fig:mono-vs-spectral}).
We suspect that this discrepancy arises from the respective difficulties of the integration problems.
In the 2D toy problems, our spectral control variate comes close to a zero-variance configuration, yielding much greater efficiency than pure NIS, which is limited to learning a monochromatic function.
In the light-transport simulation, the integrand is higher dimensional (7D) and less well behaved, leading to a larger distance between our models and the zero-variance configuration.
This hypothesis is supported by the fact that simpler scenes, such as the \SpectralBox{}, benefit more from our control variates.
Importantly, \autoref{fig:mono-vs-spectral} demonstrates that despite the smaller efficiency improvements in the light-transport simulation, unbiased spectral control variates still have a fundamental advantage over monochromatic techniques.
Furthermore, the spectrally learned integral of the control variate (see \autoref{sec:path-truncation}) results in an additional significant efficiency gain.

\paragraph{Approximation power}
\autoref{fig:cv-conv} shows that optimizing the model of the CV integral using \num{8192}spp (all other figures use much fewer spp) enables accurately approximating intricate, high-frequency signals.
This suggests that our method is not limited by the approximation power of the employed neural networks, but by the rate of learning.
We confirmed this experimentally by increasing the number of sub-flows and by making the neural network bigger. Both resulted in only minor variance reduction---much less than the added computational overhead.
For future work, it is therefore of particular interest to investigate means of increasing the training efficiency of the model.

\paragraph{Animated sequences}
\autoref{fig:light-field} demonstrates that the neural networks learn a reasonable approximation of the full light field---not just the slice visible by the camera.
When rendering an animation, it would thus be reasonable to bootstrap the training of the next frame by initializing the networks using the weights resulting from rendering the previous frame (as opposed to randomly).

\subsection{Extensions}

\paragraph{Handling of signed integrands}%
\label{sec:handling-of-signed-integrands}

In \autoref{sec:model}, we point out that our parametric neural control variates are non-negative by construction.
Inspired by Owen and Zhou's~\shortcite{owen00safe} positivisation trick for importance sampling, we show that one can construct an arbitrary \emph{signed} control variate from two such non-negative control variates.

Let $f(x)$ be a signed integrand.
Then the first non-negative control variate $\cv^+(x)$ shall approximate the positive portion of $f(x)$ and the second non-negative control variate $\cv^-(x)$ shall approximate the negative portion of $f(x)$.
Formally:
\begin{align}
    \cv^+(x) &\approx \max(f(x), 0) \\
    \cv^-(x) &\approx -\min(f(x), 0) \,.
\end{align}
The signed control variate $\cv(x) := \cv^+(x) - \cv^-(x)$ is defined as the difference of the two control variates.
It follows that $\cv(x)$ approximates $f(x)$ as desired:
\begin{align}
    \cv(x) \equiv \cv^+(x) -\cv^-(x)
    \approx \max(f(x), 0) + \min(f(x), 0) = f(x) \,.
\end{align}

\paragraph{Adjoint Russian roulette and splitting}
Many path tracers utilize Russian roulette to terminate paths and thereby probabilistically avoid evaluating low-contribution samples.
Since the neural light field approximation at the primary vertex already contains the majority of the contribution to the pixel value with only little error to be corrected by the remaining path tail, Russian roulette should be aggressively truncating the paths, leading to much greater efficiency.
Unfortunately, in our experiments, Russian roulette instead \emph{worsened} the efficiency of our approach.
We suspect that this is because it does not take the \emph{variance} of the error correction into account, which is a crucial component of efficiency-optimized Russian roulette~\citep{Veach:1997:Thesis}.
Orthogonally, it is possible to improve the accuracy of Russian roulette by taking the adjoint into account~\citep{VorbaKrivanek:2016:ADRRS}, however it is difficult to obtain a good estimate of the adjoint.
This is because our renderer is not estimating radiance, but the difference to our CV.\@
The appropriate adjoint in this case is the difference integral, which is not readily available.
We leave the investigation of efficiency-optimized Russian roulette within our parametric control variates as future work.

\paragraph{Generalization to volume rendering}
Even though we only demonstrated results for surface rendering, we believe our algorithm generalizes to volumetric rendering.
First, for estimating in-scattered radiance at a point, the only part of our algorithm that needs to change is the network input listed in \autoref{tab:parameters}.
It is straightforward to add relevant properties of volumetric points to the network input.
Then, for estimating radiance along a line through a volume, one may either use an additional 1-D instance of our model, or one may use a form of guided distance sampling~\citep{Herholz:2019}, complemented by the neural light field $\CV$ to approximate the true in-scattered radiance.

\section{Conclusion}

We present neural control variates, a model for reducing variance in parametric Monte Carlo integration.
The main challenge that we tackle is designing a model with sufficient approximation power that is efficient to evaluate.
We achieve this by employing normalizing flows to model the shape of the control variate and a second neural network to infer its integral.
To this end, prior works on normalizing flows are extended by developing the multi-channel normalizing flows, which improve the performance of multi-channel integration such as spectral rendering.
To further reduce the integration error, we utilize neural importance sampling for estimating the correction term.
We describe recipes for jointly optimizing the NCVs and the residual NIS using (i) a theoretically optimal variance loss, and (ii) an empirical composite loss for robust optimization.

We analyzed the performance of neural control variates in the setting of photorealistic image synthesis.
The NCVs yield notable improvements and perform better, on the average, than state-of-the-art competitors in both equal-time and equal-sample-count settings.
While our unbiased application of NCVs only provides a small efficiency boost, it enables a biased algorithm that improves efficiency significantly.
We expect our performance to grow further as the considerable cost of neural networks decreases with the advance of models and computer hardware.
While we demonstrate the utility of our approach for path tracing, we think it will extend well to many rendering algorithms that employ Monte Carlo integration, and we expect it to be applicable beyond light-transport algorithms due to its fundamental nature.

\paragraph{Convergence of data-driven and physically based simulation}
Our work connects neural approximation and unbiased simulation.
While data-driven approaches can obtain realistic results, correcting their errors is tedious. Physically based integrators, on the other hand, provide accurate solutions, albeit at excessive cost.
We show that the mechanism of control variates allows for combining a data-driven, high-quality neural approximation with an accurate, physically-based integrator, which can be used on demand to merely correct the errors.
The challenge for such future developments will be defining $\CV$ and $\cv$ such that $\CV=\int \cv(x)\Diff{x}$ is preserved while both are efficient to evaluate.

\begin{acks}
We thank Markus Kettunen and Nikolaus Binder for valuable feedback.
We also thank the following people for providing scenes and models that appear in our figures:
Benedikt \citet{resources16},
Frank Meinl (\CrytekSponza),
Jay-Artist (\CountryKitchen),
Johannes Hanika (\Necklace),
Marko Dabrovi\'c (\Sponza),
Miika Aittala, Samuli Laine, and Jaakko Lehtinen (\VeachDoor),
Olesya Jakob (\Torus),
Ond\v{r}ej Karl\'{\i}k (\SwimmingPool),
SlykDrako (\Bedroom),
thecali (\Spaceship),
Tiziano Portenier (\Bathroom, \Bookshelf), and
Wig42 (\Staircase).
\end{acks}

\appendix

\section{Variance of the Control Variate Estimator}%
\label{sec:variance-derivation}

In order to derive the variance of the control variate estimator
\begin{align}
    \MC{F} = \CV(\aCVParams) + \frac{f(X,Z)}{\PdfMC(X,Z;\pdfParams)}  - \frac{\acv(X;\acvParams)}{\PdfMC(X;\pdfParams)} \,,
    \label{eq:mc-integrator-with-z-appendix}
\end{align}
we recall that $\Variance[X + c] = \Variance[X]$ for any constant $c$. Hence, subtracting the constant $\aCV(\aCVParams)$ on both sides and proceeding with the definition of variance, we have:
\begin{align}
    \Variance\left[\MC{F}\right] &=
    \Variance\left[\MC{F} - \aCV(\aCVParams)\right] \nonumber \\
    &= \underbrace{\Expectation\left[ {\big({\MC{F} - \aCV(\aCVParams)}\big)}^2 \right]}_{=: \, U} - \underbrace{{\Expectation\left[ \MC{F} - \aCV(\aCVParams) \right]}^2}_{=: \, V}
    \,.
    \label{eq:mc-variance-derivation-appendix}
\end{align}
Then the expectation of the square results in the following double integral:
\begin{align}
    \lefteqn{\Expectation\left[ {\left({\MC{F} - \aCV(\aCVParams)}\right)}^2 \right]} \nonumber \\
    &=
    \Expectation\left[ {\left(
        \frac{f(X,Z)}{\PdfMC(X,Z;\pdfParams)}  - \frac{\acv(X;\acvParams)}{\PdfMC(X;\pdfParams)}
    \right)}^2 \right] \nonumber \\
    &= \int_\Domain  \int_\DomainZ  {\left( \frac{f(x,z)}{\PdfMC(x,z;\pdfParams)} - \frac{\acv(x;\acvParams)}{\PdfMC(x;\pdfParams)}\right)}^2
     \PdfMC(x,z;\pdfParams) \Diff{z}\Diff{x}
    \nonumber \\
    &= \int_\Domain  \int_\DomainZ  {\left( \frac{f(x,z)}{\PdfMC(x;\pdfParams)\PdfMC(z|x)} - \frac{\acv(x;\acvParams)}{\PdfMC(x;\pdfParams)}\right)}^2
     \PdfMC(x;\pdfParams)\PdfMC(z|x) \Diff{z}\Diff{x}
    \nonumber \\
    &= \int_\Domain  \int_\DomainZ  {\left( \frac{f(x,z)}{\PdfMC(z|x)} - \acv(x;\acvParams)\right)}^2\frac{\PdfMC(z|x)}{\PdfMC(x;\pdfParams)} \Diff{z}\Diff{x} = U\,.
\label{eq:mc-expectation-of-square-appedix}
\end{align}
The squared expectation (second term in \autoref{eq:mc-variance-derivation-appendix}) simplifies to
\begin{align}
    {\big(\Expectation\left[\MC{F}\right] - \aCV(\aCVParams) \big)}^2 = {\big(F - \aCV(\aCVParams) \big)}^2 = V\,.
    \label{eq:mc-square-of-expectation-appedix}
\end{align}
Putting the $U$ and $V$ terms together yields \autoref{eq:mc-variance-with-tail}.

\bibliography{bibliography}


\begin{thebibliography}{79}


\ifx \showCODEN    \undefined \def \showCODEN     #1{\unskip}     \fi
\ifx \showDOI      \undefined \def \showDOI       #1{#1}\fi
\ifx \showISBNx    \undefined \def \showISBNx     #1{\unskip}     \fi
\ifx \showISBNxiii \undefined \def \showISBNxiii  #1{\unskip}     \fi
\ifx \showISSN     \undefined \def \showISSN      #1{\unskip}     \fi
\ifx \showLCCN     \undefined \def \showLCCN      #1{\unskip}     \fi
\ifx \shownote     \undefined \def \shownote      #1{#1}          \fi
\ifx \showarticletitle \undefined \def \showarticletitle #1{#1}   \fi
\ifx \showURL      \undefined \def \showURL       {\relax}        \fi
\providecommand\bibfield[2]{#2}
\providecommand\bibinfo[2]{#2}
\providecommand\natexlab[1]{#1}
\providecommand\showeprint[2][]{arXiv:#2}

\bibitem[\protect\citeauthoryear{Abadi, Agarwal, Barham, Brevdo, Chen, Citro,
  Corrado, Davis, Dean, et~al\mbox{.}}{Abadi et~al\mbox{.}}{2015}]%
        {tensorflow2015-whitepaper}
\bibfield{author}{\bibinfo{person}{Mart\'{\i}n Abadi}, \bibinfo{person}{Ashish
  Agarwal}, \bibinfo{person}{Paul Barham}, \bibinfo{person}{Eugene Brevdo},
  \bibinfo{person}{Zhifeng Chen}, \bibinfo{person}{Craig Citro},
  \bibinfo{person}{Greg~S. Corrado}, \bibinfo{person}{Andy Davis},
  \bibinfo{person}{Jeffrey Dean}, {et~al\mbox{.}}}
  \bibinfo{year}{2015}\natexlab{}.
\newblock \bibinfo{title}{{TensorFlow}: Large-Scale Machine Learning on
  Heterogeneous Systems}.
\newblock
\newblock
\urldef\tempurl%
\url{http://tensorflow.org/}
\showURL{%
\tempurl}


\bibitem[\protect\citeauthoryear{Assaraf and Caffarel}{Assaraf and
  Caffarel}{1999}]%
        {PhysRevLett.83.4682}
\bibfield{author}{\bibinfo{person}{Roland Assaraf} {and}
  \bibinfo{person}{Michel Caffarel}.} \bibinfo{year}{1999}\natexlab{}.
\newblock \showarticletitle{Zero-Variance Principle for Monte Carlo
  Algorithms}.
\newblock \bibinfo{journal}{\emph{Phys. Rev. Lett.}}  \bibinfo{volume}{83}
  (\bibinfo{date}{Dec} \bibinfo{year}{1999}), \bibinfo{pages}{4682--4685}.
\newblock
Issue 23.
\urldef\tempurl%
\url{https://doi.org/10.1103/PhysRevLett.83.4682}
\showDOI{\tempurl}


\bibitem[\protect\citeauthoryear{Barth, Schwab, and Zollinger}{Barth
  et~al\mbox{.}}{2011}]%
        {Barth:2011}
\bibfield{author}{\bibinfo{person}{Andrea Barth}, \bibinfo{person}{Christoph
  Schwab}, {and} \bibinfo{person}{Nathaniel Zollinger}.}
  \bibinfo{year}{2011}\natexlab{}.
\newblock \showarticletitle{Multi-level Monte Carlo Finite Element method for
  elliptic PDEs with Stochastic Coefficients}.
\newblock \bibinfo{journal}{\emph{Numer. Math.}} \bibinfo{volume}{119},
  \bibinfo{number}{1} (\bibinfo{year}{2011}), \bibinfo{pages}{123--161}.
\newblock
\showISSN{0945-3245}
\urldef\tempurl%
\url{https://doi.org/10.1007/s00211-011-0377-0}
\showDOI{\tempurl}


\bibitem[\protect\citeauthoryear{Bekaert, Slusallek, Cools, Havran, and
  Seidel}{Bekaert et~al\mbox{.}}{2003}]%
        {bekaert2003}
\bibfield{author}{\bibinfo{person}{Philippe Bekaert}, \bibinfo{person}{Philipp
  Slusallek}, \bibinfo{person}{Ronald Cools}, \bibinfo{person}{Vlastimil
  Havran}, {and} \bibinfo{person}{Hans-Peter Seidel}.}
  \bibinfo{year}{2003}\natexlab{}.
\newblock \showarticletitle{A custom designed Density Estimation Method for
  Light Transport}.
\newblock \bibinfo{journal}{\emph{MPI-I-2003-4-004}} (\bibinfo{date}{April}
  \bibinfo{year}{2003}).
\newblock


\bibitem[\protect\citeauthoryear{Belcour, Soler, Subr, Holzschuch, and
  Durand}{Belcour et~al\mbox{.}}{2013}]%
        {Belcour:2013:COV}
\bibfield{author}{\bibinfo{person}{Laurent Belcour}, \bibinfo{person}{Cyril
  Soler}, \bibinfo{person}{Kartic Subr}, \bibinfo{person}{Nicolas Holzschuch},
  {and} \bibinfo{person}{Fredo Durand}.} \bibinfo{year}{2013}\natexlab{}.
\newblock \showarticletitle{5D Covariance Tracing for Efficient Defocus and
  Motion Blur}.
\newblock \bibinfo{journal}{\emph{ACM Trans. Graph.}} \bibinfo{volume}{32},
  \bibinfo{number}{3}, Article \bibinfo{articleno}{Article 31}
  (\bibinfo{date}{July} \bibinfo{year}{2013}), \bibinfo{numpages}{18}~pages.
\newblock
\showISSN{0730-0301}
\urldef\tempurl%
\url{https://doi.org/10.1145/2487228.2487239}
\showDOI{\tempurl}


\bibitem[\protect\citeauthoryear{Bitterli}{Bitterli}{2016}]%
        {resources16}
\bibfield{author}{\bibinfo{person}{Benedikt Bitterli}.}
  \bibinfo{year}{2016}\natexlab{}.
\newblock \bibinfo{title}{Rendering resources}.
\newblock
\newblock
\newblock
\shownote{https://benedikt-bitterli.me/resources/.}


\bibitem[\protect\citeauthoryear{Broadie and Glasserman}{Broadie and
  Glasserman}{1998}]%
        {Broadie:1998}
\bibfield{author}{\bibinfo{person}{Mark Broadie} {and} \bibinfo{person}{Paul
  Glasserman}.} \bibinfo{year}{1998}\natexlab{}.
\newblock \bibinfo{booktitle}{\emph{Risk Management and Analysis, Volume 1:
  Measuring and Modelling Financial Risk}}.
\newblock \bibinfo{publisher}{Wiley}, \bibinfo{address}{New York}, Chapter
  Simulation for option pricing and risk management, \bibinfo{pages}{173--208}.
\newblock


\bibitem[\protect\citeauthoryear{Chen, Rubanova, Bettencourt, and
  Duvenaud}{Chen et~al\mbox{.}}{2018}]%
        {chen2018neural}
\bibfield{author}{\bibinfo{person}{Tian~Qi Chen}, \bibinfo{person}{Yulia
  Rubanova}, \bibinfo{person}{Jesse Bettencourt}, {and} \bibinfo{person}{David
  Duvenaud}.} \bibinfo{year}{2018}\natexlab{}.
\newblock \showarticletitle{Neural Ordinary Differential Equations}.
\newblock \bibinfo{journal}{\emph{arXiv:1806.07366}} (\bibinfo{date}{June}
  \bibinfo{year}{2018}).
\newblock


\bibitem[\protect\citeauthoryear{Clarberg and Akenine-M{\"o}ller}{Clarberg and
  Akenine-M{\"o}ller}{2008}]%
        {Clarberg:2008}
\bibfield{author}{\bibinfo{person}{Petrik Clarberg} {and}
  \bibinfo{person}{Tomas Akenine-M{\"o}ller}.} \bibinfo{year}{2008}\natexlab{}.
\newblock \showarticletitle{Exploiting Visibility Correlation in Direct
  Illumination}.
\newblock \bibinfo{journal}{\emph{{Computer Graphics Forum}}}
  \bibinfo{volume}{27}, \bibinfo{number}{4} (\bibinfo{year}{2008}),
  \bibinfo{pages}{1125--1136}.
\newblock
\showISSN{1467-8659}
\urldef\tempurl%
\url{https://doi.org/10.1111/j.1467-8659.2008.01250.x}
\showDOI{\tempurl}


\bibitem[\protect\citeauthoryear{Dahm and Keller}{Dahm and Keller}{2018}]%
        {Dahm16}
\bibfield{author}{\bibinfo{person}{Ken Dahm} {and} \bibinfo{person}{Alexander
  Keller}.} \bibinfo{year}{2018}\natexlab{}.
\newblock \showarticletitle{Learning Light Transport the Reinforced Way}. In
  \bibinfo{booktitle}{\emph{Monte Carlo and Quasi-Monte Carlo Methods}},
  \bibfield{editor}{\bibinfo{person}{Art~B. Owen} {and}
  \bibinfo{person}{Peter~W. Glynn}} (Eds.). \bibinfo{publisher}{Springer
  International Publishing}, \bibinfo{pages}{181--195}.
\newblock
\showISBNx{978-3-319-91436-7}


\bibitem[\protect\citeauthoryear{Dinh, Krueger, and Bengio}{Dinh
  et~al\mbox{.}}{2014}]%
        {dinh2014nice}
\bibfield{author}{\bibinfo{person}{Laurent Dinh}, \bibinfo{person}{David
  Krueger}, {and} \bibinfo{person}{Yoshua Bengio}.}
  \bibinfo{year}{2014}\natexlab{}.
\newblock \showarticletitle{NICE: Non-linear Independent Components
  Estimation}.
\newblock \bibinfo{journal}{\emph{arXiv:1410.8516}} (\bibinfo{date}{Oct.}
  \bibinfo{year}{2014}).
\newblock


\bibitem[\protect\citeauthoryear{Dinh, Sohl-Dickstein, and Bengio}{Dinh
  et~al\mbox{.}}{2016}]%
        {dinh2016density}
\bibfield{author}{\bibinfo{person}{Laurent Dinh}, \bibinfo{person}{Jascha
  Sohl-Dickstein}, {and} \bibinfo{person}{Samy Bengio}.}
  \bibinfo{year}{2016}\natexlab{}.
\newblock \showarticletitle{Density Estimation using Real NVP}.
\newblock \bibinfo{journal}{\emph{arXiv:1605.08803}} (\bibinfo{date}{March}
  \bibinfo{year}{2016}).
\newblock


\bibitem[\protect\citeauthoryear{Fan, Chenney, Hu, Tsui, and Lai}{Fan
  et~al\mbox{.}}{2006}]%
        {Fan:2006:OCV}
\bibfield{author}{\bibinfo{person}{Shaohua Fan}, \bibinfo{person}{Stephen
  Chenney}, \bibinfo{person}{Bo Hu}, \bibinfo{person}{Kam-Wah Tsui}, {and}
  \bibinfo{person}{Yu-Chi Lai}.} \bibinfo{year}{2006}\natexlab{}.
\newblock \showarticletitle{Optimizing Control Variate Estimators for
  Rendering}.
\newblock \bibinfo{journal}{\emph{Computer Graphics Forum}}
  \bibinfo{volume}{25}, \bibinfo{number}{3} (\bibinfo{year}{2006}),
  \bibinfo{pages}{351--358}.
\newblock


\bibitem[\protect\citeauthoryear{Georgiev, Misso, Hachisuka, Nowrouzezahrai,
  Křivánek, and Jarosz}{Georgiev et~al\mbox{.}}{2019}]%
        {georgiev19integral}
\bibfield{author}{\bibinfo{person}{Iliyan Georgiev}, \bibinfo{person}{Zackary
  Misso}, \bibinfo{person}{Toshiya Hachisuka}, \bibinfo{person}{Derek
  Nowrouzezahrai}, \bibinfo{person}{Jaroslav Křivánek}, {and}
  \bibinfo{person}{Wojciech Jarosz}.} \bibinfo{year}{2019}\natexlab{}.
\newblock \showarticletitle{Integral formulations of volumetric transmittance}.
\newblock \bibinfo{journal}{\emph{ACM Transactions on Graphics (Proceedings of
  SIGGRAPH Asia)}} \bibinfo{volume}{38}, \bibinfo{number}{6}
  (\bibinfo{date}{Nov.} \bibinfo{year}{2019}).
\newblock
\urldef\tempurl%
\url{https://doi.org/10/dffn}
\showDOI{\tempurl}


\bibitem[\protect\citeauthoryear{Germain, Gregor, Murray, and
  Larochelle}{Germain et~al\mbox{.}}{2015}]%
        {germain2015made}
\bibfield{author}{\bibinfo{person}{Mathieu Germain}, \bibinfo{person}{Karol
  Gregor}, \bibinfo{person}{Iain Murray}, {and} \bibinfo{person}{Hugo
  Larochelle}.} \bibinfo{year}{2015}\natexlab{}.
\newblock \showarticletitle{MADE: Masked Autoencoder for Distribution
  Estimation}. In \bibinfo{booktitle}{\emph{International Conference on Machine
  Learning}}. \bibinfo{pages}{881--889}.
\newblock


\bibitem[\protect\citeauthoryear{Giles}{Giles}{2008}]%
        {Giles:2008}
\bibfield{author}{\bibinfo{person}{Michael~B. Giles}.}
  \bibinfo{year}{2008}\natexlab{}.
\newblock \showarticletitle{Improved Multilevel Monte Carlo Convergence using
  the Milstein Scheme}.
\newblock In \bibinfo{booktitle}{\emph{Monte Carlo and Quasi-Monte Carlo
  Methods 2006}}, \bibfield{editor}{\bibinfo{person}{Alexander Keller},
  \bibinfo{person}{Stefan Heinrich}, {and} \bibinfo{person}{Harald
  Niederreiter}} (Eds.). \bibinfo{publisher}{Springer},
  \bibinfo{address}{Berlin, Heidelberg}, \bibinfo{pages}{343--358}.
\newblock
\showISBNx{978-3-540-74496-2}
\urldef\tempurl%
\url{https://doi.org/10.1007/978-3-540-74496-2_20}
\showDOI{\tempurl}


\bibitem[\protect\citeauthoryear{Giles}{Giles}{2013}]%
        {giles:2013}
\bibfield{author}{\bibinfo{person}{Michael~B. Giles}.}
  \bibinfo{year}{2013}\natexlab{}.
\newblock \showarticletitle{Multilevel {M}onte {C}arlo Methods}.
\newblock In \bibinfo{booktitle}{\emph{{M}onte {C}arlo and {Q}uasi-{M}onte
  {C}arlo Methods 2012}}, \bibfield{editor}{\bibinfo{person}{Josef Dick},
  \bibinfo{person}{Y.~Frances Kuo}, \bibinfo{person}{W.~Gareth Peters}, {and}
  \bibinfo{person}{H.~Ian Sloan}} (Eds.). \bibinfo{publisher}{Springer},
  \bibinfo{address}{Berlin, Heidelberg}, \bibinfo{pages}{83--103}.
\newblock
\showISBNx{978-3-642-41095-6}
\urldef\tempurl%
\url{https://doi.org/10.1007/978-3-642-41095-6_4}
\showDOI{\tempurl}


\bibitem[\protect\citeauthoryear{Glorot and Bengio}{Glorot and Bengio}{2010}]%
        {Glorot:2010}
\bibfield{author}{\bibinfo{person}{Xavier Glorot} {and} \bibinfo{person}{Yoshua
  Bengio}.} \bibinfo{year}{2010}\natexlab{}.
\newblock \showarticletitle{Understanding the Difficulty of Training Deep
  Feedforward Neural Networks}. In \bibinfo{booktitle}{\emph{Proc.\ 13th
  International Conference on Artificial Intelligence and Statistics}} (May
  13--15). \bibinfo{publisher}{JMLR.org}, \bibinfo{pages}{249--256}.
\newblock


\bibitem[\protect\citeauthoryear{Glynn and Szechtman}{Glynn and
  Szechtman}{2002}]%
        {Glynn:2002}
\bibfield{author}{\bibinfo{person}{Peter~W. Glynn} {and}
  \bibinfo{person}{Roberto Szechtman}.} \bibinfo{year}{2002}\natexlab{}.
\newblock \showarticletitle{Some New Perspectives on the Method of Control
  Variates}.
\newblock In \bibinfo{booktitle}{\emph{Monte Carlo and Quasi-Monte Carlo
  Methods 2000}}, \bibfield{editor}{\bibinfo{person}{Kai-Tai Fang},
  \bibinfo{person}{Harald Niederreiter}, {and} \bibinfo{person}{Fred~J.
  Hickernell}} (Eds.). \bibinfo{publisher}{Springer}, \bibinfo{address}{Berlin,
  Heidelberg}, \bibinfo{pages}{27--49}.
\newblock
\showISBNx{978-3-642-56046-0}
\urldef\tempurl%
\url{https://doi.org/10.1007/978-3-642-56046-0_3}
\showDOI{\tempurl}


\bibitem[\protect\citeauthoryear{Grathwohl, Choi, Wu, Roeder, and
  Duvenaud}{Grathwohl et~al\mbox{.}}{2018}]%
        {Grathwohletal2018relax}
\bibfield{author}{\bibinfo{person}{Will Grathwohl}, \bibinfo{person}{Dami
  Choi}, \bibinfo{person}{Yuhuai Wu}, \bibinfo{person}{Geoff Roeder}, {and}
  \bibinfo{person}{David Duvenaud}.} \bibinfo{year}{2018}\natexlab{}.
\newblock \showarticletitle{Backpropagation through the Void: Optimizing
  control variates for black-box gradient estimation}.
\newblock \bibinfo{journal}{\emph{International Conference on Learning
  Representations}}.
\newblock


\bibitem[\protect\citeauthoryear{He, Zhang, Ren, and Sun}{He
  et~al\mbox{.}}{2016}]%
        {He:2015}
\bibfield{author}{\bibinfo{person}{Kaiming He}, \bibinfo{person}{Xiangyu
  Zhang}, \bibinfo{person}{Shaoqing Ren}, {and} \bibinfo{person}{Jian Sun}.}
  \bibinfo{year}{2016}\natexlab{}.
\newblock \showarticletitle{Deep Residual Learning for Image Recognition}. In
  \bibinfo{booktitle}{\emph{IEEE Conference on Computer Vision and Pattern
  Recognition (CVPR)}}.
\newblock


\bibitem[\protect\citeauthoryear{Heinrich}{Heinrich}{1998}]%
        {Heinrich:1998}
\bibfield{author}{\bibinfo{person}{Stefan Heinrich}.}
  \bibinfo{year}{1998}\natexlab{}.
\newblock \showarticletitle{Monte Carlo Complexity of Global Solution of
  Integral Equations}.
\newblock \bibinfo{journal}{\emph{Journal of Complexity}} \bibinfo{volume}{14},
  \bibinfo{number}{2} (\bibinfo{year}{1998}), \bibinfo{pages}{151 -- 175}.
\newblock
\showISSN{0885-064X}
\urldef\tempurl%
\url{https://doi.org/10.1006/jcom.1998.0471}
\showDOI{\tempurl}


\bibitem[\protect\citeauthoryear{Heinrich}{Heinrich}{2000}]%
        {Heinrich:2000}
\bibfield{author}{\bibinfo{person}{Stefan Heinrich}.}
  \bibinfo{year}{2000}\natexlab{}.
\newblock \showarticletitle{The Multilevel Method of Dependent Tests}.
\newblock In \bibinfo{booktitle}{\emph{Advances in Stochastic Simulation
  Methods}}. \bibinfo{publisher}{Birkh{\"a}user Boston},
  \bibinfo{address}{Boston, MA}, \bibinfo{pages}{47--61}.
\newblock
\showISBNx{978-1-4612-1318-5}
\urldef\tempurl%
\url{https://doi.org/10.1007/978-1-4612-1318-5_4}
\showDOI{\tempurl}


\bibitem[\protect\citeauthoryear{Herholz, Zhao, Elek, Nowrouzezahrai, Lensch,
  and K\v{r}iv\'{a}nek}{Herholz et~al\mbox{.}}{2019}]%
        {Herholz:2019}
\bibfield{author}{\bibinfo{person}{Sebastian Herholz},
  \bibinfo{person}{Yangyang Zhao}, \bibinfo{person}{Oskar Elek},
  \bibinfo{person}{Derek Nowrouzezahrai}, \bibinfo{person}{Hendrik P.~A.
  Lensch}, {and} \bibinfo{person}{Jaroslav K\v{r}iv\'{a}nek}.}
  \bibinfo{year}{2019}\natexlab{}.
\newblock \showarticletitle{Volume Path Guiding Based on Zero-Variance Random
  Walk Theory}.
\newblock \bibinfo{journal}{\emph{ACM Trans. Graph.}} \bibinfo{volume}{38},
  \bibinfo{number}{3}, Article \bibinfo{articleno}{25} (\bibinfo{date}{June}
  \bibinfo{year}{2019}), \bibinfo{numpages}{19}~pages.
\newblock
\showISSN{0730-0301}
\urldef\tempurl%
\url{https://doi.org/10.1145/3230635}
\showDOI{\tempurl}


\bibitem[\protect\citeauthoryear{Hermosilla, Maisch, Ritschel, and
  Ropinski}{Hermosilla et~al\mbox{.}}{2019}]%
        {hermosilla2019}
\bibfield{author}{\bibinfo{person}{Pedro Hermosilla},
  \bibinfo{person}{Sebastian Maisch}, \bibinfo{person}{Tobias Ritschel}, {and}
  \bibinfo{person}{Timo Ropinski}.} \bibinfo{year}{2019}\natexlab{}.
\newblock \showarticletitle{Deep-learning the Latent Space of Light Transport}.
\newblock \bibinfo{journal}{\emph{Computer Graphics Forum}}
  \bibinfo{volume}{38}, \bibinfo{number}{4} (\bibinfo{year}{2019}).
\newblock


\bibitem[\protect\citeauthoryear{Hesterberg and Nelson}{Hesterberg and
  Nelson}{1998}]%
        {Hesterberg:1998}
\bibfield{author}{\bibinfo{person}{Timothy~C. Hesterberg} {and}
  \bibinfo{person}{Barry~L. Nelson}.} \bibinfo{year}{1998}\natexlab{}.
\newblock \showarticletitle{Control Variates for Probability and Quantile
  Estimation}.
\newblock \bibinfo{journal}{\emph{Management Science}} \bibinfo{volume}{44},
  \bibinfo{number}{9} (\bibinfo{date}{Sept.} \bibinfo{year}{1998}),
  \bibinfo{pages}{1295--1312}.
\newblock
\showISSN{0025-1909}
\urldef\tempurl%
\url{https://doi.org/10.1287/mnsc.44.9.1295}
\showDOI{\tempurl}


\bibitem[\protect\citeauthoryear{Huang, Krueger, Lacoste, and Courville}{Huang
  et~al\mbox{.}}{2018}]%
        {Huang2018NeuralAF}
\bibfield{author}{\bibinfo{person}{Chin-Wei Huang}, \bibinfo{person}{David
  Krueger}, \bibinfo{person}{Alexandre Lacoste}, {and}
  \bibinfo{person}{Aaron~C. Courville}.} \bibinfo{year}{2018}\natexlab{}.
\newblock \showarticletitle{Neural Autoregressive Flows}.
\newblock \bibinfo{journal}{\emph{arXiv:1804.00779}} (\bibinfo{date}{April}
  \bibinfo{year}{2018}).
\newblock


\bibitem[\protect\citeauthoryear{Ioffe and Szegedy}{Ioffe and Szegedy}{2015}]%
        {ioffe15batchnorm}
\bibfield{author}{\bibinfo{person}{Sergey Ioffe} {and}
  \bibinfo{person}{Christian Szegedy}.} \bibinfo{year}{2015}\natexlab{}.
\newblock \showarticletitle{Batch Normalization: Accelerating Deep Network
  Training by Reducing Internal Covariate Shift}.
\newblock \bibinfo{journal}{\emph{arXiv:1502.03167}} (\bibinfo{year}{2015}).
\newblock


\bibitem[\protect\citeauthoryear{Jakob}{Jakob}{2010}]%
        {Mitsuba}
\bibfield{author}{\bibinfo{person}{Wenzel Jakob}.}
  \bibinfo{year}{2010}\natexlab{}.
\newblock \bibinfo{title}{Mitsuba Renderer}.
\newblock
\newblock
\newblock
\shownote{http://www.mitsuba-renderer.org.}


\bibitem[\protect\citeauthoryear{Kallweit, M\"{u}ller, McWilliams, Gross, and
  Nov\'{a}k}{Kallweit et~al\mbox{.}}{2017}]%
        {Kallweit2017DeepScattering}
\bibfield{author}{\bibinfo{person}{Simon Kallweit}, \bibinfo{person}{Thomas
  M\"{u}ller}, \bibinfo{person}{Brian McWilliams}, \bibinfo{person}{Markus
  Gross}, {and} \bibinfo{person}{Jan Nov\'{a}k}.}
  \bibinfo{year}{2017}\natexlab{}.
\newblock \showarticletitle{Deep Scattering: Rendering Atmospheric Clouds with
  Radiance-Predicting Neural Networks}.
\newblock \bibinfo{journal}{\emph{ACM Trans. Graph.}} \bibinfo{volume}{36},
  \bibinfo{number}{6}, Article \bibinfo{articleno}{231} (\bibinfo{date}{Nov.}
  \bibinfo{year}{2017}), \bibinfo{numpages}{11}~pages.
\newblock
\urldef\tempurl%
\url{https://doi.org/10.1145/3130800.3130880}
\showDOI{\tempurl}


\bibitem[\protect\citeauthoryear{Keller}{Keller}{2001}]%
        {Keller:2001}
\bibfield{author}{\bibinfo{person}{Alexander Keller}.}
  \bibinfo{year}{2001}\natexlab{}.
\newblock \showarticletitle{Hierarchical Monte Carlo Image Synthesis}.
\newblock \bibinfo{journal}{\emph{Mathematics and Computers in Simulation}}
  \bibinfo{volume}{55}, \bibinfo{number}{1–3} (\bibinfo{year}{2001}),
  \bibinfo{pages}{79 -- 92}.
\newblock
\showISSN{0378-4754}
\urldef\tempurl%
\url{https://doi.org/10.1016/S0378-4754(00)00248-2}
\showDOI{\tempurl}
\newblock
\shownote{The Second \{IMACS\} Seminar on Monte Carlo Methods.}


\bibitem[\protect\citeauthoryear{Kemna and Vorst}{Kemna and Vorst}{1990}]%
        {Kemna:1990}
\bibfield{author}{\bibinfo{person}{Angelien Kemna} {and} \bibinfo{person}{Ton
  Vorst}.} \bibinfo{year}{1990}\natexlab{}.
\newblock \showarticletitle{A Pricing Method for Options based on Average Asset
  Values}.
\newblock \bibinfo{journal}{\emph{Journal of Banking \& Finance}}
  \bibinfo{volume}{14}, \bibinfo{number}{1} (\bibinfo{year}{1990}),
  \bibinfo{pages}{113--129}.
\newblock
\showISSN{0378-4266}
\urldef\tempurl%
\url{https://doi.org/10.1016/0378-4266(90)90039-5}
\showDOI{\tempurl}


\bibitem[\protect\citeauthoryear{Kingma and Ba}{Kingma and Ba}{2014}]%
        {KingmaB14}
\bibfield{author}{\bibinfo{person}{Diederik~P. Kingma} {and}
  \bibinfo{person}{Jimmy Ba}.} \bibinfo{year}{2014}\natexlab{}.
\newblock \showarticletitle{Adam: {A} Method for Stochastic Optimization}.
\newblock \bibinfo{journal}{\emph{arXiv:1412.6980}} (\bibinfo{date}{June}
  \bibinfo{year}{2014}).
\newblock


\bibitem[\protect\citeauthoryear{Kingma and Dhariwal}{Kingma and
  Dhariwal}{2018}]%
        {kingma18glow}
\bibfield{author}{\bibinfo{person}{Diederik~P. Kingma} {and}
  \bibinfo{person}{Prafulla Dhariwal}.} \bibinfo{year}{2018}\natexlab{}.
\newblock \showarticletitle{{Glow: Generative Flow with Invertible 1x1
  Convolutions}}.
\newblock \bibinfo{journal}{\emph{arXiv:1807.03039}} (\bibinfo{date}{July}
  \bibinfo{year}{2018}).
\newblock


\bibitem[\protect\citeauthoryear{Kingma, Salimans, Jozefowicz, Chen, Sutskever,
  and Welling}{Kingma et~al\mbox{.}}{2016}]%
        {kingma2016improved}
\bibfield{author}{\bibinfo{person}{Diederik~P. Kingma}, \bibinfo{person}{Tim
  Salimans}, \bibinfo{person}{Rafal Jozefowicz}, \bibinfo{person}{Xi Chen},
  \bibinfo{person}{Ilya Sutskever}, {and} \bibinfo{person}{Max Welling}.}
  \bibinfo{year}{2016}\natexlab{}.
\newblock \showarticletitle{Improved Variational Inference with inverse
  Autoregressive Flow}. In \bibinfo{booktitle}{\emph{Advances in Neural
  Information Processing Systems}}. \bibinfo{pages}{4743--4751}.
\newblock


\bibitem[\protect\citeauthoryear{Kobyzev, Prince, and Brubaker}{Kobyzev
  et~al\mbox{.}}{2019}]%
        {kobyzev2019normalizing}
\bibfield{author}{\bibinfo{person}{Ivan Kobyzev}, \bibinfo{person}{Simon
  Prince}, {and} \bibinfo{person}{Marcus~A. Brubaker}.}
  \bibinfo{year}{2019}\natexlab{}.
\newblock \bibinfo{title}{Normalizing Flows: An Introduction and Review of
  Current Methods}.
\newblock
\newblock
\showeprint[arxiv]{stat.ML/1908.09257}


\bibitem[\protect\citeauthoryear{Kondapaneni, Vevoda, Grittmann, Sk\v{r}ivan,
  Slusallek, and K\v{r}iv\'{a}nek}{Kondapaneni et~al\mbox{.}}{2019}]%
        {Kondapaneni:2019}
\bibfield{author}{\bibinfo{person}{Ivo Kondapaneni}, \bibinfo{person}{Petr
  Vevoda}, \bibinfo{person}{Pascal Grittmann}, \bibinfo{person}{Tom\'{a}\v{s}
  Sk\v{r}ivan}, \bibinfo{person}{Philipp Slusallek}, {and}
  \bibinfo{person}{Jaroslav K\v{r}iv\'{a}nek}.}
  \bibinfo{year}{2019}\natexlab{}.
\newblock \showarticletitle{Optimal Multiple Importance Sampling}.
\newblock \bibinfo{journal}{\emph{ACM Trans. Graph.}} \bibinfo{volume}{38},
  \bibinfo{number}{4}, Article \bibinfo{articleno}{37} (\bibinfo{date}{July}
  \bibinfo{year}{2019}), \bibinfo{numpages}{14}~pages.
\newblock
\showISSN{0730-0301}
\urldef\tempurl%
\url{https://doi.org/10.1145/3306346.3323009}
\showDOI{\tempurl}


\bibitem[\protect\citeauthoryear{Lafortune and Willems}{Lafortune and
  Willems}{1994}]%
        {Lafortune:1994:TAT}
\bibfield{author}{\bibinfo{person}{Eric~P. Lafortune} {and}
  \bibinfo{person}{Yves~D. Willems}.} \bibinfo{year}{1994}\natexlab{}.
\newblock \showarticletitle{The Ambient Term as a Variance Reducing Technique
  for {M}onte {C}arlo Ray Tracing}. In \bibinfo{booktitle}{\emph{Proc.\ EGWR}}.
  \bibinfo{pages}{163--171}.
\newblock


\bibitem[\protect\citeauthoryear{Lafortune and Willems}{Lafortune and
  Willems}{1995}]%
        {Lafortune:1995:A5T}
\bibfield{author}{\bibinfo{person}{Eric~P. Lafortune} {and}
  \bibinfo{person}{Yves~D. Willems}.} \bibinfo{year}{1995}\natexlab{}.
\newblock \showarticletitle{A 5D Tree to Reduce the Variance of {M}onte {C}arlo
  Ray Tracing}. In \bibinfo{booktitle}{\emph{Proc.\ EGWR}}.
  \bibinfo{pages}{11--20}.
\newblock


\bibitem[\protect\citeauthoryear{Lavenberg, Moeller, and Welch}{Lavenberg
  et~al\mbox{.}}{1982}]%
        {Lavenberg:1982}
\bibfield{author}{\bibinfo{person}{Stephen~S. Lavenberg},
  \bibinfo{person}{Thomas~L. Moeller}, {and} \bibinfo{person}{Peter~D. Welch}.}
  \bibinfo{year}{1982}\natexlab{}.
\newblock \showarticletitle{Statistical Results on Control Variables with
  Application to Queueing Network Simulation}.
\newblock \bibinfo{journal}{\emph{Operations Research}} \bibinfo{volume}{30},
  \bibinfo{number}{1} (\bibinfo{year}{1982}), \bibinfo{pages}{182--202}.
\newblock
\urldef\tempurl%
\url{https://doi.org/10.1287/opre.30.1.182}
\showDOI{\tempurl}


\bibitem[\protect\citeauthoryear{Lehtinen, Munkberg, Hasselgren, Laine, Karras,
  Aittala, and Aila}{Lehtinen et~al\mbox{.}}{2018}]%
        {Lehtinen:2018}
\bibfield{author}{\bibinfo{person}{Jaakko Lehtinen}, \bibinfo{person}{Jacob
  Munkberg}, \bibinfo{person}{Jon Hasselgren}, \bibinfo{person}{Samuli Laine},
  \bibinfo{person}{Tero Karras}, \bibinfo{person}{Miika Aittala}, {and}
  \bibinfo{person}{Timo Aila}.} \bibinfo{year}{2018}\natexlab{}.
\newblock \bibinfo{title}{Noise2Noise: Learning Image Restoration without Clean
  Data}.
\newblock
\newblock
\showeprint[arxiv]{cs.CV/1803.04189}


\bibitem[\protect\citeauthoryear{Lombardi, Simon, Saragih, Schwartz, Lehrmann,
  and Sheikh}{Lombardi et~al\mbox{.}}{2019}]%
        {Lombardi:2019}
\bibfield{author}{\bibinfo{person}{Stephen Lombardi}, \bibinfo{person}{Tomas
  Simon}, \bibinfo{person}{Jason Saragih}, \bibinfo{person}{Gabriel Schwartz},
  \bibinfo{person}{Andreas Lehrmann}, {and} \bibinfo{person}{Yaser Sheikh}.}
  \bibinfo{year}{2019}\natexlab{}.
\newblock \showarticletitle{Neural Volumes: Learning Dynamic Renderable Volumes
  from Images}.
\newblock \bibinfo{journal}{\emph{ACM Trans. Graph.}} \bibinfo{volume}{38},
  \bibinfo{number}{4}, Article \bibinfo{articleno}{65} (\bibinfo{date}{July}
  \bibinfo{year}{2019}), \bibinfo{numpages}{14}~pages.
\newblock
\showISSN{0730-0301}
\urldef\tempurl%
\url{https://doi.org/10.1145/3306346.3323020}
\showDOI{\tempurl}


\bibitem[\protect\citeauthoryear{Maximov, Leal-Taixe, Fritz, and
  Ritschel}{Maximov et~al\mbox{.}}{2019}]%
        {Maximov_2019_ICCV}
\bibfield{author}{\bibinfo{person}{Maxim Maximov}, \bibinfo{person}{Laura
  Leal-Taixe}, \bibinfo{person}{Mario Fritz}, {and} \bibinfo{person}{Tobias
  Ritschel}.} \bibinfo{year}{2019}\natexlab{}.
\newblock \showarticletitle{Deep Appearance Maps}. In
  \bibinfo{booktitle}{\emph{The IEEE International Conference on Computer
  Vision (ICCV)}}.
\newblock


\bibitem[\protect\citeauthoryear{Meka, H\"{a}ne, Pandey, Zollh\"{o}fer,
  Fanello, Fyffe, Kowdle, Yu, Busch, Dourgarian, Denny, Bouaziz, Lincoln,
  Whalen, Harvey, Taylor, Izadi, Tagliasacchi, Debevec, Theobalt, Valentin, and
  Rhemann}{Meka et~al\mbox{.}}{2019}]%
        {Meka2019}
\bibfield{author}{\bibinfo{person}{Abhimitra Meka}, \bibinfo{person}{Christian
  H\"{a}ne}, \bibinfo{person}{Rohit Pandey}, \bibinfo{person}{Michael
  Zollh\"{o}fer}, \bibinfo{person}{Sean Fanello}, \bibinfo{person}{Graham
  Fyffe}, \bibinfo{person}{Adarsh Kowdle}, \bibinfo{person}{Xueming Yu},
  \bibinfo{person}{Jay Busch}, \bibinfo{person}{Jason Dourgarian},
  \bibinfo{person}{Peter Denny}, \bibinfo{person}{Sofien Bouaziz},
  \bibinfo{person}{Peter Lincoln}, \bibinfo{person}{Matt Whalen},
  \bibinfo{person}{Geoff Harvey}, \bibinfo{person}{Jonathan Taylor},
  \bibinfo{person}{Shahram Izadi}, \bibinfo{person}{Andrea Tagliasacchi},
  \bibinfo{person}{Paul Debevec}, \bibinfo{person}{Christian Theobalt},
  \bibinfo{person}{Julien Valentin}, {and} \bibinfo{person}{Christoph
  Rhemann}.} \bibinfo{year}{2019}\natexlab{}.
\newblock \showarticletitle{Deep Reflectance Fields: High-quality Facial
  Reflectance Field Inference from Color Gradient Illumination}.
\newblock \bibinfo{journal}{\emph{ACM Trans. Graph.}} \bibinfo{volume}{38},
  \bibinfo{number}{4}, Article \bibinfo{articleno}{77} (\bibinfo{date}{July}
  \bibinfo{year}{2019}), \bibinfo{numpages}{12}~pages.
\newblock
\showISSN{0730-0301}
\urldef\tempurl%
\url{https://doi.org/10.1145/3306346.3323027}
\showDOI{\tempurl}


\bibitem[\protect\citeauthoryear{Mira, Solgi, and Imparato}{Mira
  et~al\mbox{.}}{2013}]%
        {Mira2013ZeroVM}
\bibfield{author}{\bibinfo{person}{Antonietta Mira}, \bibinfo{person}{Reza
  Solgi}, {and} \bibinfo{person}{Daniele Imparato}.}
  \bibinfo{year}{2013}\natexlab{}.
\newblock \showarticletitle{Zero variance Markov chain Monte Carlo for Bayesian
  estimators}.
\newblock \bibinfo{journal}{\emph{Statistics and Computing}}
  \bibinfo{volume}{23} (\bibinfo{year}{2013}), \bibinfo{pages}{653--662}.
\newblock


\bibitem[\protect\citeauthoryear{M\"{u}ller}{M\"{u}ller}{2019}]%
        {mueller19guiding}
\bibfield{author}{\bibinfo{person}{Thomas M\"{u}ller}.}
  \bibinfo{year}{2019}\natexlab{}.
\newblock \showarticletitle{``Practical Path Guiding'' in Production}. In
  \bibinfo{booktitle}{\emph{ACM SIGGRAPH Courses: Path Guiding in Production,
  Chapter 10}}. \bibinfo{publisher}{ACM}, \bibinfo{address}{New York, NY, USA},
  \bibinfo{pages}{18:1--18:77}.
\newblock
\urldef\tempurl%
\url{https://doi.org/10.1145/3305366.3328091}
\showDOI{\tempurl}


\bibitem[\protect\citeauthoryear{M\"{u}ller, Gross, and Nov\'{a}k}{M\"{u}ller
  et~al\mbox{.}}{2017}]%
        {mueller2017practical}
\bibfield{author}{\bibinfo{person}{Thomas M\"{u}ller}, \bibinfo{person}{Markus
  Gross}, {and} \bibinfo{person}{Jan Nov\'{a}k}.}
  \bibinfo{year}{2017}\natexlab{}.
\newblock \showarticletitle{Practical Path Guiding for Efficient
  Light-Transport Simulation}.
\newblock \bibinfo{journal}{\emph{Computer Graphics Forum}}
  \bibinfo{volume}{36}, \bibinfo{number}{4} (\bibinfo{date}{June}
  \bibinfo{year}{2017}), \bibinfo{pages}{91--100}.
\newblock
\showISSN{1467-8659}
\urldef\tempurl%
\url{https://doi.org/10.1111/cgf.13227}
\showDOI{\tempurl}


\bibitem[\protect\citeauthoryear{M\"uller, Mcwilliams, Rousselle, Gross, and
  Nov\'ak}{M\"uller et~al\mbox{.}}{2019}]%
        {mueller2019nis}
\bibfield{author}{\bibinfo{person}{Thomas M\"uller}, \bibinfo{person}{Brian
  Mcwilliams}, \bibinfo{person}{Fabrice Rousselle}, \bibinfo{person}{Markus
  Gross}, {and} \bibinfo{person}{Jan Nov\'ak}.}
  \bibinfo{year}{2019}\natexlab{}.
\newblock \showarticletitle{Neural Importance Sampling}.
\newblock \bibinfo{journal}{\emph{ACM Trans. Graph.}} \bibinfo{volume}{38},
  \bibinfo{number}{5}, Article \bibinfo{articleno}{145} (\bibinfo{date}{Oct.}
  \bibinfo{year}{2019}), \bibinfo{numpages}{19}~pages.
\newblock
\showISSN{0730-0301}
\urldef\tempurl%
\url{https://doi.org/10.1145/3341156}
\showDOI{\tempurl}


\bibitem[\protect\citeauthoryear{Nalbach, Arabadzhiyska, Mehta, Seidel, and
  Ritschel}{Nalbach et~al\mbox{.}}{2017}]%
        {Nalbach2017b}
\bibfield{author}{\bibinfo{person}{Oliver Nalbach}, \bibinfo{person}{Elena
  Arabadzhiyska}, \bibinfo{person}{Dushyant Mehta}, \bibinfo{person}{Hans-Peter
  Seidel}, {and} \bibinfo{person}{Tobias Ritschel}.}
  \bibinfo{year}{2017}\natexlab{}.
\newblock \showarticletitle{Deep Shading: Convolutional Neural Networks for
  Screen-Space Shading}.
\newblock  \bibinfo{volume}{36}, \bibinfo{number}{4} (\bibinfo{year}{2017}).
\newblock


\bibitem[\protect\citeauthoryear{Nelson}{Nelson}{1990}]%
        {Nelson:1990}
\bibfield{author}{\bibinfo{person}{Barry~L. Nelson}.}
  \bibinfo{year}{1990}\natexlab{}.
\newblock \showarticletitle{Control Variate Remedies}.
\newblock \bibinfo{journal}{\emph{Operations Research}} \bibinfo{volume}{38},
  \bibinfo{number}{6} (\bibinfo{year}{1990}), \bibinfo{pages}{974--992}.
\newblock
\urldef\tempurl%
\url{https://doi.org/10.1287/opre.38.6.974}
\showDOI{\tempurl}


\bibitem[\protect\citeauthoryear{Novák, Selle, and Jarosz}{Novák
  et~al\mbox{.}}{2014}]%
        {Novak:14:RRT}
\bibfield{author}{\bibinfo{person}{Jan Novák}, \bibinfo{person}{Andrew Selle},
  {and} \bibinfo{person}{Wojciech Jarosz}.} \bibinfo{year}{2014}\natexlab{}.
\newblock \showarticletitle{Residual Ratio Tracking for Estimating Attenuation
  in Participating Media}.
\newblock \bibinfo{journal}{\emph{ACM Trans. Graph.}} \bibinfo{volume}{33},
  \bibinfo{number}{6} (\bibinfo{date}{Nov.} \bibinfo{year}{2014}).
\newblock
\urldef\tempurl%
\url{https://doi.org/10.1145/2661229.2661292}
\showDOI{\tempurl}


\bibitem[\protect\citeauthoryear{Oates, Girolami, and Chopin}{Oates
  et~al\mbox{.}}{2014}]%
        {Oates2014ControlFF}
\bibfield{author}{\bibinfo{person}{C. Oates}, \bibinfo{person}{M. Girolami},
  {and} \bibinfo{person}{N. Chopin}.} \bibinfo{year}{2014}\natexlab{}.
\newblock \showarticletitle{Control functionals for Monte Carlo integration}.
\newblock \bibinfo{journal}{\emph{Journal of The Royal Statistical Society
  Series B-statistical Methodology}}  \bibinfo{volume}{79}
  (\bibinfo{year}{2014}), \bibinfo{pages}{695--718}.
\newblock


\bibitem[\protect\citeauthoryear{Owen and Zhou}{Owen and Zhou}{2000}]%
        {owen00safe}
\bibfield{author}{\bibinfo{person}{Art Owen} {and} \bibinfo{person}{Yi Zhou}.}
  \bibinfo{year}{2000}\natexlab{}.
\newblock \showarticletitle{Safe and Effective Importance Sampling}.
\newblock \bibinfo{journal}{\emph{J. Amer. Statist. Assoc.}}
  \bibinfo{volume}{95}, \bibinfo{number}{449} (\bibinfo{year}{2000}),
  \bibinfo{pages}{135--143}.
\newblock
\showISSN{01621459}
\urldef\tempurl%
\url{http://www.jstor.org/stable/2669533}
\showURL{%
\tempurl}


\bibitem[\protect\citeauthoryear{Papamakarios, Murray, and
  Pavlakou}{Papamakarios et~al\mbox{.}}{2017}]%
        {papamakarios2017masked}
\bibfield{author}{\bibinfo{person}{George Papamakarios}, \bibinfo{person}{Iain
  Murray}, {and} \bibinfo{person}{Theo Pavlakou}.}
  \bibinfo{year}{2017}\natexlab{}.
\newblock \showarticletitle{Masked Autoregressive Flow for Density Estimation}.
  In \bibinfo{booktitle}{\emph{Advances in Neural Information Processing
  Systems}}. \bibinfo{pages}{2338--2347}.
\newblock


\bibitem[\protect\citeauthoryear{Papamakarios, Nalisnick, Rezende, Mohamed, and
  Lakshminarayanan}{Papamakarios et~al\mbox{.}}{2019}]%
        {papamakarios2019normalizing}
\bibfield{author}{\bibinfo{person}{George Papamakarios}, \bibinfo{person}{Eric
  Nalisnick}, \bibinfo{person}{Danilo~Jimenez Rezende}, \bibinfo{person}{Shakir
  Mohamed}, {and} \bibinfo{person}{Balaji Lakshminarayanan}.}
  \bibinfo{year}{2019}\natexlab{}.
\newblock \bibinfo{title}{Normalizing Flows for Probabilistic Modeling and
  Inference}.
\newblock
\newblock
\showeprint[arxiv]{stat.ML/1912.02762}


\bibitem[\protect\citeauthoryear{Pegoraro, Brownlee, Shirley, and
  Parker}{Pegoraro et~al\mbox{.}}{2008a}]%
        {PegoraroIRT08TIGIESMCA}
\bibfield{author}{\bibinfo{person}{Vincent Pegoraro}, \bibinfo{person}{Carson
  Brownlee}, \bibinfo{person}{Peter~S. Shirley}, {and}
  \bibinfo{person}{Steven~G. Parker}.} \bibinfo{year}{2008}\natexlab{a}.
\newblock \showarticletitle{Towards Interactive Global Illumination Effects via
  Sequential Monte Carlo Adaptation}. In \bibinfo{booktitle}{\emph{Proceedings
  of the 3rd IEEE Symposium on Interactive Ray Tracing}}.
  \bibinfo{pages}{107--114}.
\newblock


\bibitem[\protect\citeauthoryear{Pegoraro, Wald, and Parker}{Pegoraro
  et~al\mbox{.}}{2008b}]%
        {PegoraroEGSR08SMCALAPM}
\bibfield{author}{\bibinfo{person}{Vincent Pegoraro}, \bibinfo{person}{Ingo
  Wald}, {and} \bibinfo{person}{Steven~G. Parker}.}
  \bibinfo{year}{2008}\natexlab{b}.
\newblock \showarticletitle{Sequential Monte Carlo Adaptation in Low-Anisotropy
  Participating Media}.
\newblock \bibinfo{journal}{\emph{Computer Graphics Forum}}
  \bibinfo{volume}{27}, \bibinfo{number}{4} (\bibinfo{year}{2008}),
  \bibinfo{pages}{1097--1104}.
\newblock


\bibitem[\protect\citeauthoryear{Pharr, Jacob, and Humphreys}{Pharr
  et~al\mbox{.}}{2016}]%
        {PBRT}
\bibfield{author}{\bibinfo{person}{Matt Pharr}, \bibinfo{person}{Wenzel Jacob},
  {and} \bibinfo{person}{Greg Humphreys}.} \bibinfo{year}{2016}\natexlab{}.
\newblock \bibinfo{booktitle}{\emph{{Physically Based Rendering - From Theory
  to Implementation}}}.
\newblock \bibinfo{publisher}{Morgan Kaufmann, Third Edition}.
\newblock


\bibitem[\protect\citeauthoryear{Ren, Wang, Gong, Lin, Tong, and Guo}{Ren
  et~al\mbox{.}}{2013}]%
        {Ren:2013}
\bibfield{author}{\bibinfo{person}{Peiran Ren}, \bibinfo{person}{Jiaping Wang},
  \bibinfo{person}{Minmin Gong}, \bibinfo{person}{Stephen Lin},
  \bibinfo{person}{Xin Tong}, {and} \bibinfo{person}{Baining Guo}.}
  \bibinfo{year}{2013}\natexlab{}.
\newblock \showarticletitle{Global Illumination with Radiance Regression
  Functions}.
\newblock \bibinfo{journal}{\emph{ACM Trans. Graph.}} \bibinfo{volume}{32},
  \bibinfo{number}{4}, Article \bibinfo{articleno}{130} (\bibinfo{date}{July}
  \bibinfo{year}{2013}), \bibinfo{numpages}{12}~pages.
\newblock
\showISSN{0730-0301}
\urldef\tempurl%
\url{https://doi.org/10.1145/2461912.2462009}
\showDOI{\tempurl}


\bibitem[\protect\citeauthoryear{Rezende and Mohamed}{Rezende and
  Mohamed}{2015}]%
        {rezende2015variational}
\bibfield{author}{\bibinfo{person}{Danilo Rezende} {and}
  \bibinfo{person}{Shakir Mohamed}.} \bibinfo{year}{2015}\natexlab{}.
\newblock \showarticletitle{Variational Inference with Normalizing Flows}. In
  \bibinfo{booktitle}{\emph{International Conference on Machine Learning}}.
  \bibinfo{pages}{1530--1538}.
\newblock


\bibitem[\protect\citeauthoryear{Rousselle, Jarosz, and Nov\'{a}k}{Rousselle
  et~al\mbox{.}}{2016}]%
        {Rousselle:2016:ICV}
\bibfield{author}{\bibinfo{person}{Fabrice Rousselle},
  \bibinfo{person}{Wojciech Jarosz}, {and} \bibinfo{person}{Jan Nov\'{a}k}.}
  \bibinfo{year}{2016}\natexlab{}.
\newblock \showarticletitle{Image-space Control Variates for Rendering}.
\newblock \bibinfo{journal}{\emph{ACM Trans. Graph.}} \bibinfo{volume}{35},
  \bibinfo{number}{6}, Article \bibinfo{articleno}{169} (\bibinfo{date}{Nov.}
  \bibinfo{year}{2016}), \bibinfo{numpages}{12}~pages.
\newblock
\showISSN{0730-0301}
\urldef\tempurl%
\url{https://doi.org/10.1145/2980179.2982443}
\showDOI{\tempurl}


\bibitem[\protect\citeauthoryear{Rousselle, Knaus, and Zwicker}{Rousselle
  et~al\mbox{.}}{2011}]%
        {Rousselle:2011}
\bibfield{author}{\bibinfo{person}{Fabrice Rousselle}, \bibinfo{person}{Claude
  Knaus}, {and} \bibinfo{person}{Matthias Zwicker}.}
  \bibinfo{year}{2011}\natexlab{}.
\newblock \showarticletitle{Adaptive Sampling and Reconstruction Using Greedy
  Error Minimization}.
\newblock \bibinfo{journal}{\emph{ACM Trans. Graph.}} \bibinfo{volume}{30},
  \bibinfo{number}{6}, Article \bibinfo{articleno}{159} (\bibinfo{date}{Dec.}
  \bibinfo{year}{2011}), \bibinfo{numpages}{12}~pages.
\newblock
\showISSN{0730-0301}
\urldef\tempurl%
\url{https://doi.org/10.1145/2070781.2024193}
\showDOI{\tempurl}


\bibitem[\protect\citeauthoryear{Sitzmann, Thies, Heide, Nie{\ss}ner,
  Wetzstein, and Zollh{\"o}fer}{Sitzmann et~al\mbox{.}}{2018}]%
        {Sitzmann2018DeepVoxelsLP}
\bibfield{author}{\bibinfo{person}{Vincent Sitzmann}, \bibinfo{person}{Justus
  Thies}, \bibinfo{person}{Felix Heide}, \bibinfo{person}{Matthias
  Nie{\ss}ner}, \bibinfo{person}{Gordon Wetzstein}, {and}
  \bibinfo{person}{Michael Zollh{\"o}fer}.} \bibinfo{year}{2018}\natexlab{}.
\newblock \showarticletitle{DeepVoxels: Learning Persistent 3D Feature
  Embeddings}. In \bibinfo{booktitle}{\emph{CVPR}}.
\newblock


\bibitem[\protect\citeauthoryear{Stein}{Stein}{1972}]%
        {stein1972}
\bibfield{author}{\bibinfo{person}{Charles Stein}.}
  \bibinfo{year}{1972}\natexlab{}.
\newblock \showarticletitle{A bound for the error in the normal approximation
  to the distribution of a sum of dependent random variables}. In
  \bibinfo{booktitle}{\emph{Proceedings of the Sixth Berkeley Symposium on
  Mathematical Statistics and Probability, Volume 2: Probability Theory}}.
  \bibinfo{publisher}{University of California Press},
  \bibinfo{address}{Berkeley, Calif.}, \bibinfo{pages}{583--602}.
\newblock
\urldef\tempurl%
\url{https://projecteuclid.org/euclid.bsmsp/1200514239}
\showURL{%
\tempurl}


\bibitem[\protect\citeauthoryear{Sz{\'e}csi, Sbert, and
  Szirmay-Kalos}{Sz{\'e}csi et~al\mbox{.}}{2004}]%
        {Szecsi:2004:CCI}
\bibfield{author}{\bibinfo{person}{L{\'a}szl{\'o} Sz{\'e}csi},
  \bibinfo{person}{Mateu Sbert}, {and} \bibinfo{person}{L{\'a}szl{\'o}
  Szirmay-Kalos}.} \bibinfo{year}{2004}\natexlab{}.
\newblock \showarticletitle{Combined Correlated and Importance Sampling in
  Direct Light Source Computation and Environment Mapping}.
\newblock \bibinfo{journal}{\emph{Computer Graphics Forum}}
  \bibinfo{volume}{23} (\bibinfo{year}{2004}), \bibinfo{pages}{585--594}.
\newblock


\bibitem[\protect\citeauthoryear{Szirmay-Kalos, T\'oth, and
  Magdics}{Szirmay-Kalos et~al\mbox{.}}{2011}]%
        {SzirmayKalos11}
\bibfield{author}{\bibinfo{person}{L\'aszl\'o Szirmay-Kalos},
  \bibinfo{person}{Bal\'azs T\'oth}, {and} \bibinfo{person}{Mil\'an Magdics}.}
  \bibinfo{year}{2011}\natexlab{}.
\newblock \showarticletitle{Free Path Sampling in High Resolution Inhomogeneous
  Participating Media}.
\newblock \bibinfo{journal}{\emph{{Computer Graphics Forum}}}
  \bibinfo{volume}{30}, \bibinfo{number}{1} (\bibinfo{year}{2011}),
  \bibinfo{pages}{85--97}.
\newblock


\bibitem[\protect\citeauthoryear{Tabak and Turner}{Tabak and Turner}{2013}]%
        {Tabak:2013}
\bibfield{author}{\bibinfo{person}{Esteban Tabak} {and}
  \bibinfo{person}{{Cristina V.} Turner}.} \bibinfo{year}{2013}\natexlab{}.
\newblock \showarticletitle{A Family of Nonparametric Density Estimation
  Algorithms}.
\newblock \bibinfo{journal}{\emph{Communications on Pure and Applied
  Mathematics}} \bibinfo{volume}{66}, \bibinfo{number}{2}
  (\bibinfo{year}{2013}), \bibinfo{pages}{145--164}.
\newblock
\urldef\tempurl%
\url{https://doi.org/10.1002/cpa.21423}
\showDOI{\tempurl}


\bibitem[\protect\citeauthoryear{Tabak and {Vanden Eijnden}}{Tabak and {Vanden
  Eijnden}}{2010}]%
        {Tabak:2010}
\bibfield{author}{\bibinfo{person}{Esteban Tabak} {and} \bibinfo{person}{Eric
  {Vanden Eijnden}}.} \bibinfo{year}{2010}\natexlab{}.
\newblock \showarticletitle{Density Estimation by dual Ascent of the
  Log-Likelihood}.
\newblock \bibinfo{journal}{\emph{Communications in Mathematical Sciences}}
  \bibinfo{volume}{8}, \bibinfo{number}{1} (\bibinfo{year}{2010}),
  \bibinfo{pages}{217--233}.
\newblock
\showISSN{1539-6746}


\bibitem[\protect\citeauthoryear{Tewari, Fried, Thies, Sitzmann, Lombardi,
  Sunkavalli, Martin-Brualla, Simon, Saragih, Nießner, Pandey, Fanello,
  Wetzstein, Zhu, Theobalt, Agrawala, Shechtman, Goldman, and
  Zollhöfer}{Tewari et~al\mbox{.}}{2020}]%
        {tewari2020state}
\bibfield{author}{\bibinfo{person}{Ayush Tewari}, \bibinfo{person}{Ohad Fried},
  \bibinfo{person}{Justus Thies}, \bibinfo{person}{Vincent Sitzmann},
  \bibinfo{person}{Stephen Lombardi}, \bibinfo{person}{Kalyan Sunkavalli},
  \bibinfo{person}{Ricardo Martin-Brualla}, \bibinfo{person}{Tomas Simon},
  \bibinfo{person}{Jason Saragih}, \bibinfo{person}{Matthias Nießner},
  \bibinfo{person}{Rohit Pandey}, \bibinfo{person}{Sean Fanello},
  \bibinfo{person}{Gordon Wetzstein}, \bibinfo{person}{Jun-Yan Zhu},
  \bibinfo{person}{Christian Theobalt}, \bibinfo{person}{Maneesh Agrawala},
  \bibinfo{person}{Eli Shechtman}, \bibinfo{person}{Dan~B Goldman}, {and}
  \bibinfo{person}{Michael Zollhöfer}.} \bibinfo{year}{2020}\natexlab{}.
\newblock \bibinfo{title}{State of the Art on Neural Rendering}.
\newblock
\newblock
\showeprint[arxiv]{cs.CV/2004.03805}


\bibitem[\protect\citeauthoryear{Thies, Zollh{\"o}fer, and Nie{\ss}ner}{Thies
  et~al\mbox{.}}{2019}]%
        {Thies:2019}
\bibfield{author}{\bibinfo{person}{Justus Thies}, \bibinfo{person}{Michael
  Zollh{\"o}fer}, {and} \bibinfo{person}{Matthias Nie{\ss}ner}.}
  \bibinfo{year}{2019}\natexlab{}.
\newblock \showarticletitle{Deferred Neural Rendering: Image Synthesis Using
  Neural Textures}.
\newblock \bibinfo{journal}{\emph{ACM Trans. Graph.}} \bibinfo{volume}{38},
  \bibinfo{number}{4}, Article \bibinfo{articleno}{66} (\bibinfo{date}{July}
  \bibinfo{year}{2019}), \bibinfo{numpages}{12}~pages.
\newblock
\showISSN{0730-0301}
\urldef\tempurl%
\url{https://doi.org/10.1145/3306346.3323035}
\showDOI{\tempurl}


\bibitem[\protect\citeauthoryear{Veach}{Veach}{1997}]%
        {Veach:1997:Thesis}
\bibfield{author}{\bibinfo{person}{Eric Veach}.}
  \bibinfo{year}{1997}\natexlab{}.
\newblock \emph{\bibinfo{title}{Robust {M}onte {C}arlo methods for light
  transport simulation}}.
\newblock \bibinfo{thesistype}{Ph.D. Dissertation}. \bibinfo{address}{Stanford,
  CA, USA}.
\newblock
\showISBNx{0-591-90780-1}


\bibitem[\protect\citeauthoryear{Veach and Guibas}{Veach and Guibas}{1995}]%
        {Veach:1995:MIS}
\bibfield{author}{\bibinfo{person}{Eric Veach} {and}
  \bibinfo{person}{Leonidas~J. Guibas}.} \bibinfo{year}{1995}\natexlab{}.
\newblock \showarticletitle{Optimally Combining Sampling Techniques for Monte
  Carlo Rendering}. In \bibinfo{booktitle}{\emph{Proc.\ {SIGGRAPH}}}.
  \bibinfo{pages}{419--428}.
\newblock
\showISBNx{0-89791-701-4}
\urldef\tempurl%
\url{https://doi.org/10.1145/218380.218498}
\showDOI{\tempurl}


\bibitem[\protect\citeauthoryear{Vicini, Koltun, and Jakob}{Vicini
  et~al\mbox{.}}{2019}]%
        {Vicini:2019}
\bibfield{author}{\bibinfo{person}{Delio Vicini}, \bibinfo{person}{Vladlen
  Koltun}, {and} \bibinfo{person}{Wenzel Jakob}.}
  \bibinfo{year}{2019}\natexlab{}.
\newblock \showarticletitle{A Learned Shape-Adaptive Subsurface Scattering
  Model}.
\newblock \bibinfo{journal}{\emph{ACM Trans. Graph.}} \bibinfo{volume}{38},
  \bibinfo{number}{4}, Article \bibinfo{articleno}{127} (\bibinfo{date}{July}
  \bibinfo{year}{2019}), \bibinfo{numpages}{15}~pages.
\newblock
\showISSN{0730-0301}
\urldef\tempurl%
\url{https://doi.org/10.1145/3306346.3322974}
\showDOI{\tempurl}


\bibitem[\protect\citeauthoryear{Vidales, Siska, and Szpruch}{Vidales
  et~al\mbox{.}}{2018}]%
        {vidales2018unbiased}
\bibfield{author}{\bibinfo{person}{Marc~Sabate Vidales}, \bibinfo{person}{David
  Siska}, {and} \bibinfo{person}{Lukasz Szpruch}.}
  \bibinfo{year}{2018}\natexlab{}.
\newblock \showarticletitle{Unbiased deep solvers for parametric PDEs}.
\newblock \bibinfo{journal}{\emph{arXiv:1810.05094}} (\bibinfo{date}{Oct.}
  \bibinfo{year}{2018}).
\newblock


\bibitem[\protect\citeauthoryear{Vorba, Karl\'ik, \v{S}ik, Ritschel, and
  K\v{r}iv\'{a}nek}{Vorba et~al\mbox{.}}{2014}]%
        {Vorba:2014:OnlineLearningPMMinLTS}
\bibfield{author}{\bibinfo{person}{Ji\v{r}\'i Vorba},
  \bibinfo{person}{Ond\v{r}ej Karl\'ik}, \bibinfo{person}{Martin \v{S}ik},
  \bibinfo{person}{Tobias Ritschel}, {and} \bibinfo{person}{Jaroslav
  K\v{r}iv\'{a}nek}.} \bibinfo{year}{2014}\natexlab{}.
\newblock \showarticletitle{On-line Learning of Parametric Mixture Models for
  Light Transport Simulation}.
\newblock \bibinfo{journal}{\emph{ACM Trans. Graph.}} \bibinfo{volume}{33},
  \bibinfo{number}{4} (\bibinfo{date}{Aug.} \bibinfo{year}{2014}).
\newblock


\bibitem[\protect\citeauthoryear{Vorba and K\v{r}iv\'{a}nek}{Vorba and
  K\v{r}iv\'{a}nek}{2016}]%
        {VorbaKrivanek:2016:ADRRS}
\bibfield{author}{\bibinfo{person}{Ji\v{r}\'i Vorba} {and}
  \bibinfo{person}{Jaroslav K\v{r}iv\'{a}nek}.}
  \bibinfo{year}{2016}\natexlab{}.
\newblock \showarticletitle{Adjoint-Driven Russian Roulette and Splitting in
  Light Transport Simulation}.
\newblock \bibinfo{journal}{\emph{ACM Trans. Graph.}} \bibinfo{volume}{35},
  \bibinfo{number}{4} (\bibinfo{date}{jul} \bibinfo{year}{2016}).
\newblock


\bibitem[\protect\citeauthoryear{Wan, Zhong, Xiong, and Zhu}{Wan
  et~al\mbox{.}}{2019}]%
        {wan2019neural}
\bibfield{author}{\bibinfo{person}{Ruosi Wan}, \bibinfo{person}{Mingjun Zhong},
  \bibinfo{person}{Haoyi Xiong}, {and} \bibinfo{person}{Zhanxing Zhu}.}
  \bibinfo{year}{2019}\natexlab{}.
\newblock \showarticletitle{Neural Control Variates for Variance Reduction}.
\newblock \bibinfo{journal}{\emph{arXiv:1806.00159}} (\bibinfo{date}{Oct.}
  \bibinfo{year}{2019}).
\newblock


\bibitem[\protect\citeauthoryear{Yamaguchi, Yatagawa, and Morishima}{Yamaguchi
  et~al\mbox{.}}{2018}]%
        {yamaguchi18efficient}
\bibfield{author}{\bibinfo{person}{Tomoya Yamaguchi}, \bibinfo{person}{Tatsuya
  Yatagawa}, {and} \bibinfo{person}{Shigeo Morishima}.}
  \bibinfo{year}{2018}\natexlab{}.
\newblock \showarticletitle{{Efficient Metropolis Path Sampling for Material
  Editing and Re-rendering}}. In \bibinfo{booktitle}{\emph{Pacific Graphics
  Short Papers}}, \bibfield{editor}{\bibinfo{person}{Hongbo Fu},
  \bibinfo{person}{Abhijeet Ghosh}, {and} \bibinfo{person}{Johannes Kopf}}
  (Eds.). \bibinfo{publisher}{The Eurographics Association}.
\newblock
\showISBNx{978-3-03868-073-4}
\urldef\tempurl%
\url{https://doi.org/10.2312/pg.20181271}
\showDOI{\tempurl}


\bibitem[\protect\citeauthoryear{Zheng and Zwicker}{Zheng and Zwicker}{2019}]%
        {Zheng:2019}
\bibfield{author}{\bibinfo{person}{Quan Zheng} {and} \bibinfo{person}{Matthias
  Zwicker}.} \bibinfo{year}{2019}\natexlab{}.
\newblock \showarticletitle{Learning to Importance Sample in Primary Sample
  Space}.
\newblock \bibinfo{journal}{\emph{{Computer Graphics Forum}}}
  \bibinfo{volume}{38}, \bibinfo{number}{2} (\bibinfo{year}{2019}),
  \bibinfo{pages}{169--179}.
\newblock
\urldef\tempurl%
\url{https://doi.org/10.1111/cgf.13628}
\showDOI{\tempurl}


\end{thebibliography}
\bibliographystyle{ACM-Reference-Format}

\end{document}